\documentclass[preprint,12pt]{elsarticle}

\usepackage{amssymb}
\usepackage{ amsmath}
\usepackage{hyperref}
\usepackage{multirow,tabularx}
\usepackage{adjustbox}
\usepackage{textcomp}
\usepackage{etoolbox,lipsum,cleveref}
\usepackage[font=footnotesize,labelfont=bf]{caption}
\usepackage[font=footnotesize,labelfont=bf]{subcaption}

\usepackage{pifont}
\newcommand{\cmark}{\ding{51}}%
\newcommand{\xmark}{\ding{55}}%
\usepackage{footnote}
\usepackage[flushleft]{threeparttable}

\usepackage{makecell}

\begin{document}

\begin{frontmatter}



\title{ Deep learning for table detection and structure recognition: A survey}


\author[inst1]{Mahmoud Kasem}
\author[inst1]{Abdelrahman Abdallah}
\author[inst4]{Alexander Berendeyev}
\author[inst1]{Ebrahem Elkady}
\author[inst5]{Mahmoud Abdalla}
\author[inst1,inst6]{Mohamed Mahmoud}
\author[inst2]{Mohamed Hamada}
\author[inst4,inst3]{Daniyar Nurseitov}
\author[inst1]{Islam Taj-Eddin}

\affiliation[inst1]{organization={Information Technology Department, FCI},
            addressline={Assiut University}, 
            city={Assiut},
            postcode={71515}, 
            state={Assiut},
            country={Egypt}}

\affiliation[inst2]{
    organization={Department of Information System, International IT University,},
    city={Almaty},
    postcode={050000}, 
    country={Kazakhstan}
}

\affiliation[inst3]{
    organization={KazMunayGas Engineering LLP},
    city={Nur-Sultan},
    postcode={010000}, 
    country={Kazakhstan}
}

\affiliation[inst4]{
    organization={Satbayev University},
    city={Almaty},
    postcode={050013}, 
    country={Kazakhstan}
}

\affiliation[inst5]{
    organization={Information Technology Institute(ITI)},
    city={Alexandria},
    postcode={5310002}, 
    country={Egypt}
}

\affiliation[inst6]{
    organization={College of Electrical and Computer Engineering, Chungbuk National University},
    city={Cheongju},
    postcode={28644}, 
    country={South Korea}
}

\begin{abstract}
Tables are everywhere, from scientific journals, papers, websites, and newspapers all the way to items we buy at the supermarket. Detecting them is thus of utmost importance to automatically understanding the content of a document. The performance of table detection has substantially increased thanks to the rapid development of deep learning networks. The goals of this survey are to provide a profound comprehension of the major developments in the field of Table Detection, offer insight into the different methodologies, and provide a systematic taxonomy of the different approaches. Furthermore, we provide an analysis of both classic and new applications in the field. Lastly, the datasets and source code of the existing models are organized to provide the reader with a compass on this vast literature. Finally, we go over the architecture of utilizing various object detection and table structure recognition methods to create an effective and efficient system, as well as a set of development trends to keep up with state-of-the-art algorithms and future research. We have also set up a public GitHub repository where we will be updating the most recent publications, open data, and source code. The GitHub repository is available at  \url{https://github.com/abdoelsayed2016/table-detection-structure-recognition}.

\end{abstract}


\end{frontmatter}


\section{Introduction}
\label{sec:sample1}
Textbooks, lists, formulae, graphs, tables, and other elements are common in documents. Most papers, in particular, contain several sorts of tables. Tables, as a significant part of papers, may convey more information in fewer words and allow readers to quickly explore, compare, and comprehend the content. Table detection and structure identification are crucial tasks in image analysis because they allow  retrieving vital information from tables in a digital format. Because of the document's type and the variety of document layouts, detecting and extracting images or document tables is tough. Researchers have previously used heuristic techniques to recognize tables or to break pages into many parts for table extraction. Few studies have focused on table structure recognition in documents following table detection.

The layout and content analysis of documents are used to detect tables. Tables come in a number of layouts and formats. As a result, creating a general method for table detection and table structure recognition is quite difficult. Table detection is regarded as a difficult subject in scientific circles. A large number of studies have been conducted in this sector, although the majority of them have limitations. Existing commercial and open-source document analysis algorithms, such as Tesseract, are unable to fully detect table areas from document images.
 \cite{hu2002evaluating}.

Machine learning and deep learning have been proven to be very effective in computer vision research. On computer vision tasks such as picture classification, object detection, object position estimation, learning, and so on, deep convolutional neural networks (types of feed-forward artificial neural systems) have outperformed alternative learning models. The effectiveness of Convolutional Neural Networks (CNNs) in object identification is based on their ability to learn substantial mid-level visual properties rather than the hand-crafted low-level representations that are often utilized in particular approaches to image categorization. The object is defined by its major characteristics, which include shape, size, color, texture, and other characteristics. To identify such an item, a picture must clearly reveal the object's presence and, moreover, its position \cite{dollar2014fast}. 

Object detection may thus be described as a method of locating real-world items in photographs. Detection is closely connected to categorization because it includes determining the existence and location of a certain item in an image. There are many items that may be identified in a picture, including automobiles, buildings, tables, human faces, and so on.
Deep learning approaches, such as deep neural networks, region-based convolutional neural networks, and deeply convolutional neural networks, can improve object identification precision, and efficacy.

In recent years, a variety of remarkable and creative strategies have been used to improve deep learning model detection accuracy and solve complex challenges encountered during the training and testing process of deep learning object recognition models. Modification of the activation function of deep CNNs \cite{yang2018modified}, Transfer learning \cite{li2019detection,masita2018pedestrian}, cancer diagnosis, detection \cite{hu2018deep,redmon2016you,ABDALLAH202279}, and classification\cite{fakoor2013using}, and medical question answers\cite{minaee2017automatic,abdallah2020automated}, as well as software engineering applications such as optimizing the time and schedule of software projects\cite{arpteg2018software,hamada2021neural}, Intrusion Detection in IoT \cite{mahmoud2022ae,xu2021improving} and handwritten recognition for various languages\cite{mahmoud2014khatt,nurseitov2021handwritten,toiganbayeva2022kohtd,fischer2013fast}., and inventive ways in the combined selection of the activation function and the optimization system for the proposed deep learning model are among these unique strategies.
Among the various variables and initiatives that have contributed to the rapid advancement of table detection algorithms, the development of deep convolutional neural networks and GPU computational capacity should be credited. Deep learning models are now widely used in many aspects of computer vision, including general table detection\cite{schreiber2017deepdesrt,traquair2019deep,gilani2017table,tran2015table,hao2016table}. Table structures, on the other hand, receive far less attention, and the table structure is typically characterized by the rows and columns of a table \cite{mao2003document,kara2020holistic,sarkar2020document}.

\begin{figure}[h!]
    
    \begin{subfigure}{0.50\columnwidth}
        \includegraphics[width=\textwidth,height=\textwidth]{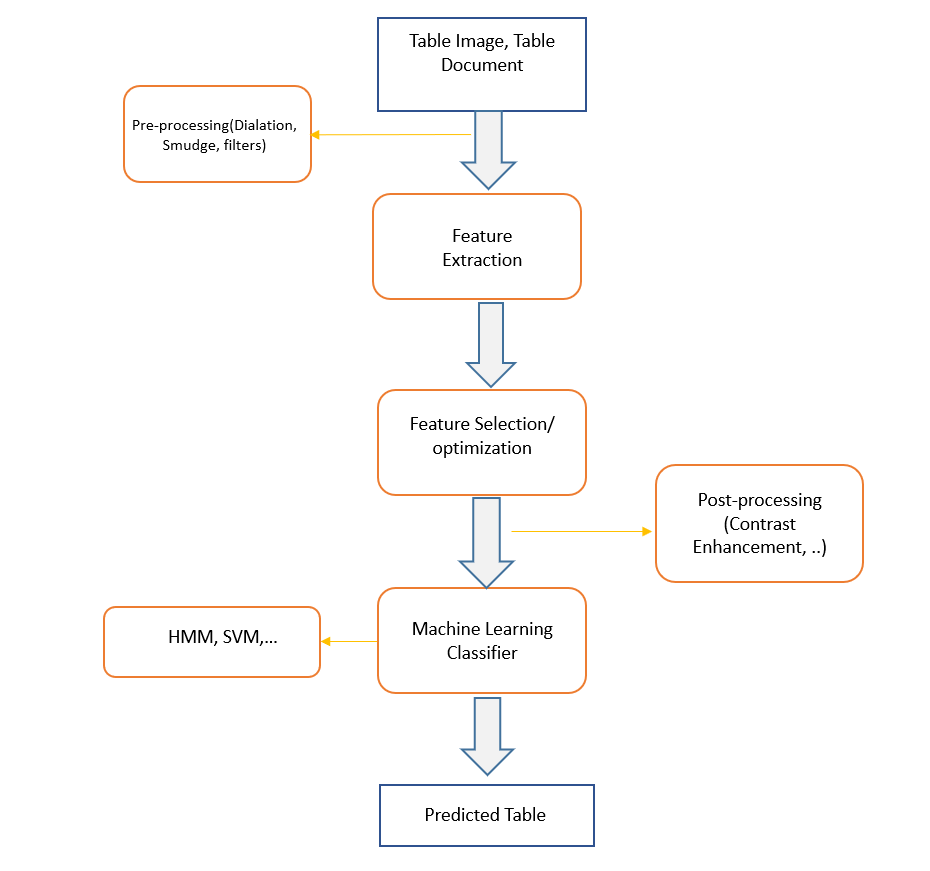}

        \caption{ Traditional Table Detection approaches }
    \end{subfigure}
    \begin{subfigure}{0.50\columnwidth}
        \includegraphics[width=\textwidth,height=\textwidth]{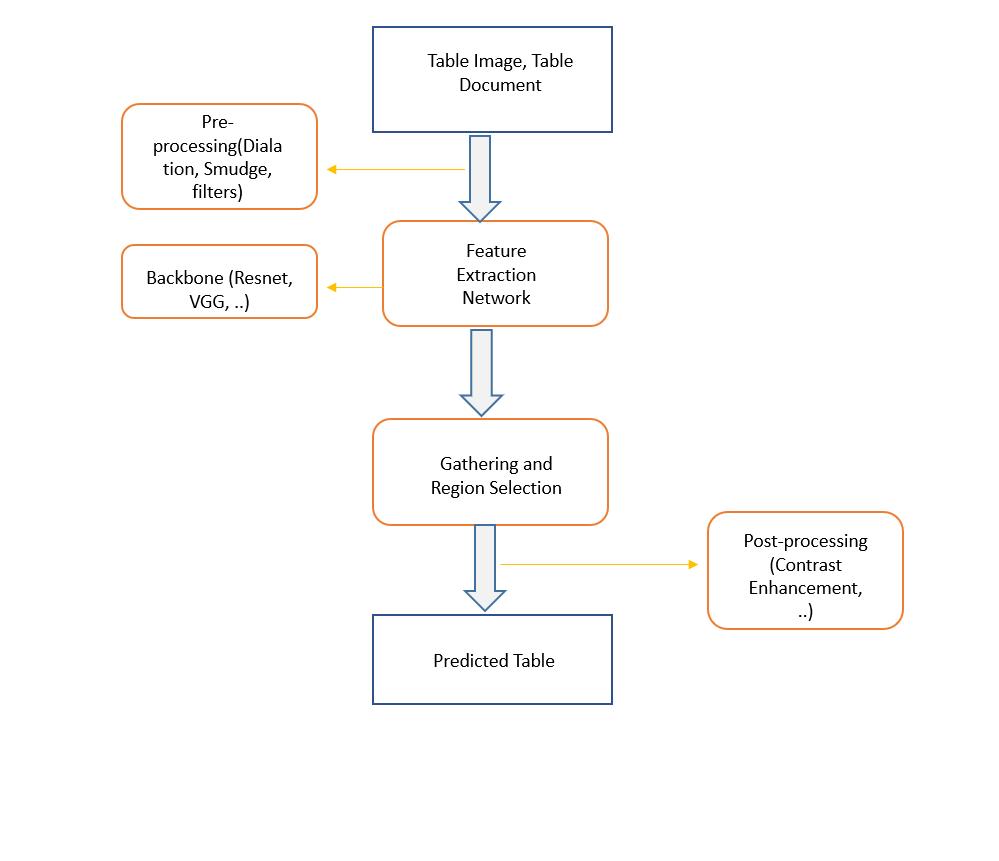}
        \caption{ Deep Learning approaches for Table Detection}
    \end{subfigure}
\caption{Table analysis pipeline comparison of conventional and deep learning methods. While convolutional networks are used in deep learning techniques, classical approaches primarily perform feature extraction through image processing techniques. Deep learning methods for interpreting tables are more generalizable and independent of data than conventional approaches.} 
\label{fig:Traditional and Deep Learning Approaches}
\end{figure}

Figure \ref{fig:Traditional and Deep Learning Approaches} shows a basic pipeline comparison of deep learning techniques and conventional approaches for the task of understanding tables. Traditional table recognition techniques either can't handle varied datasets well enough or need extra metadata from PDF files. Extensive pre- and post-processing were also used in the majority of early approaches to improve the effectiveness of conventional table recognition systems. However, deep learning algorithms retrieve features using neural networks, primarily convolutional neural networks \cite{traquair2019deep}, instead of manually created features. Object detection or segmentation networks then try to differentiate the tabular portion that is further broken down and recognized in a document image.

This survey examines deep learning-based table detection and classification architectures in depth. While current evaluations are comprehensive \cite{zanibbi2004survey,embley2006table}, the majority of them do not address recent advancements in the field.

The following are the paper's main contributions:
\begin{enumerate}
    \item We provide a brief history of Table Datasets and the differences between them.
    \item The paper examines important table detection methods, as well as the evolution of these methods over time.
    \item  We give a thorough analysis of table structure recognition in-depth. 
    \item We provide Table Classification methods and compare these methods. There was no study that provided a broad summary of these issues that we could identify.
    \item Experiments Result on some datasets of Table detection
\end{enumerate}

\subsection{Comparison with Previous Reviews}
\label{sec:sample1}
\begin{figure}[h!]
    
        \includegraphics[width=\textwidth]{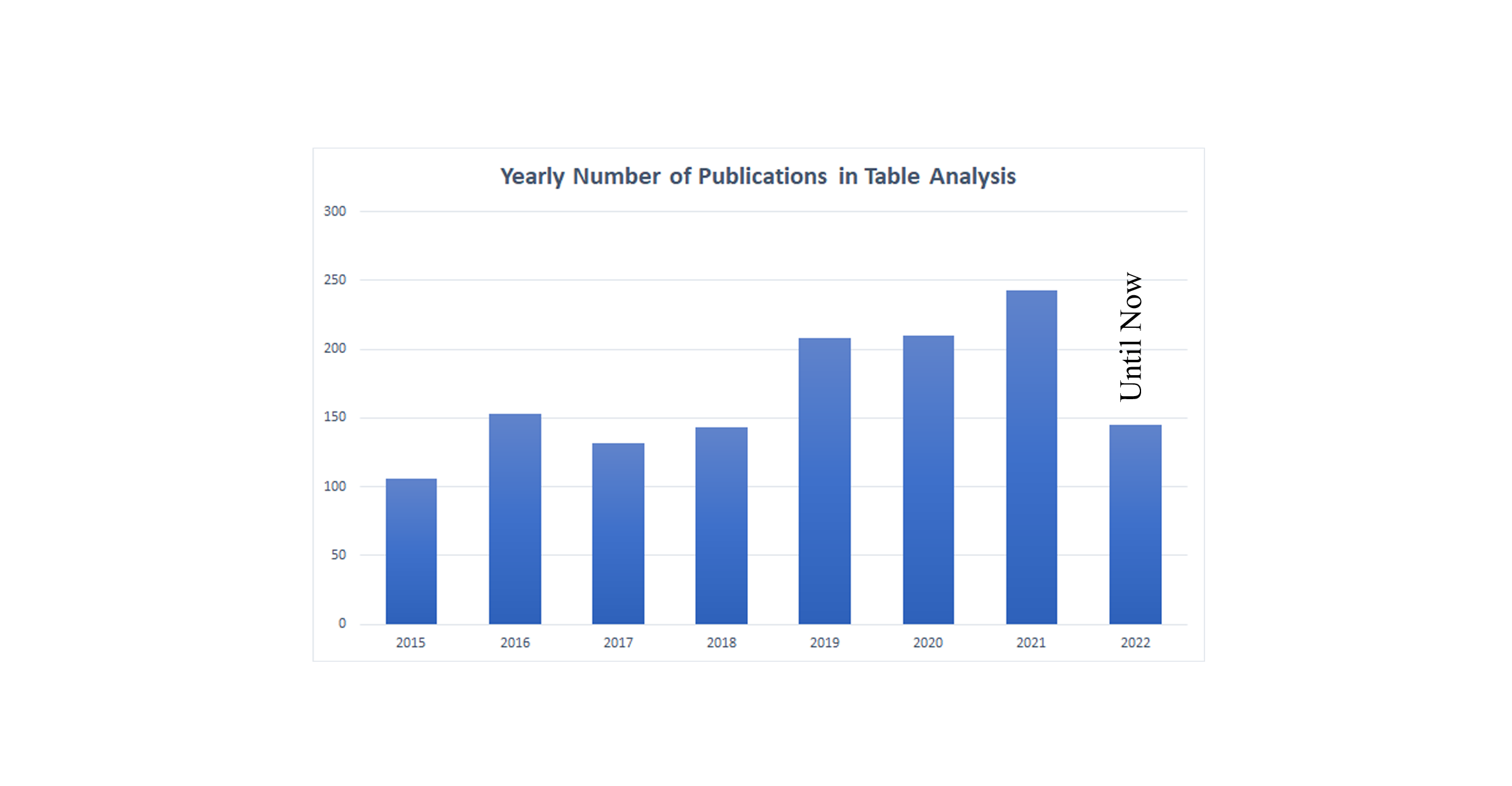}

\caption{shows an illustration of an expanding trend in the area of table analysis. This information was gathered by looking through the annual reports on table detection and table identification from the years 2015 to 2022, this analysis values were derived from Scopus. } 
\label{fig:Chart about Table analysis publications}
\end{figure}

For many years, the issue with table analysis has been widely acknowledged. Figure \ref{fig:Chart about Table analysis publications} shows the upward trend in publications during the previous eight years, this analysis values were derived from Scopus. There have been notable table detection and table classification surveys published. There are outstanding studies on the subject of table detection in these surveys \cite{zanibbi2004survey,embley2006table}. There have been few recent surveys that specifically address the subject of table detection and classification. B. Coüasnon \cite{couasnon2014recognition} released another review on table recognition and forms. The review gives a quick rundown of the most recent techniques at the time, S. Khusro \cite{khusro2015methods} released the most recent review on the identification and extraction of tables in PDF documents the following year, according to our knowledge. Deep learning enables computational models to learn fantastically complex, subtle, and abstract representations, resulting in significant advancements in a wide range of problems such as visual recognition, object detection, speech recognition, natural language processing, and medical image analysis. In contrast, despite the fact that various deep learning-based algorithms for table identification have been presented, we are unaware of any recent thorough survey. For further advancement in table detection, a detailed review and explanation of prior work are required, especially for researchers new to the topic.

\subsection{Scope}
\label{sec:sample1}
The quantity of studies on deep learning-based table detection is staggering. They are so numerous that any complete examination of the state of the art would be beyond the scope of any acceptable length paper. As a result, selection criteria must be established, and we have narrowed our attention to the best journal and conference articles.

The main goal of this paper is to provide a comprehensive survey of deep learning-based table detection and classification techniques, as well as some taxonomy, a high-level perspective, and organization, based on popular datasets, evaluation metrics, context modeling, and detection proposal methods. Our goal is for our classification to make it easier for readers to comprehend the similarities and differences across a wide range of tactics. The suggested taxonomy provides a framework for researchers to comprehend existing research and highlight open research problems for the future.

\section{Major Challenges}
\label{sec:Major Challenges}
\subsection{Object detection Challenges}
Developing a general-purpose algorithm that fulfills two competing criteria of high quality/accuracy and great efficiency is ideal for object detection. High-quality detection must accurately localize and recognize objects in images or video frames, allowing for distinction of a wide range of object categories in the real world and localization and recognition of object instances from the same category, despite intra-class appearance variations, for high robustness. High efficiency necessitates that the full detection process is completed in real-time while maintaining reasonable memory and storage requirements.

\subsection{Table detection Challenges}
Although a trained segmentation model can accurately locate tables, conventional machine learning techniques have flaws in the structural identification of tables. A major issue is a large number of things in such a little space. As a result, the network misses out on critical visual cues that may aid in the detection and recognition of tables \cite{schreiber2017deepdesrt}. As physical rules are available, intersections of horizontal and vertical lines are computed to recognize table formations. The Hough transform is a prominent approach in computer vision that aids in the detection of lines in document scans \cite{szeliski2010computer}.  Length, rotation, and average darkness of a line are utilized to filter out false positives and determine if the line is, in fact, a table line \cite{lee2017line}. The intersections of the remaining horizontal and vertical lines are computed after the Hough lines have been filtered. Table cells are created based on the crossings.

\section{A Quick Overview of Deep Learning}
\label{sec:Deep Learning}
From image classification and video processing to speech recognition and natural language understanding, deep learning has transformed a wide range of machine learning activities. Given the incredible rate of change\cite{liu2020deep}, there is a plethora of current survey studies on deep learning \cite{bengio2013representation,lecun2015deep,goodfellow2016deep,litjens2017survey,zhu2017deep,gu2018recent,pouyanfar2018survey,young2018recent,zhou2020graph,zhang2018deep,wu2020comprehensive} , medical image analysis applications \cite{litjens2017survey}, natural language processing \cite{young2018recent}, and speech recognition systems \cite{zhang2018deep}.

Convolutional neural networks (CNNs), the most common deep learning model, can use the fundamental properties of actual signals: translation invariance, local connection, and compositional hierarchies. A typical CNN comprises a hierarchical structure and numerous layers for learning data representations at different levels of abstraction \cite{lecun2015deep}. We start with a convolution
\begin{equation}
\begin{split}
x^{l-1} * w^{l}
\end{split}
\end{equation}
between a feature map from the previous layer l-1 and an input feature map $x^{l-1}$ , convolved using a 2D convolutional kernel (or filter or weights) $w^l$. This convolution is seen as a series of layers that have been subjected to a nonlinear process $\sigma$, such
that 
\begin{equation}
\begin{split}
x^{l}_{j} = \sigma \Bigg(\sum_{i=1}^{N^{l-1}} x^{l-1}_{i} * w^{l}_{i,j} + b^{l}_{j} \Bigg)
\end{split}
\end{equation}
with a bias term $b^l_j$ and a convolution between the $N^{l-1}$ input feature maps $x^{l-1}_i$  and the matching kernel $w^{l}_{i,j}$ . For each element, the element-wise nonlinear function $\sigma(.)$  is commonly a rectified linear unit (ReLU) for each element,

\begin{equation}
\begin{split}
\sigma(x) = max \{x,0\}
\end{split}
\end{equation}

Finally, pooling is the process of downsampling and upsampling feature maps. Deep convolution neural networks(DCNNs) are CNNs with a large number of layers, often known as "deep" networks . A CNN's most basic layers consist of a series of feature maps, each of which operates as a neuron. A set of weights $w_{i,j}$ connects each neuron in a convolutional layer to feature maps from the preceding layer (essentially a set of 2D filters).
Whereas convolutional and pooling layers make up the early CNN layers, the subsequent layers are usually completely connected. The input picture is repeatedly convolved from earlier to later layers, and the receptive field or region of support grows with each layer. In general, the first CNN layers extract low-level characteristics (such as edges), whereas subsequent layers extract more generic features of increasing complexity. \cite{bengio2013representation,zeiler2014visualizing,oquab2014learning,lecun2015deep}.

DCNNs have a hierarchical structure that allows them to learn data representations at numerous levels of abstraction, the ability to learn highly complicated functions, and the ability to learn feature representations directly and automatically from data with minimum domain expertise. The availability of huge size labeled datasets and GPUs with extremely high computational capabilities is what has made DCNNs so successful.

Despite the enormous achievements, there are still acknowledged flaws. There is a critical need for labeled training data as well as expensive computational resources, and selecting proper learning parameters and network designs still requires substantial expertise and experience. Trained networks are difficult to comprehend, lack resistance to degradations and many DCNNs have been proven to be vulnerable to assaults \cite{goodfellow2016deep}, all of which restrict their applicability in real-world applications.

\section{Datasets and Evaluation Metrics}

\subsection{Datasets}
This section will describe datasets that are available and have been most commonly used for table detection, table structure recognition, and classification tasks.

\subsubsection{ICDAR 2013}
\begin{figure}[h!]
    \begin{subfigure}{0.50\columnwidth}
        \includegraphics[width=\textwidth]{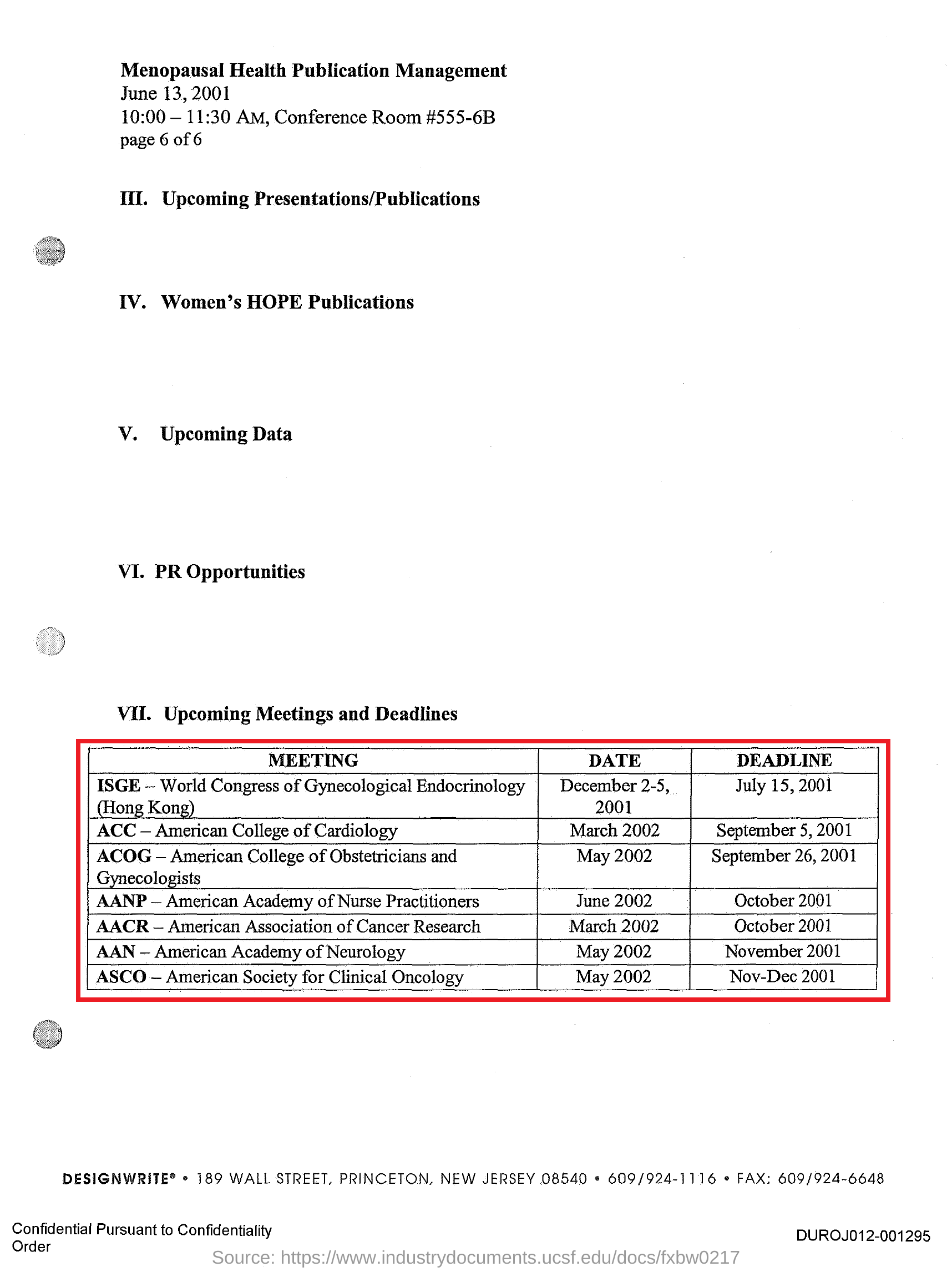}
        
        \caption{ }
    \end{subfigure}
    \begin{subfigure}{0.50\columnwidth}
        \includegraphics[width=\textwidth]{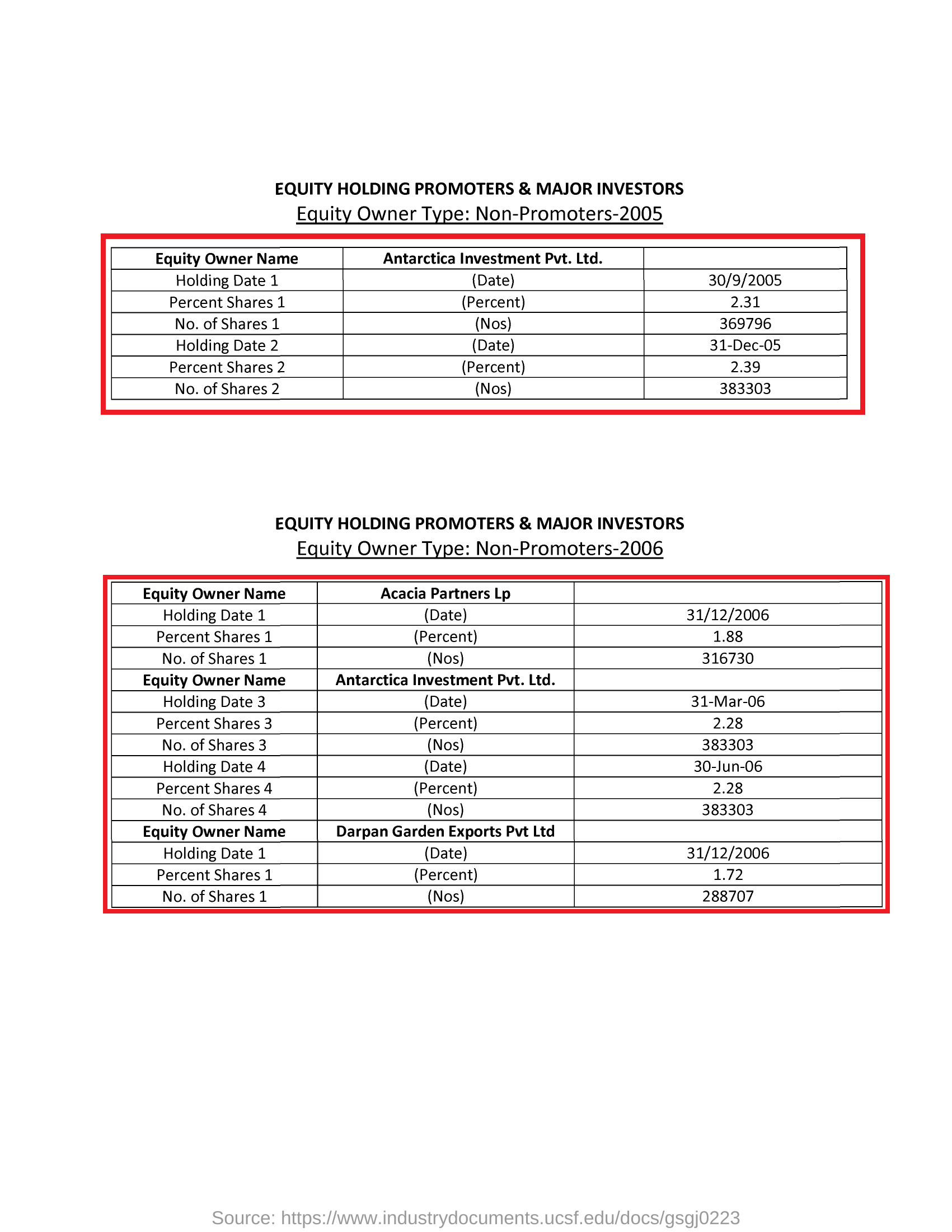}
        \caption{}
    \end{subfigure}

\caption{Examples of images in ICDAR 2013} 
\label{fig:ICDAR2013}
\end{figure}

ICDAR2013 dataset\cite{gobel2013icdar} referred to in served as the competition's official practice dataset. Rather than focusing on a specific subset of documents, authors have always intended to evaluate systems as broadly as possible, and this dataset, as well as the actual competition dataset, were generated by systematically collecting PDFs from a Google search in order to make the selection as objective as possible. they are limited to two governmental sources with the additional search terms site:europa.eu and site:*.gov in order to obtain documents whose publications are known to be in the public domain. The ICDAR2013 dataset contains 150 tables, 75 of which are in 27 EU excerpts and 75 of which are in 40 US Government excerpts. Table regions are rectangular areas of a page whose coordinates define them. Multiple regions can be included in the same table because a table can span multiple pages. Table detection or location and table structure recognition are the two sub-tasks of ICDAR2013. The task of table structure recognition compares methods for determining table cell structure given precise location information. Figure \ref{fig:ICDAR2013} presents a few examples from this dataset.

\subsubsection{ICDAR 2017 POD}
\begin{figure}[h!]
    
    \begin{subfigure}{0.50\columnwidth}
        \includegraphics[width=\textwidth]{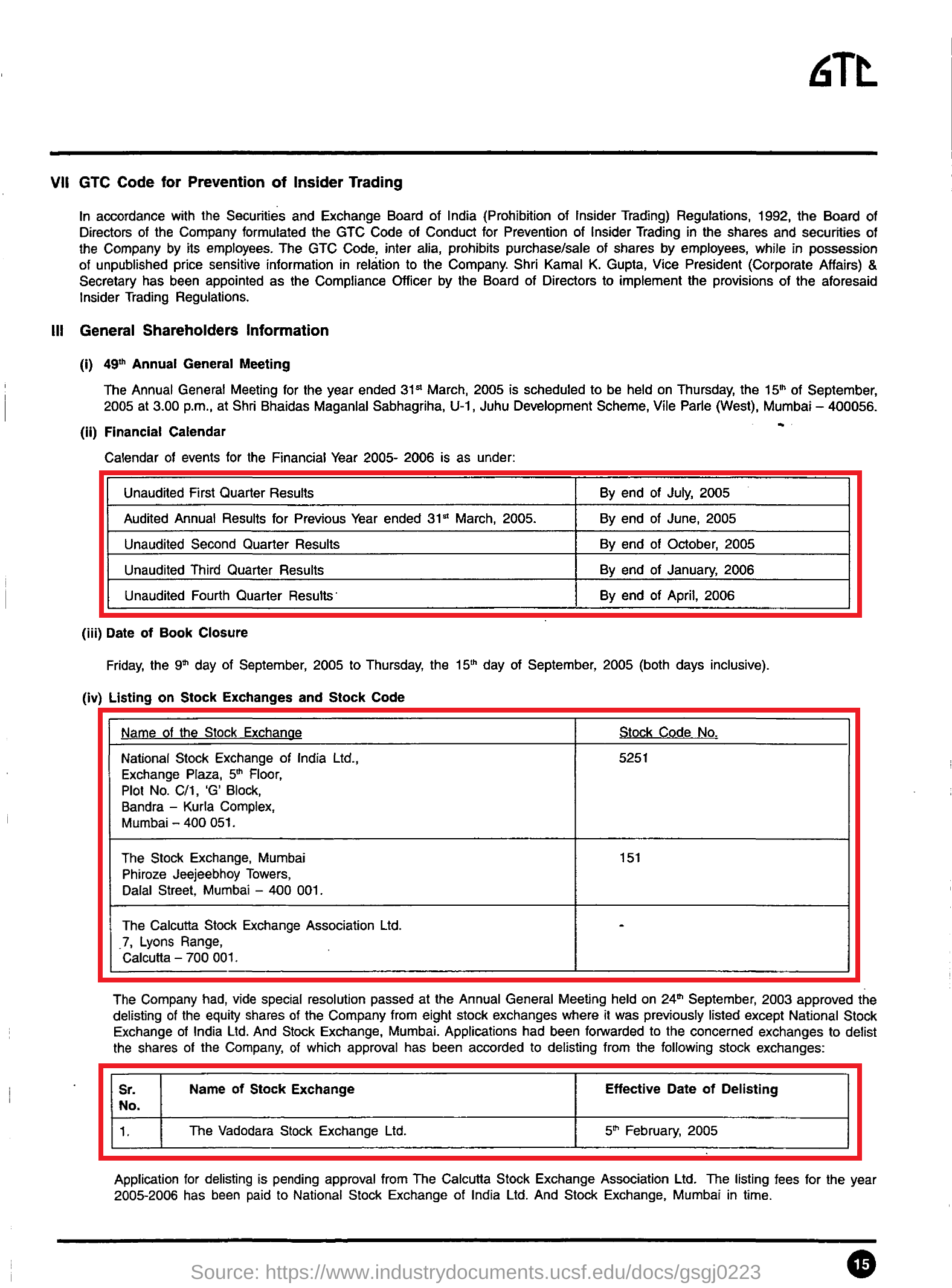}
        
        \caption{}
    \end{subfigure}
    \begin{subfigure}{0.50\columnwidth}
        \includegraphics[width=\textwidth]{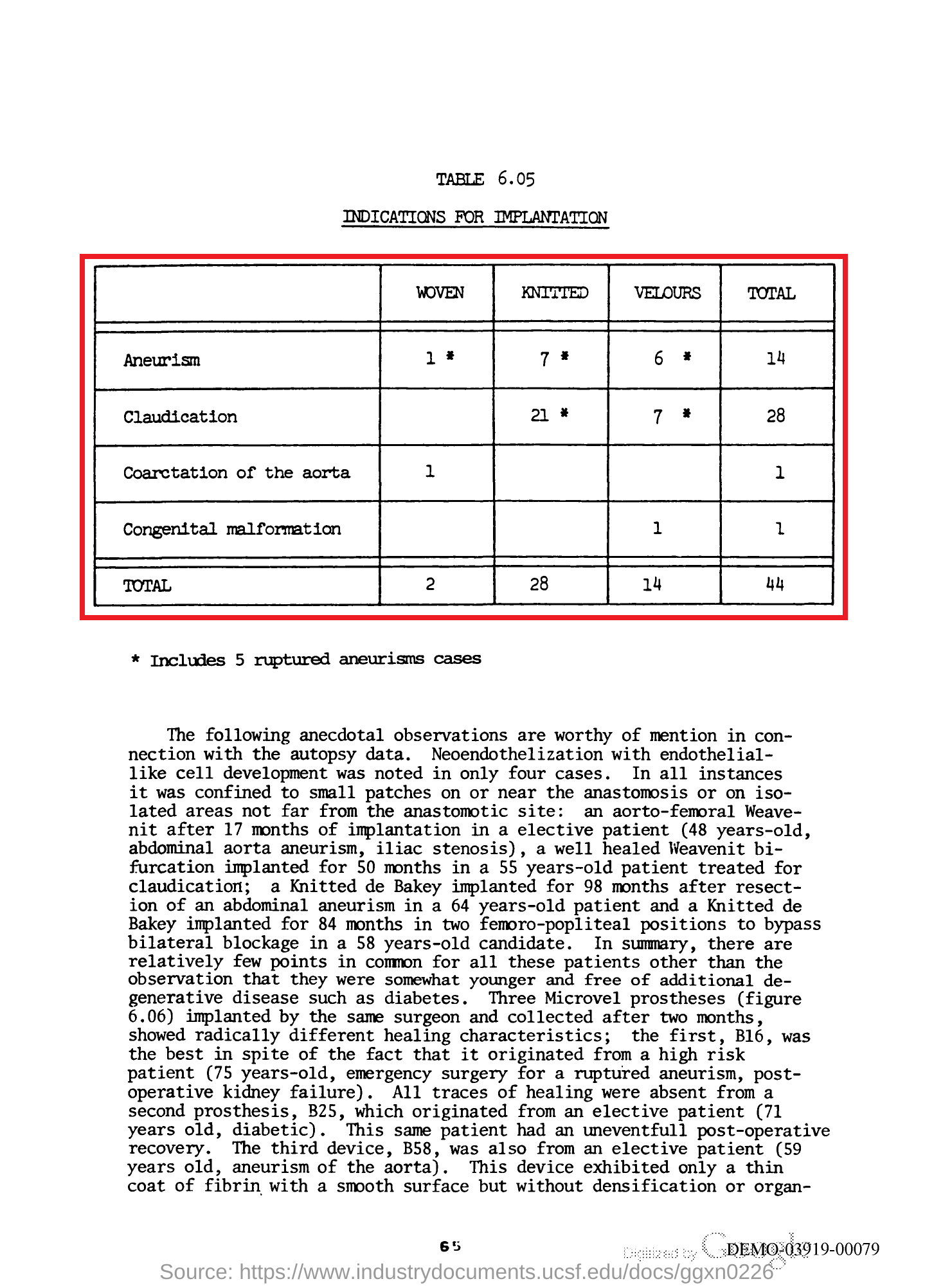}
        \caption{ }
    \end{subfigure}
\caption{Examples of images in ICDAR 2017} 
\label{fig:ICDAR2017}
\end{figure}

Dataset \cite{gao2017icdar2017} was published for the ICDAR2017  Page Object Detection (POD) competition. This dataset is frequently used to test different methods of table detection. Compared to the ICDAR2013 table dataset, this dataset is significantly larger. There are 2417 total images in it, including figures, tables, and formulae. The dataset is commonly split into 1600 photos which have 731 tabular areas for training, and the remaining 817 images which have 350 tabular regions for testing. Figure \ref{fig:ICDAR2017} illustrates two examples of this dataset.

\subsubsection{ICDAR2019}
\begin{figure}[h!]
    
    \begin{subfigure}{0.50\columnwidth}
        \includegraphics[width=\textwidth]{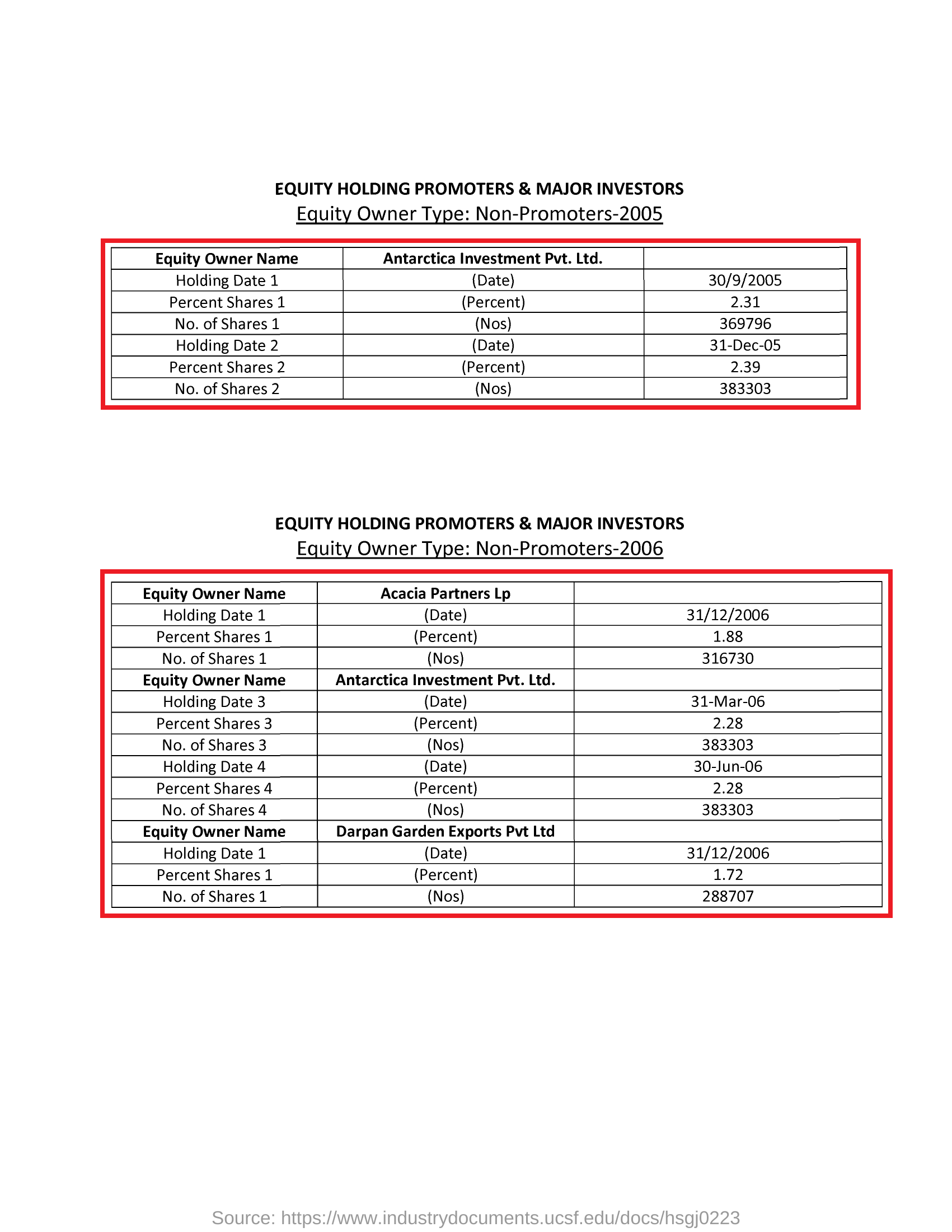}

        \caption{}
    \end{subfigure}
    \begin{subfigure}{0.50\columnwidth}
        \includegraphics[width=\textwidth]{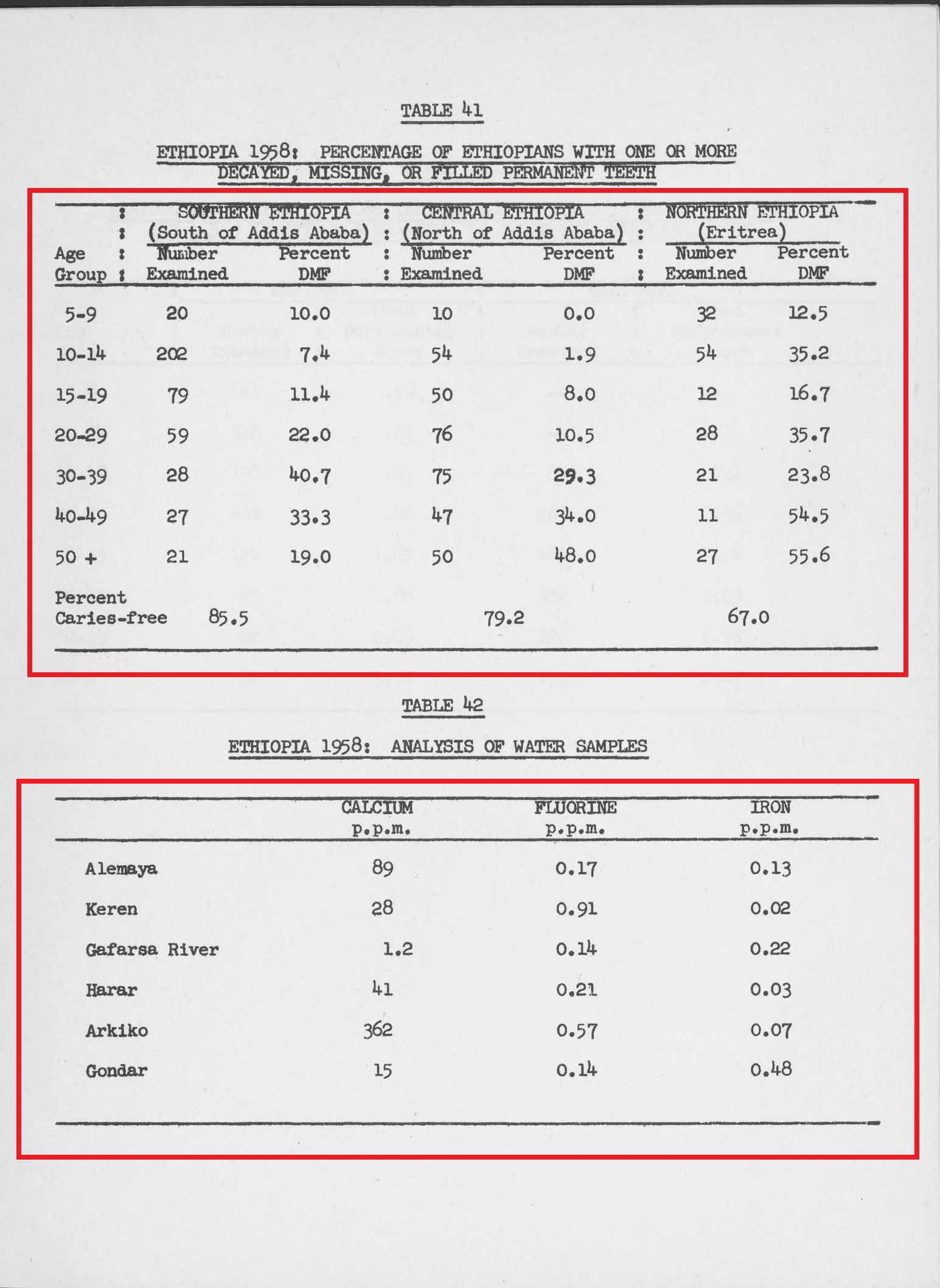}
        \caption{ }
    \end{subfigure}
\caption{Examples of images in ICDAR 2019} 
\label{fig:ICDAR2019}
\end{figure}

ICDAR2019 \cite{gao2019icdar} proposed a dataset for table detection (TRACK A) and table recognition (TRACK B). The dataset is divided into two types, historical and modern datasets. It contains 1600 images for training and 839 images for testing. The historical type contains 1200 images in tracks A and B for training and 499 images for testing. The modern type contains 600 images in tracks A and B for training and 340 images for testing. Document images containing one or more tables are provided for TRACK A. TRACK B has two sub-tracks: the first (B.1) provides the table region, and only the table structure recognition is required. The second sub-track (B.2) contains no prior knowledge. That is, both table region detection and table structure recognition must be performed. For the annotation of the dataset, a similar notation was derived from the ICDAR 2013 \cite{gobel2013icdar} Table Competition format, and the structures were stored in a single XML file. Each table element corresponds to a table with a single Coords element with a points attribute indicating the coordinates of the bounding polygon with N vertices. Each table element contains a list of cell elements as well. The attributes start-row, start-col, end-row, and end-col denote the position of each cell element in the table. The cell element's Coords denote the coordinates of the bounding polygon of this cell box, and the content is the text within this cell. Figure \ref{fig:ICDAR2019} presents a few examples from this dataset.

\subsubsection{TabStructDB}
\begin{figure}[h!]
    
    \begin{subfigure}{0.50\columnwidth}
        \includegraphics[width=\textwidth]{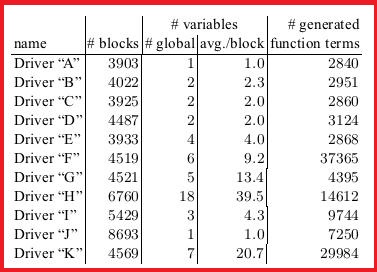}

        \caption{}
    \end{subfigure}
    \begin{subfigure}{0.50\columnwidth}
        \includegraphics[width=\textwidth]{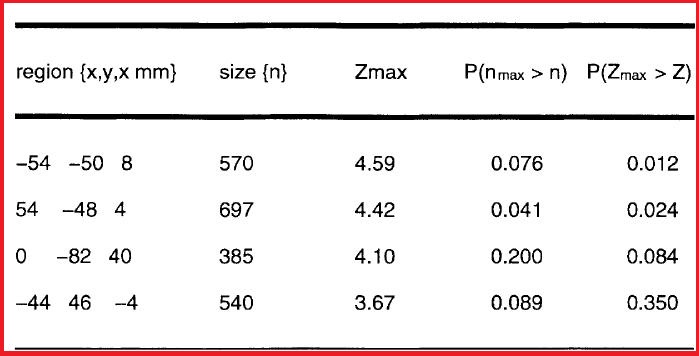}
        \caption{ }
    \end{subfigure}
\caption{Examples of images in TabStructDB 
} 
\label{fig:TabStructDB}  
\end{figure}

TabStructDB is a different publicly available image-based table structure recognition dataset that was promoted by SA Siddiqui \cite{siddiqui2019deeptabstr}. The well-known ICDAR 2017 page object detection dataset, which contains pictures annotated with structural details, was used to curate this dataset. Figure \ref{fig:TabStructDB} illustrates two examples of this dataset.

\subsubsection{TABLE2LATEX-450K}
\begin{figure}[h!]
    
    \begin{subfigure}{0.50\columnwidth}
        \includegraphics[width=\textwidth]{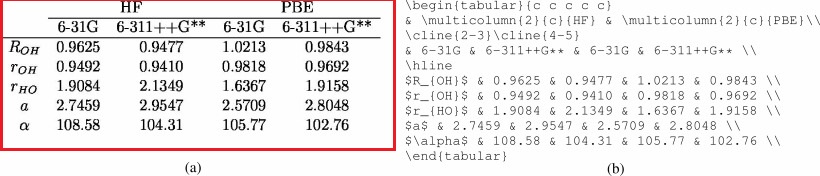}

        \caption{}
    \end{subfigure}
    \begin{subfigure}{0.50\columnwidth}
        \includegraphics[width=\textwidth]{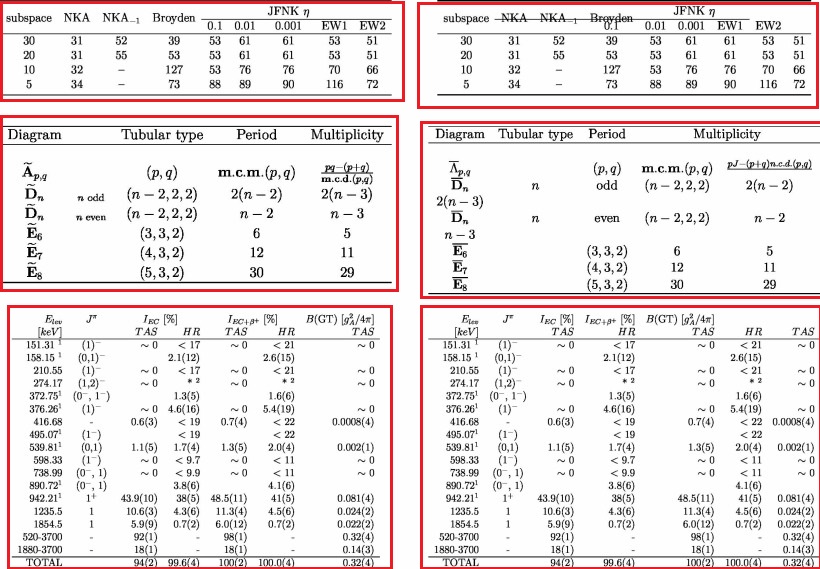}
        \caption{ }
    \end{subfigure}
\caption{Examples of images in TABLE2LATEX-450K 
} 
\label{fig:TABLE2LATEX-450K}  
\end{figure}

TABLE2LATEX-450K \cite{deng2019challenges} is another sizable dataset that was released at the most recent ICDAR conference. The dataset includes 450,000 annotated tables and the associated pictures. This enormous dataset was created by crawling through all of the LaTeX source documents and ArXiv publications from 1991 to 2016. The high-quality labeled dataset is obtained via source code extraction and further refining. Figure \ref{fig:TABLE2LATEX-450K} presents a few examples from this dataset.

\subsubsection{RVL-CDIP (SUBSET)}

\begin{figure}[h!]
    
    \begin{subfigure}{0.50\columnwidth}
        \includegraphics[width=\textwidth]{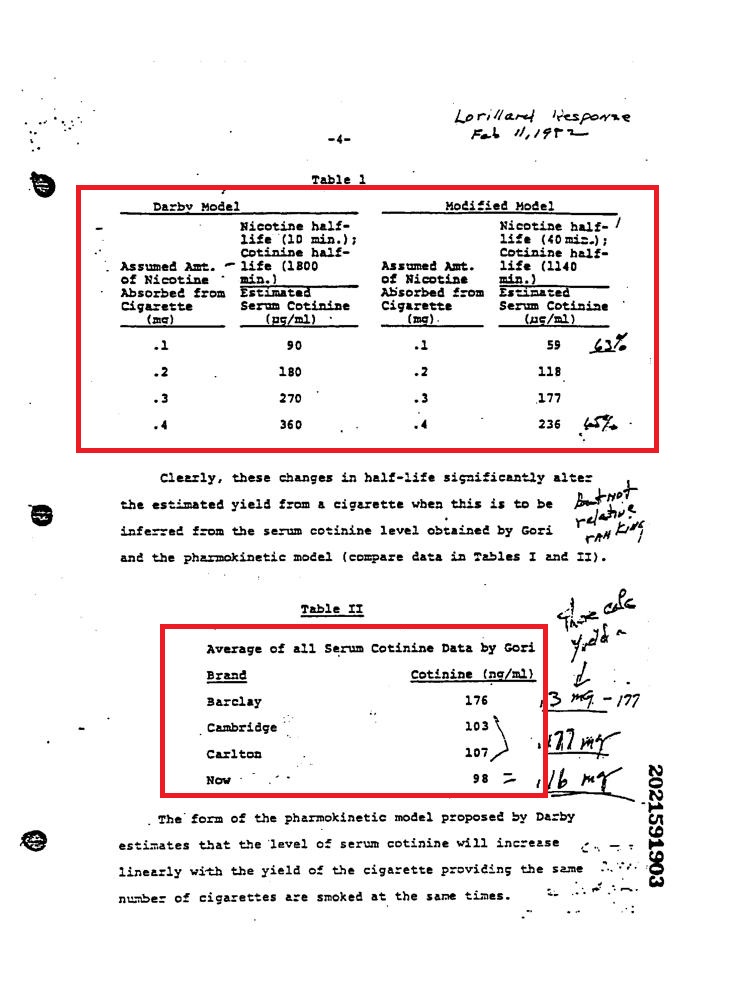}

        \caption{}
    \end{subfigure}
    \begin{subfigure}{0.50\columnwidth}
        \includegraphics[width=\textwidth]{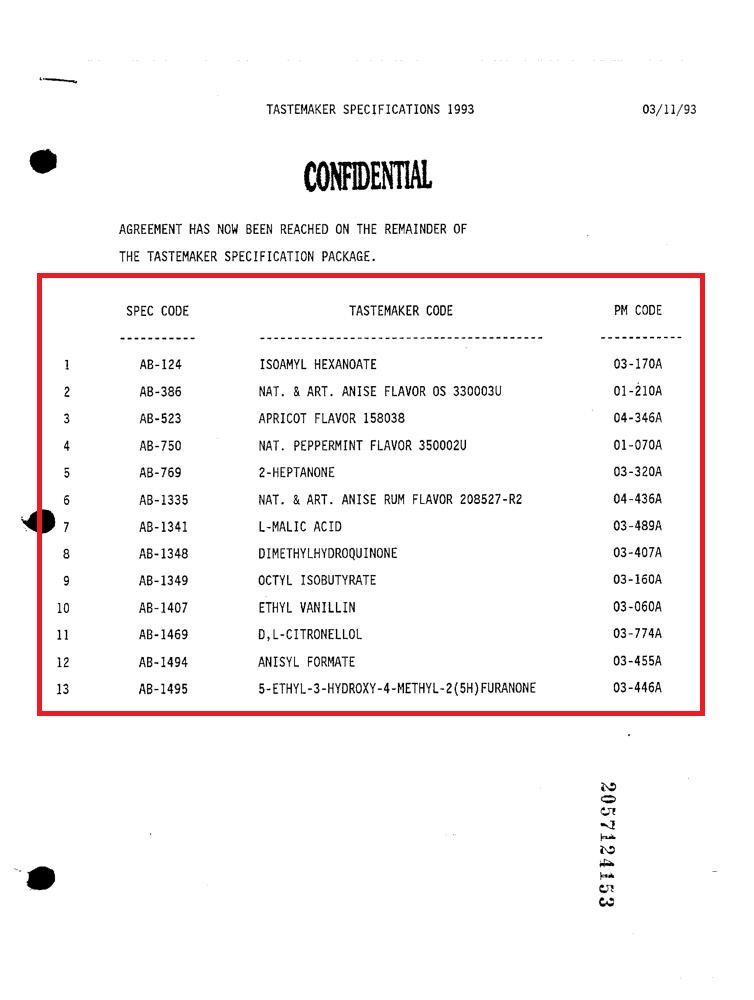}
        \caption{ }
    \end{subfigure}
\caption{Examples of images in RVL-CDIP (SUBSET)} 
\label{fig:RVL-CDIP}
\end{figure}

A well-known dataset in the world of document analysis is the RVL-CDIP (Ryerson Vision Lab Complex Document Information Processing) \cite{harley2015evaluation}. 400,000 photos are included, evenly spread over 16 classifications. For the purpose of detecting tables, P Riba \cite{riba2019table} makes subset dataset by annotating the 518 invoices in the RVL-CDIP dataset. The dataset has been made available to the general public. For testing table identification methods especially created for invoice document pictures, this subset of the real RVL-CDIP dataset \cite{harley2015evaluation} is a significant contribution. Figure \ref{fig:RVL-CDIP} presents a few examples from this dataset.

\subsubsection{IIIT-AR-13K}
\begin{figure}[h!]
    
    \begin{subfigure}{0.50\columnwidth}
        \includegraphics[width=\textwidth]{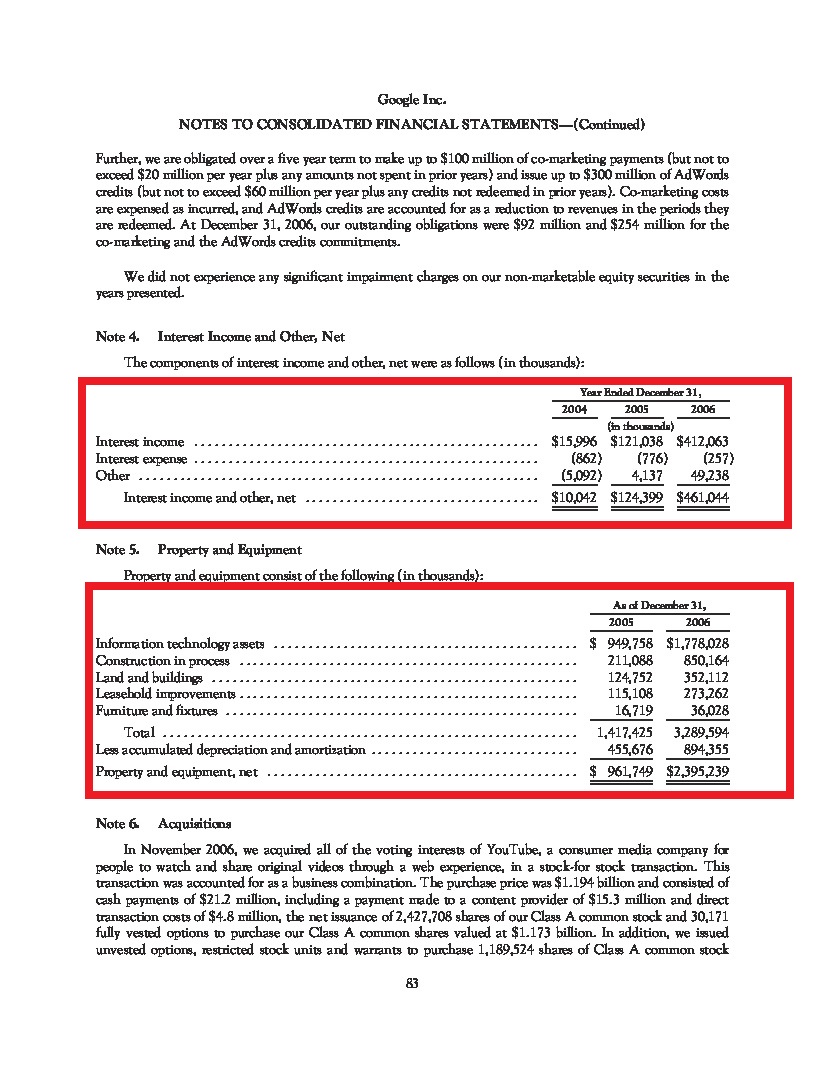}

        \caption{}
    \end{subfigure}
    \begin{subfigure}{0.50\columnwidth}
        \includegraphics[width=\textwidth]{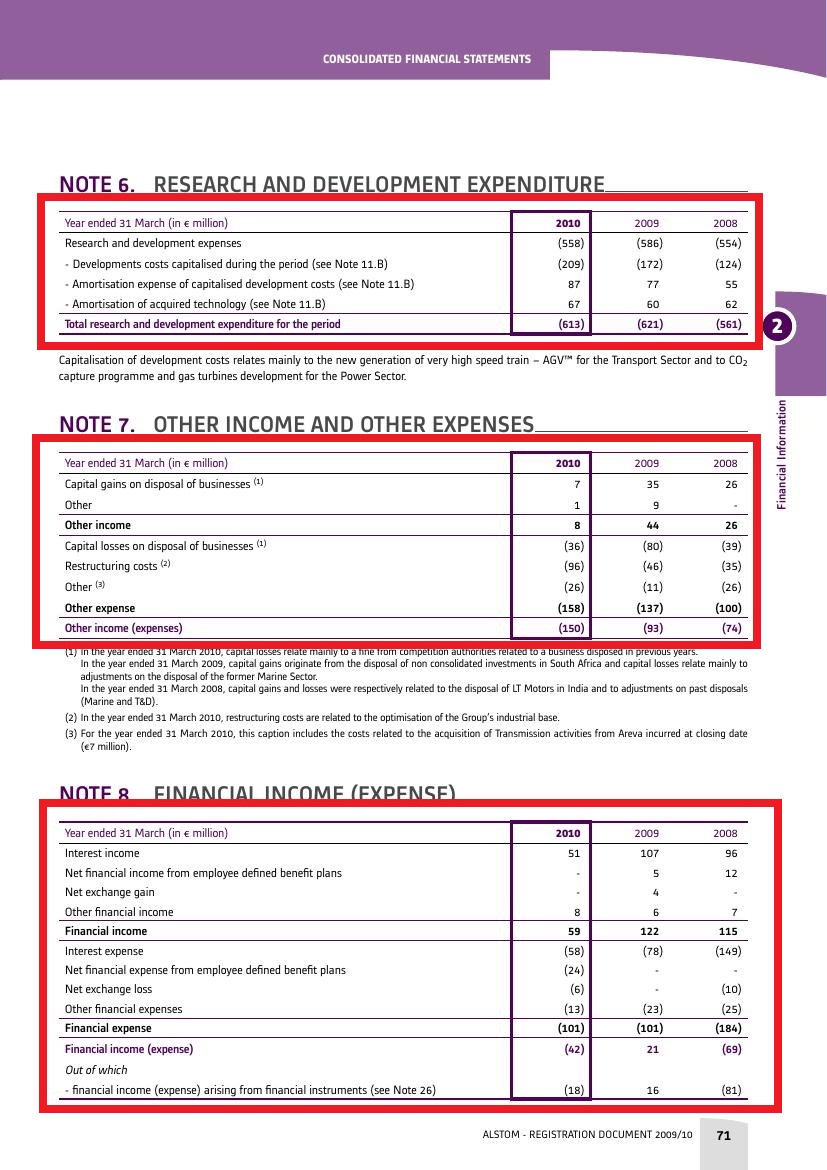}
        \caption{ }
    \end{subfigure}
\caption{Examples of images in IIIT-AR-13K} 
\label{fig:IIIT-AR-13K}
\end{figure}

IIT-AR-13K is a brand-new dataset that is introduced by A Mondal \cite{mondal2020iiit}. The yearly reports that were publicly available and written in English and other languages were collected  to create this dataset. The biggest manually annotated dataset for the problem of graphical page object recognition, according to the authors, has been released. Annotations for figures, natural imagery, logos, and signatures are included in the dataset in addition to the tables. For numerous tasks of page object detection, the authors have included the train, validation, and test splits. In order to train for table detection, 11,000 samples are employed, while 2000 and 3000 samples are allotted for validation and testing, respectively. Figure \ref{fig:IIIT-AR-13K} illustrates two examples of this dataset.

\subsubsection{CamCap}
\begin{figure}[h!]
    
    \begin{subfigure}{0.50\columnwidth}
        \includegraphics[width=\textwidth]{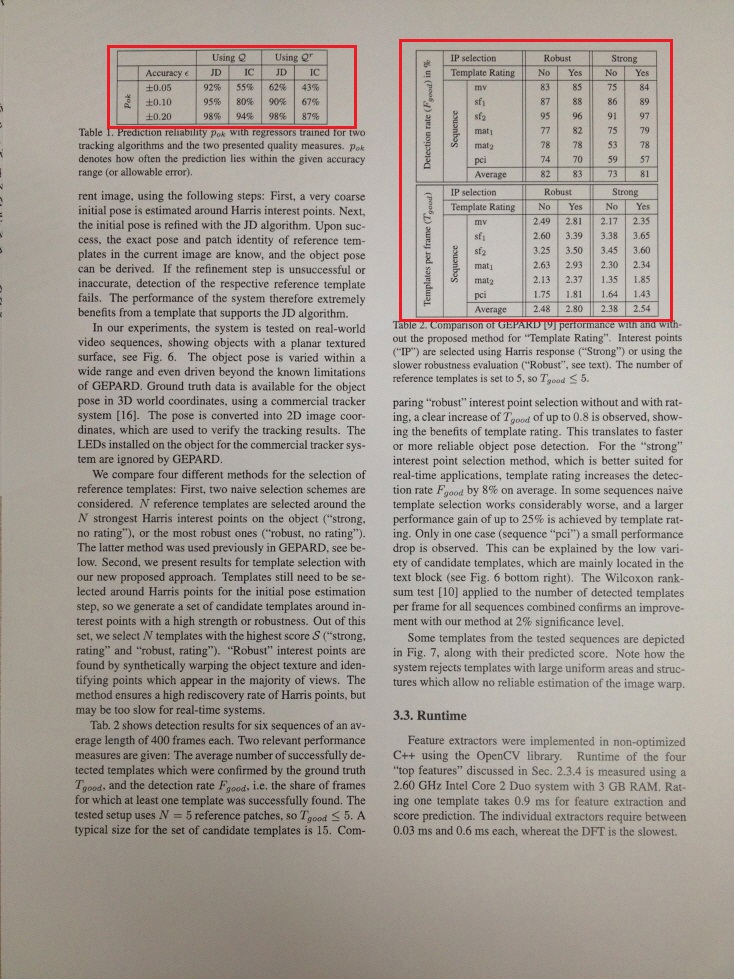}

        \caption{}
    \end{subfigure}
    \begin{subfigure}{0.50\columnwidth}
        \includegraphics[width=\textwidth]{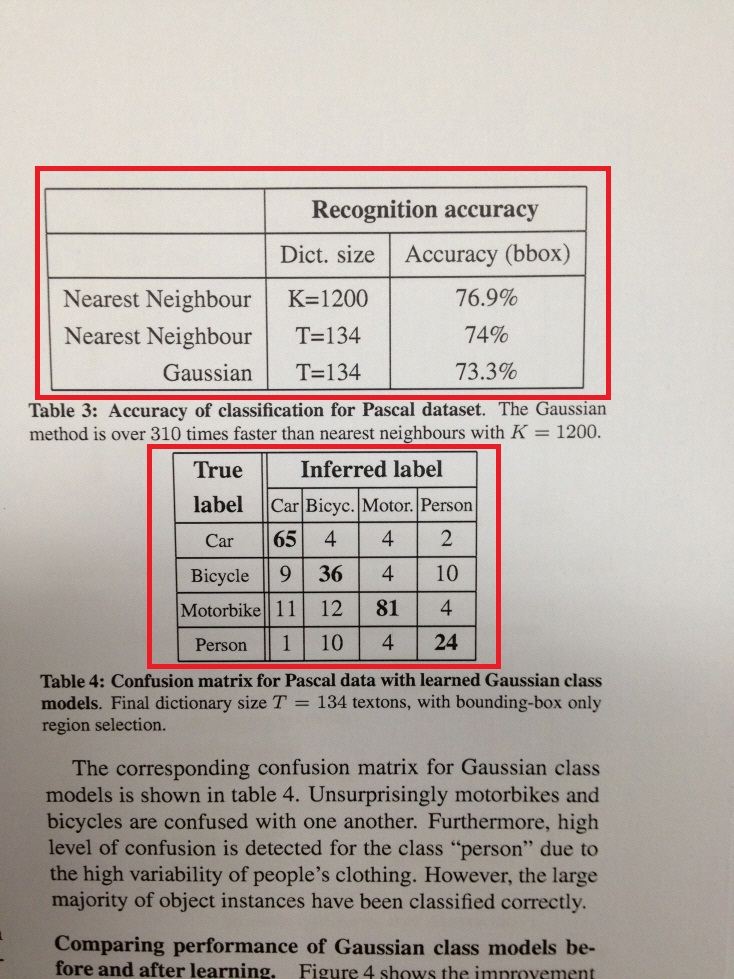}
        \caption{ }
    \end{subfigure}
\caption{Examples of images in CamCap} 
\label{fig:CamCap}
\end{figure}

CamCap is a collection of camera-captured photos suggested by W Seo \cite{seo2015junction}. Only 85 photos are present (38 tables on curved surfaces having 1295 cells and 47 tables on planar surfaces consisting of 1162 cells). For the sake of detecting tables and identifying their structures, the suggested dataset is accessible to the general public. This dataset is a significant addition to evaluating the reliability of table identification techniques on camera-captured document pictures. Figure \ref{fig:CamCap} illustrates two examples of this dataset.

\subsubsection{UNLV Table}
\begin{figure}[h!]
    
    \begin{subfigure}{0.50\columnwidth}
        \includegraphics[width=\textwidth,height=\textwidth]{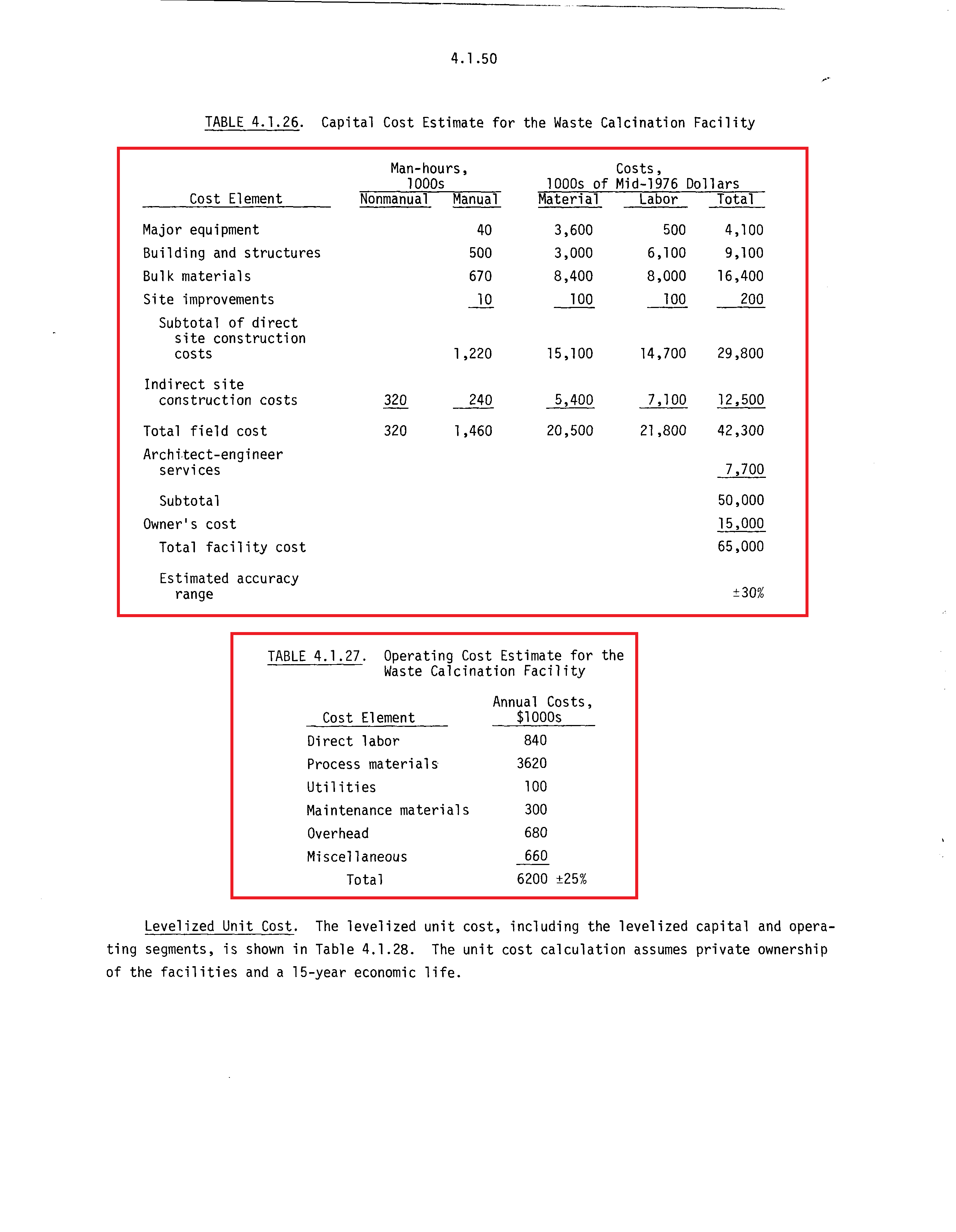}

        \caption{}
    \end{subfigure}
    \begin{subfigure}{0.50\columnwidth}
        \includegraphics[width=\textwidth,height=\textwidth]{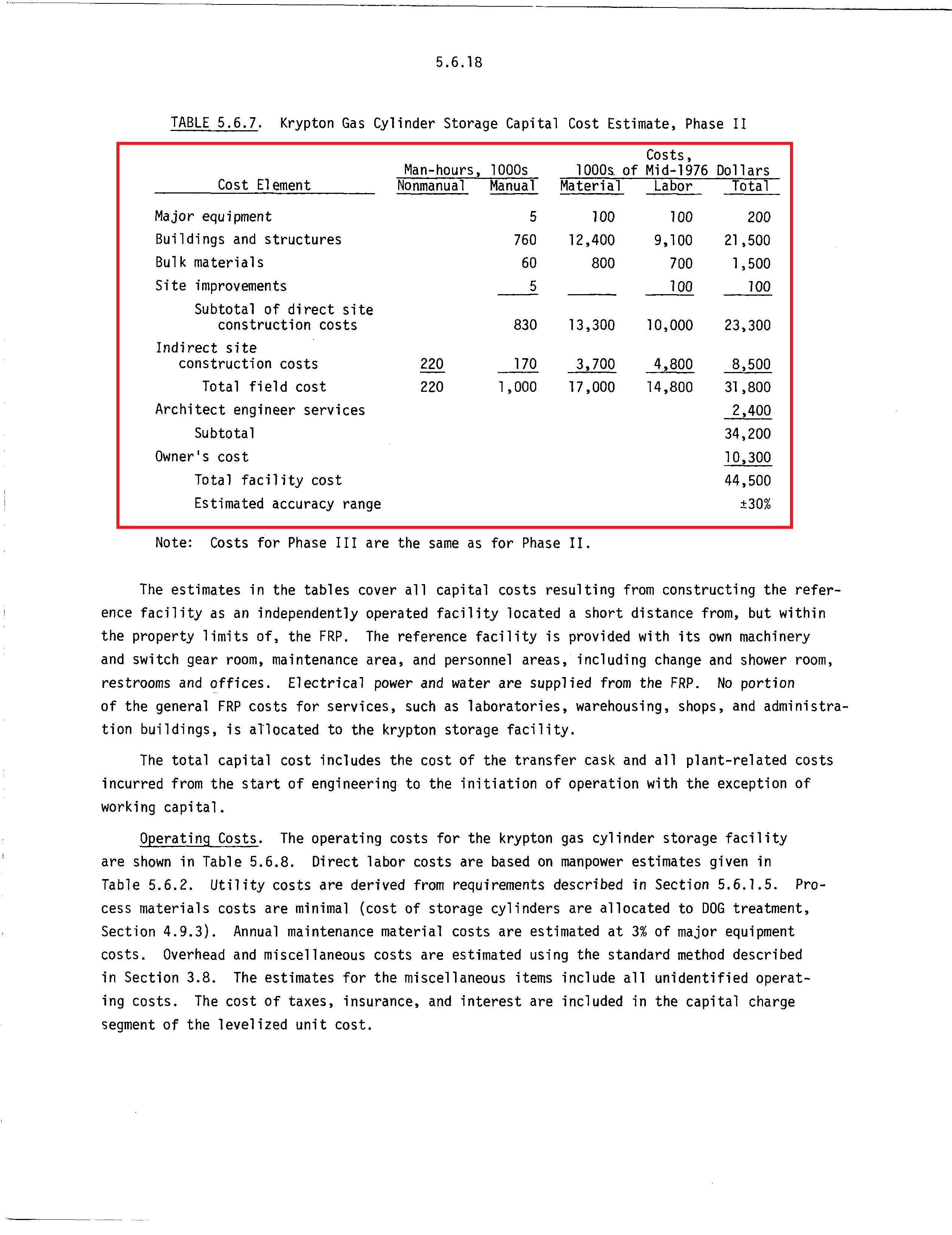}
        \caption{ }
    \end{subfigure}
\caption{Examples of images in UNLV} 
\label{fig:UNLV}
\end{figure}

The UNLV Table dataset\cite{shahab2010open} includes 2889 pages of scanned document images gathered from various sources (magazines, newspapers, business letters, annual reports, etc). Scanned images are available in bitonal, grayscale, and fax formats, with resolutions ranging from 200 to 300 DPI. There is ground truth data in addition to the original dataset, which contains manually marked zones; zone types are provided in text format. Figure \ref{fig:UNLV} illustrates two examples of this dataset.

\subsubsection{UW-3 Table}
\begin{figure}[h!]
    
    \begin{subfigure}{0.50\columnwidth}
        \includegraphics[width=\textwidth,height=\textwidth]{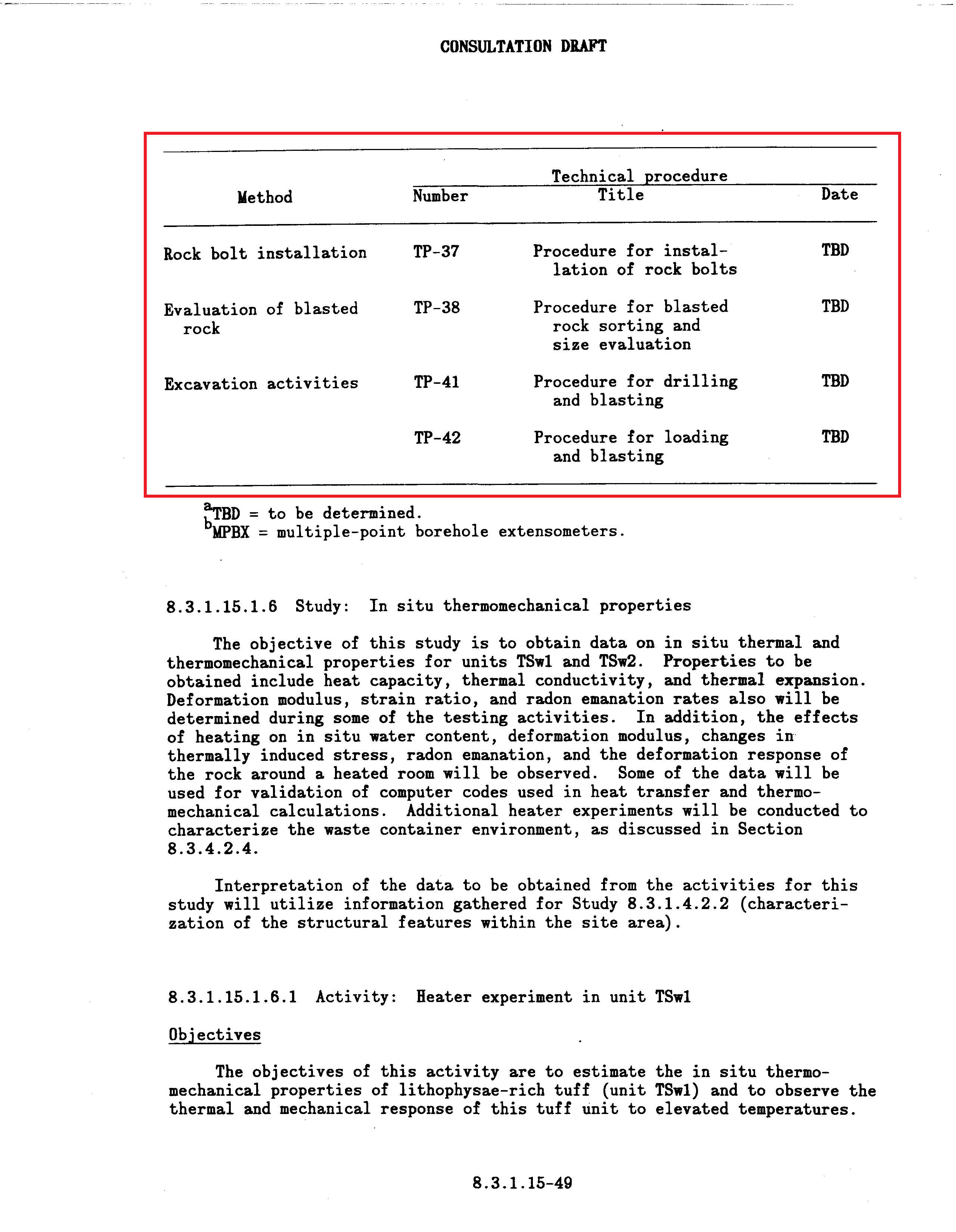}

        \caption{}
    \end{subfigure}
    \begin{subfigure}{0.50\columnwidth}
        \includegraphics[width=\textwidth,height=\textwidth]{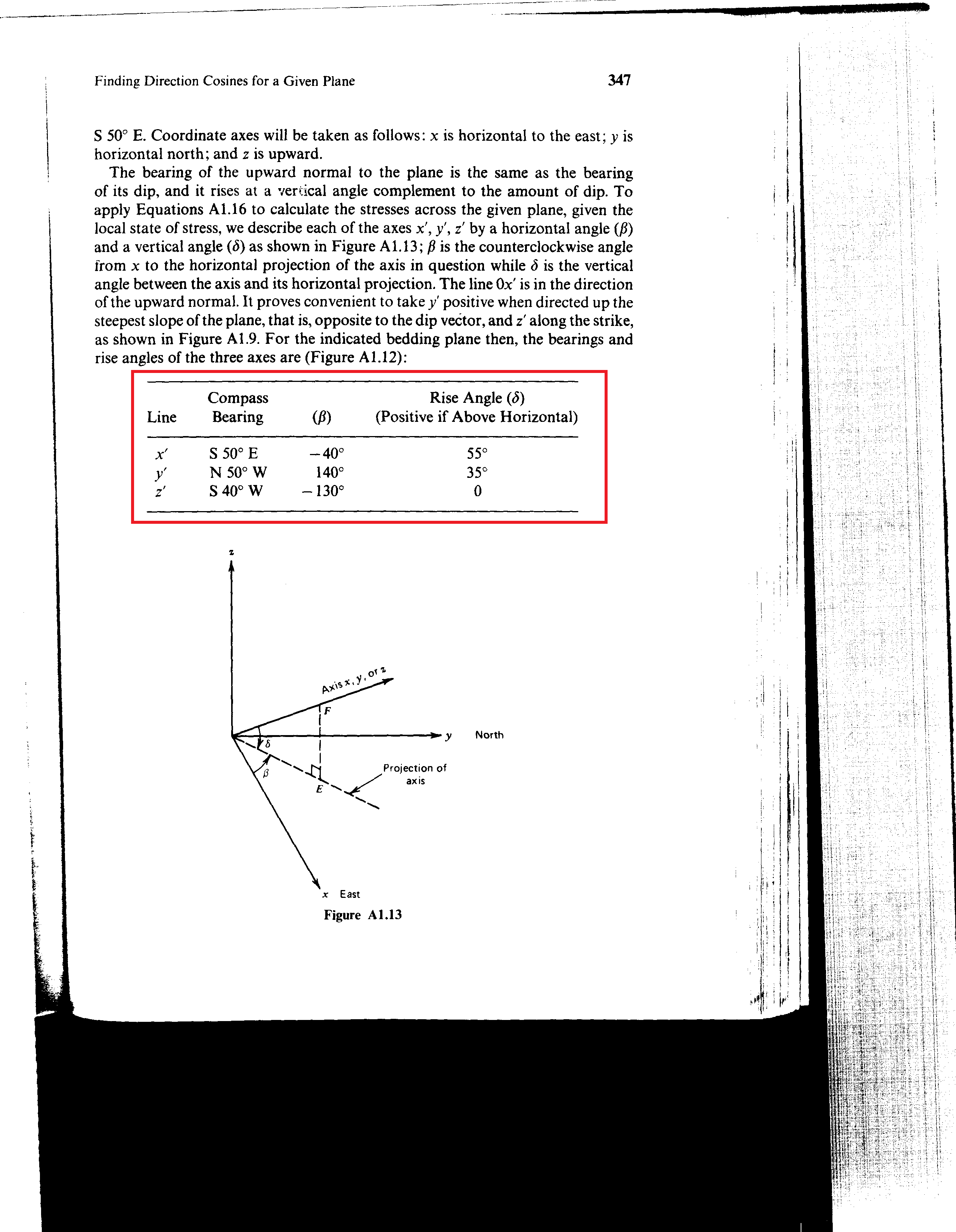}
        \caption{ }
    \end{subfigure}
\caption{Examples of images in UW3} 
\label{fig:UW3}
\end{figure}

The UW-3 Table dataset \cite{phillips1996user}  contains 1600 skew-corrected English document images with manually edited entity bounding box ground-truth. Page frames, text and non-text zones, text lines, and words are all surrounded by these bounding boxes. Each zone's type (text, math, table, half-tone, etc.) is also indicated. There are approximately 120 document images with at least one marked table zone. The UNLV and  UW-3 Table dataset taught a user how to use the T-Truth tool and asked him to prepare ground truth for the target images in the above dataset. Each image's ground truth is stored in an XML. Another expert manually validated the ground truths using the T-Truth tool's preview edit mode, and incorrect ground truths were corrected. These iterations were repeated several times to ensure that the ground truth was correct. authors discovered that the majority of errors occur when labeling column spanning cells where the column boundaries coincide with the word boundaries. Problems can also arise when there are multiple interpretations of a table structure, as described by Nagy\cite{hu2001table}, and domain knowledge is required to correctly label the table structure. Figure \ref{fig:UW3} illustrates two examples of this dataset.

\subsubsection{Marmot}
\begin{figure}[h!]
    
    \begin{subfigure}{0.50\columnwidth}
        \includegraphics[width=\textwidth]{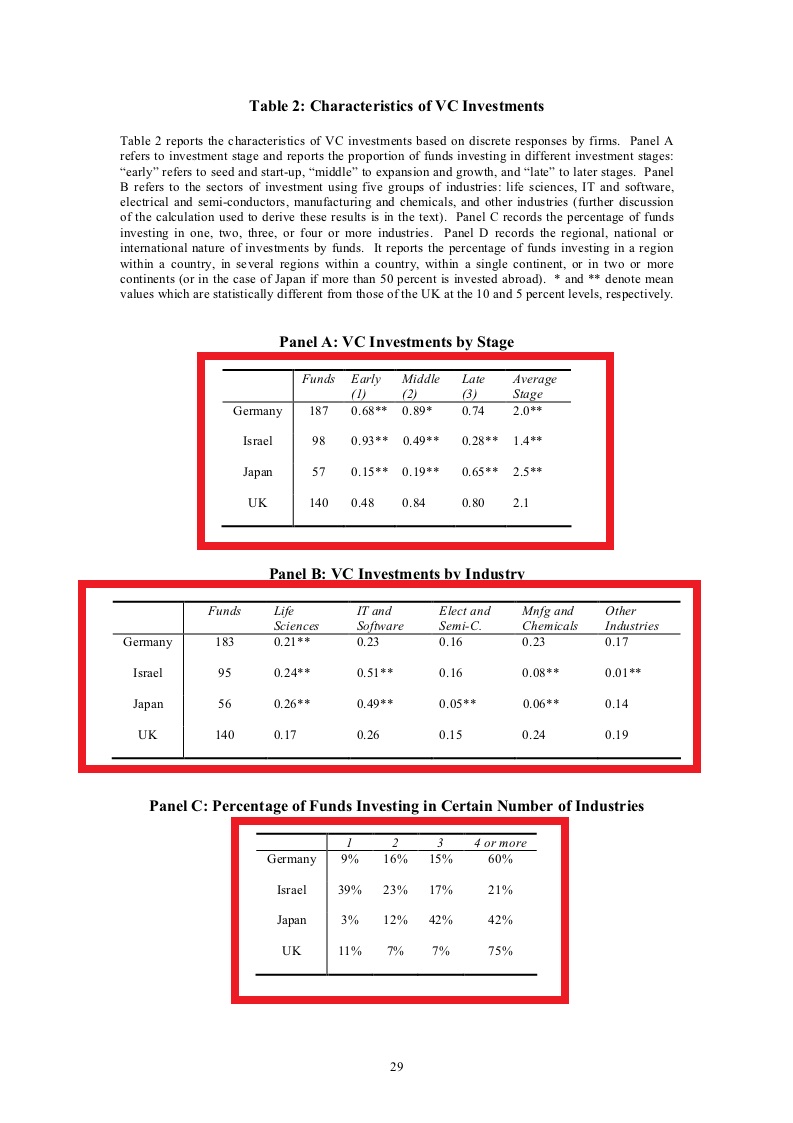}

        \caption{}
    \end{subfigure}
    \begin{subfigure}{0.50\columnwidth}
        \includegraphics[width=\textwidth]{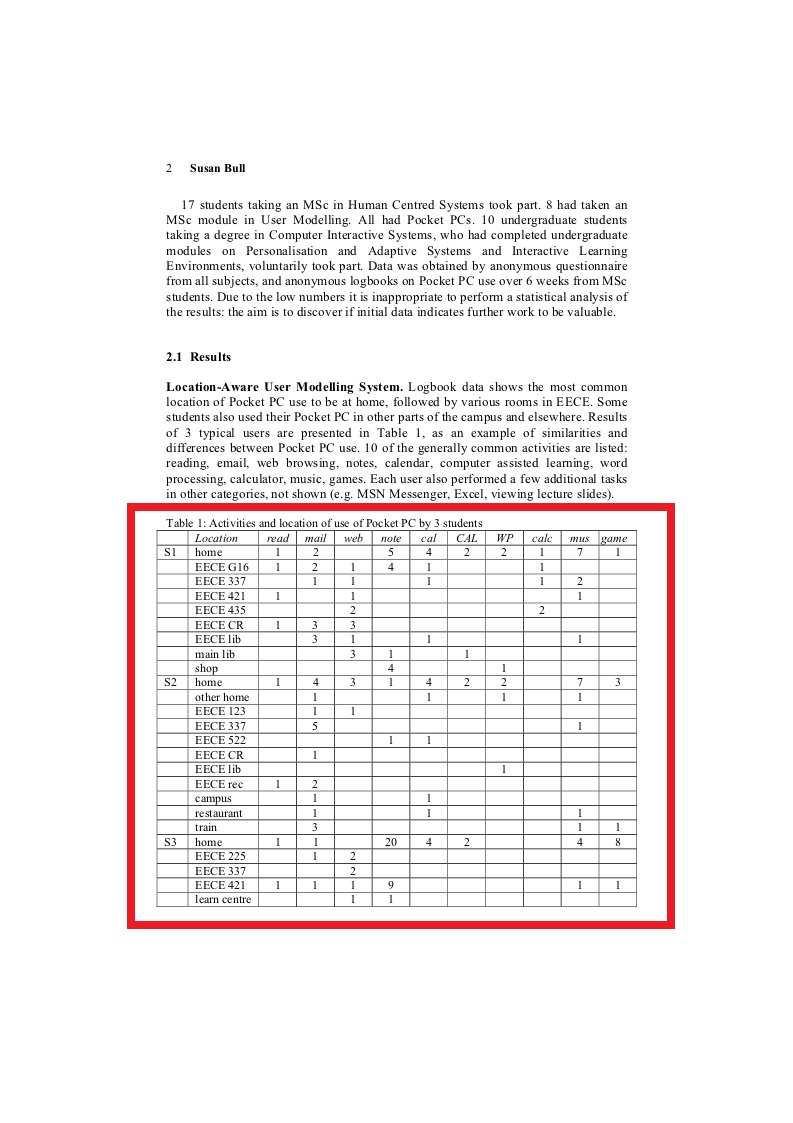}
        \caption{ }
    \end{subfigure}
\caption{Examples of images in Marmot} 
\label{fig:Marmot}
\end{figure}

The Marmot dataset\cite{fang2012dataset} is considered the first large dataset in table detection. it contains 2000 PDF pages with ground truth data. This labeling task was completed by 15 people. To reduce subjectivity, a unified labeling standard was established, and each ground-truth file is double-checked. The dataset's size is still growing. The e-document pages in the dataset have a wide range of language types, page layouts, and table styles. First, it is made up of roughly equal parts Chinese and English pages. The Chinese pages were chosen from over 120 e-Books with diverse subject areas from Founder Apabi's digital library, with no more than 15 pages chosen from each book. The English pages were retrieved from the web. Over 1500 conference and journal papers from 1970 to 2011 were crawled, covering a wide range of topics. The Chinese e-Book pages are mostly one column, whereas the English pages are printed in both one and two columns. This dataset includes a wide range of table types, from ruled tables to partially and non-ruled tables, horizontal tables to vertical tables, inside-column tables to span-column tables, and so on. A few samples from this dataset are shown in Figure \ref{fig:Marmot}.

\subsubsection{TableBank}
\begin{figure}[h!]
    
    \begin{subfigure}{0.50\columnwidth}
        \includegraphics[width=\textwidth]{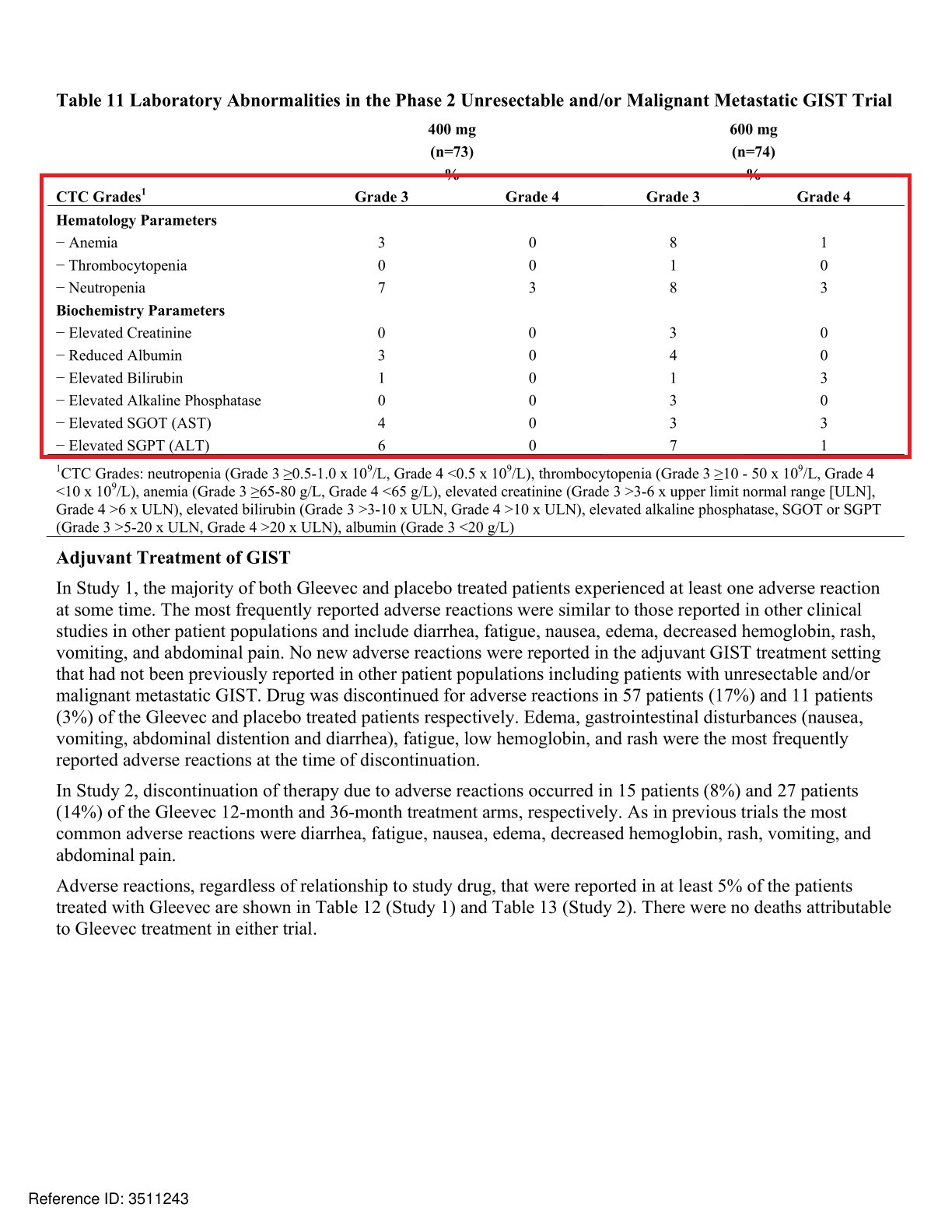}

        \caption{}
    \end{subfigure}
    \begin{subfigure}{0.50\columnwidth}
        \includegraphics[width=\textwidth]{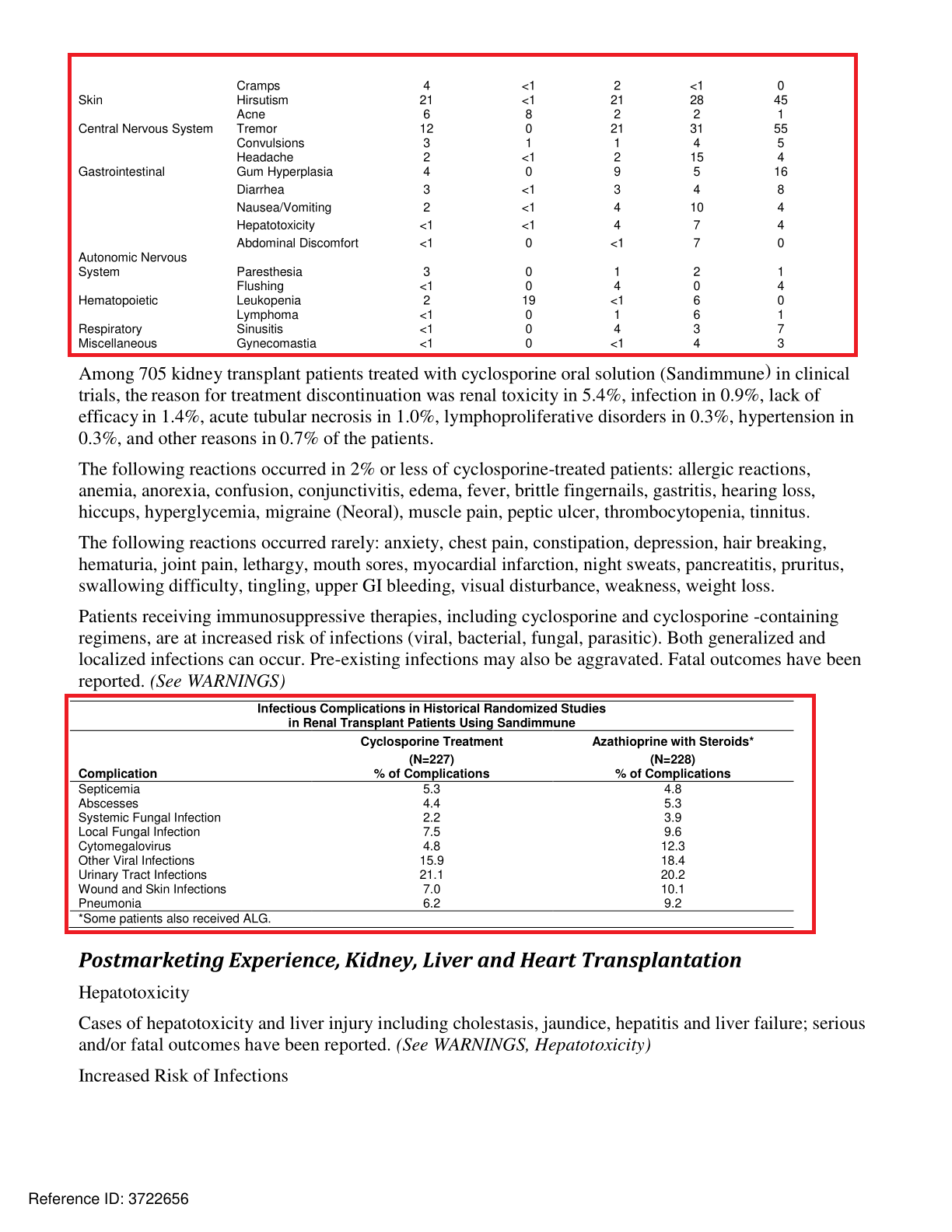}
        \caption{ }
    \end{subfigure}
\caption{Examples of images in TableBank} 
\label{fig:TableBank}
\end{figure}

The TableBank  \cite{li2020tablebank}  proposed a novel weak supervision approach for automatically creating the dataset, which is orders of magnitude larger than existing human-labeled datasets for table analysis. Unlike the traditional weakly supervised training set, this approach can generate not only large amounts of but also high-quality training data. There are many electronic documents available on the internet nowadays, such as Microsoft Word (.docx) and Latex (.tex) files. By definition, these online documents contain mark-up tags for tables in their source code. Intuitively, these source codes manipulate by adding bounding boxes within each document using the mark-up language. The Office XML code for Word documents can be modified to identify the borderline of each table. The code for Latex documents can also be modified to recognize table bounding boxes. This method generates high-quality labeled data for a wide range of domains, including business documents, official filings, research papers, and so on, which is extremely useful for large-scale table analysis tasks. The TableBank dataset is made up of 417,234 high-quality labeled tables and their original documents from a variety of domains. A few samples from this dataset are shown in Figure \ref{fig:TableBank}.

\subsubsection{DeepFigures}
\begin{figure}[h!]
        \centering
        \includegraphics[width=0.5\textwidth]{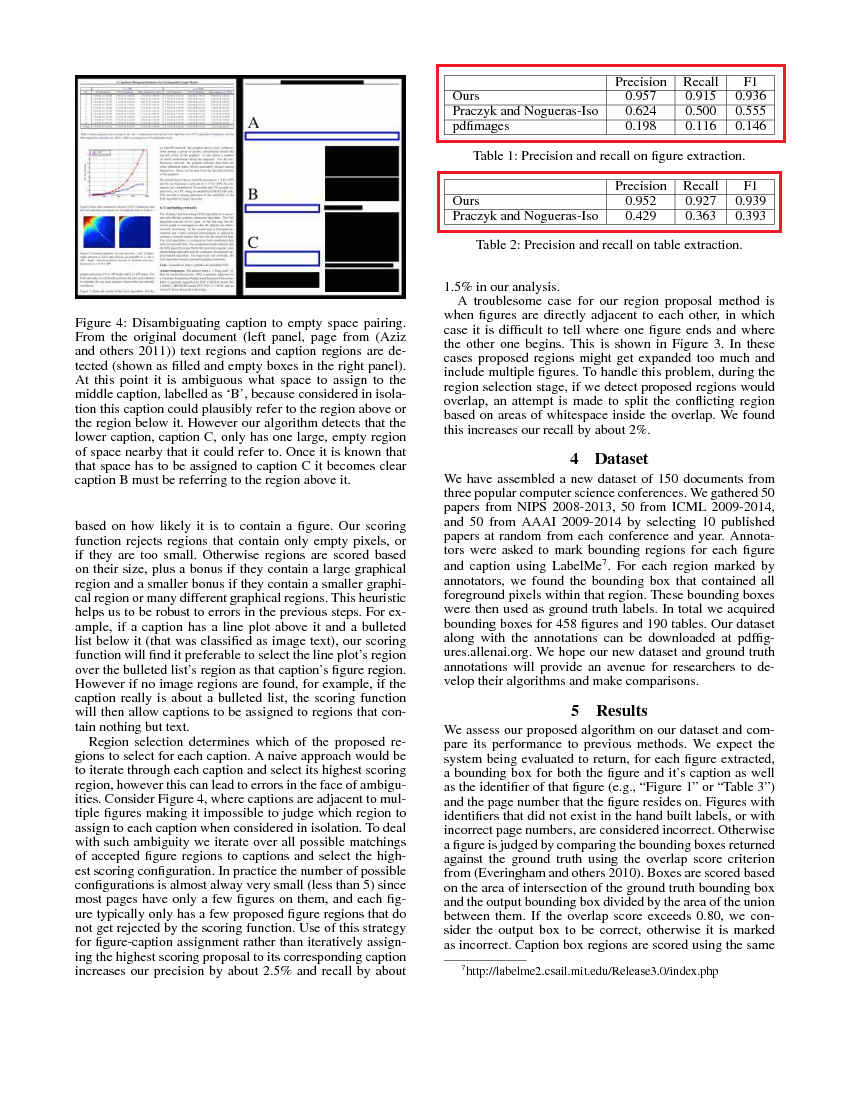}

\caption{Examples of images in DeepFigures} 
\label{fig:DeepFigures}
\end{figure}

DeepFigures \cite{siegel2018extracting} uses no human assistance to generate high-quality training labels for the task of figure extraction in a huge number of scientific papers. authors do this by locating figures and captions in the rasterized PDF using supplementary data from two big web collections of scientific articles (PubMed and arXiv). authors provide the resulting dataset of approximately 5.5 million tables and figures induced labels to facilitate the development of modern data-driven approaches for this task, which is 4,000 times larger than the previous largest figure extraction dataset and has an average precision of 96.8\%.  Samples from this dataset are shown in Figure \ref{fig:DeepFigures}.

\subsubsection{PubTables-1M}
\begin{figure}[h!]
    
    \begin{subfigure}{0.50\columnwidth}
        \includegraphics[width=\textwidth]{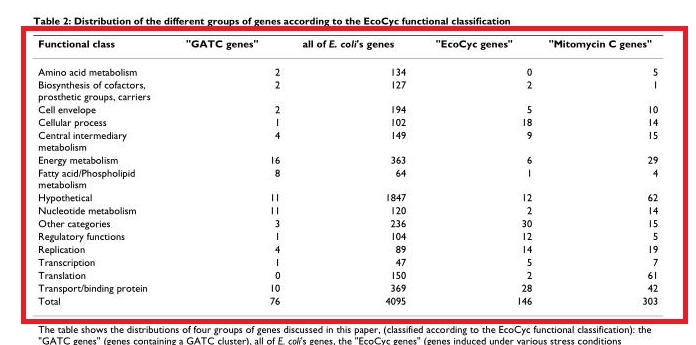}

        \caption{}
    \end{subfigure}
    \begin{subfigure}{0.50\columnwidth}
        \includegraphics[width=\textwidth]{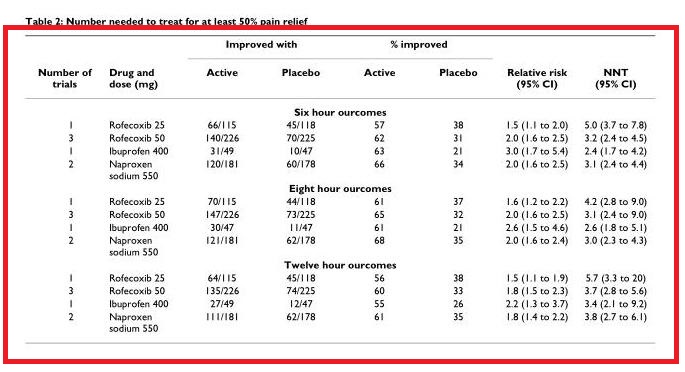}
        \caption{ }
    \end{subfigure}
\caption{Examples of images in PubTables-1M} 
\label{fig:PubTables-1M}
\end{figure}

PubTables-1M \cite{smock2021pubtables} contains nearly one million tables from scientific articles, supports multiple input modalities and contains detailed
header and location information for table structures, making it useful for a wide variety of modeling approaches. It also addresses a significant source of ground truth inconsistency observed in prior datasets called over-segmentation, using a novel canonicalization procedure. We demonstrate that these improvements lead to a significant increase in training performance and a more reliable estimate of model performance at evaluation for table structure recognition. Further, authors show that transformer-based object detection models trained on PubTables-1M produce excellent results for all three tasks of detection, structure recognition, and functional analysis without the need for any special customization for these tasks.  Figure \ref{fig:PubTables-1M} illustrates two examples of this dataset.

\subsubsection{SciTSR}
\begin{figure}[h!]
    
    \begin{subfigure}{0.50\columnwidth}
        \includegraphics[width=\textwidth]{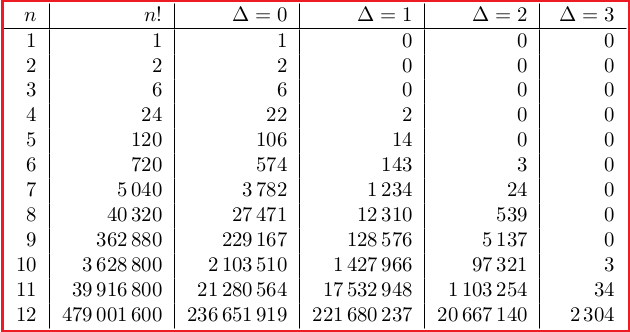}

        \caption{}
    \end{subfigure}
    \begin{subfigure}{0.50\columnwidth}
        \includegraphics[width=\textwidth]{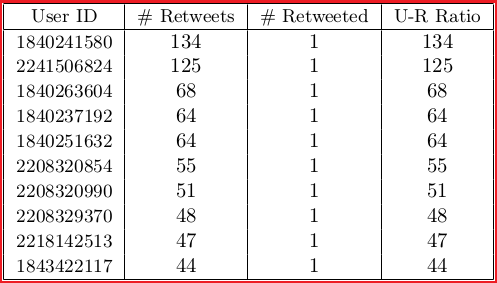}
        \caption{ }
    \end{subfigure}
\caption{Examples of images in SciTSR} 
\label{fig:SciTSR}
\end{figure}

SciTSR \cite{chi2019complicated} presents a large-scale table structure recognition dataset derived from scientific articles that comprise 15,000 tables from PDF files and their related structural labels.  Figure \ref{fig:SciTSR} illustrates two examples of this dataset.


\subsubsection{FinTabNet}
\begin{figure}[h!]
    
    \begin{subfigure}{0.50\columnwidth}
        \includegraphics[width=\textwidth]{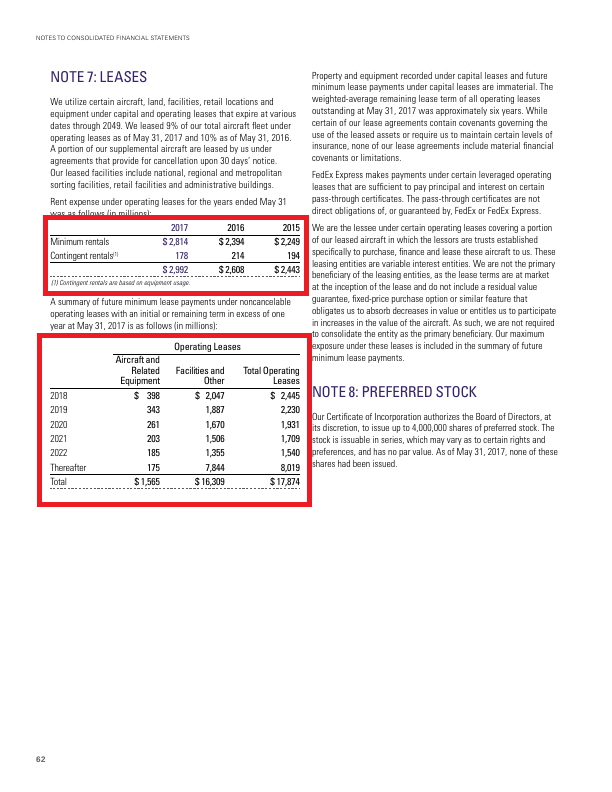}

        \caption{}
    \end{subfigure}
    \begin{subfigure}{0.50\columnwidth}
        \includegraphics[width=\textwidth]{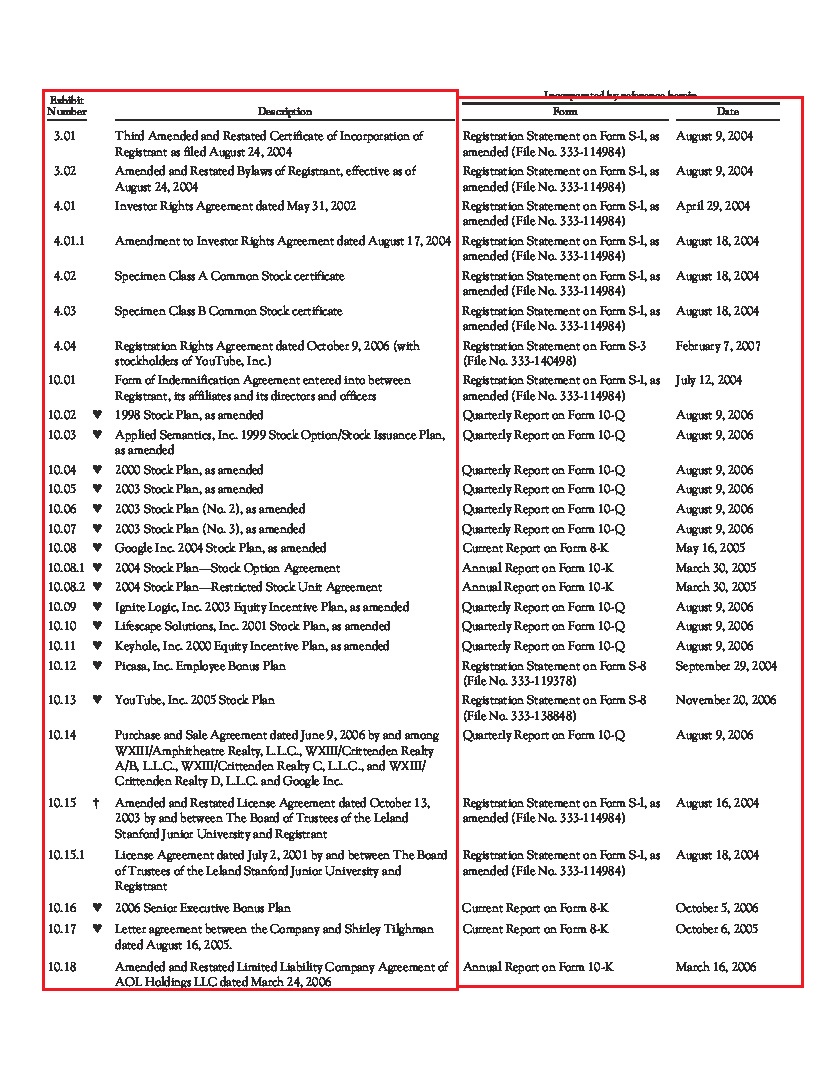}
        \caption{ }
    \end{subfigure}
\caption{Examples of images in FinTabNet} 
\label{fig:FinTabNet}
\end{figure}

FinTabNet \cite{zheng2020global} introduces GTE, a vision-guided systematic framework for combined table detection and cell structured identification that can be constructed on top of any object detection model. Create a new penalty based on the natural cell containment constraint of tables with GTE-Table to train their table network with the help of cell location predictions. GTE-Cell is a novel hierarchical cell detection network that uses table layouts to detect cells. Build a technique for automatically labeling table and cell structures in existing texts to create a huge corpus of training and test data for a low cost. FinTabNet is a collection of real-world and complicated scientific and financial datasets with thorough table structure annotations to aid in structure identification training and testing.  Figure \ref{fig:FinTabNet} illustrates two examples of this dataset.

\subsubsection{PubTabNet}
\begin{figure}[h!]
    
    \begin{subfigure}{0.50\columnwidth}
        \includegraphics[width=\textwidth]{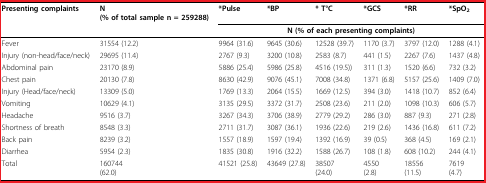}

        \caption{}
    \end{subfigure}
    \begin{subfigure}{0.50\columnwidth}
        \includegraphics[width=\textwidth]{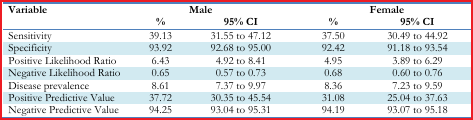}
        \caption{ }
    \end{subfigure}
\caption{Examples of images in PubTabNet} 
\label{fig:PubTabNet}
\end{figure}

PubTabNet \cite{zhong2020image} one of the biggest openly accessible table recognition collection, including 568k table pictures and structured HTML representations. PubTabNet is built automatically by comparing the XML and PDF formats of scientific publications in the PubMed CentralTM Open Access Subset (PMCOA). authors also suggest an attention-based encoder-dual-decoder (EDD) architecture for converting table graphics to HTML code. A structure decoder is included in the model, which reconstructs the table structure and assists the cell decoder in recognizing cell content. Furthermore, authors also propose a new Tree-Edit-Distance-based Similarity (TEDS) metric for table recognition that better captures multi-hop cell misalignment and OCR errors than the existing metric.  Figure \ref{fig:PubTabNet} illustrates two examples of this dataset.

\subsubsection{TNCR}
\begin{figure}[h!]
    
    \begin{subfigure}{0.50\columnwidth}
        \includegraphics[width=\textwidth]{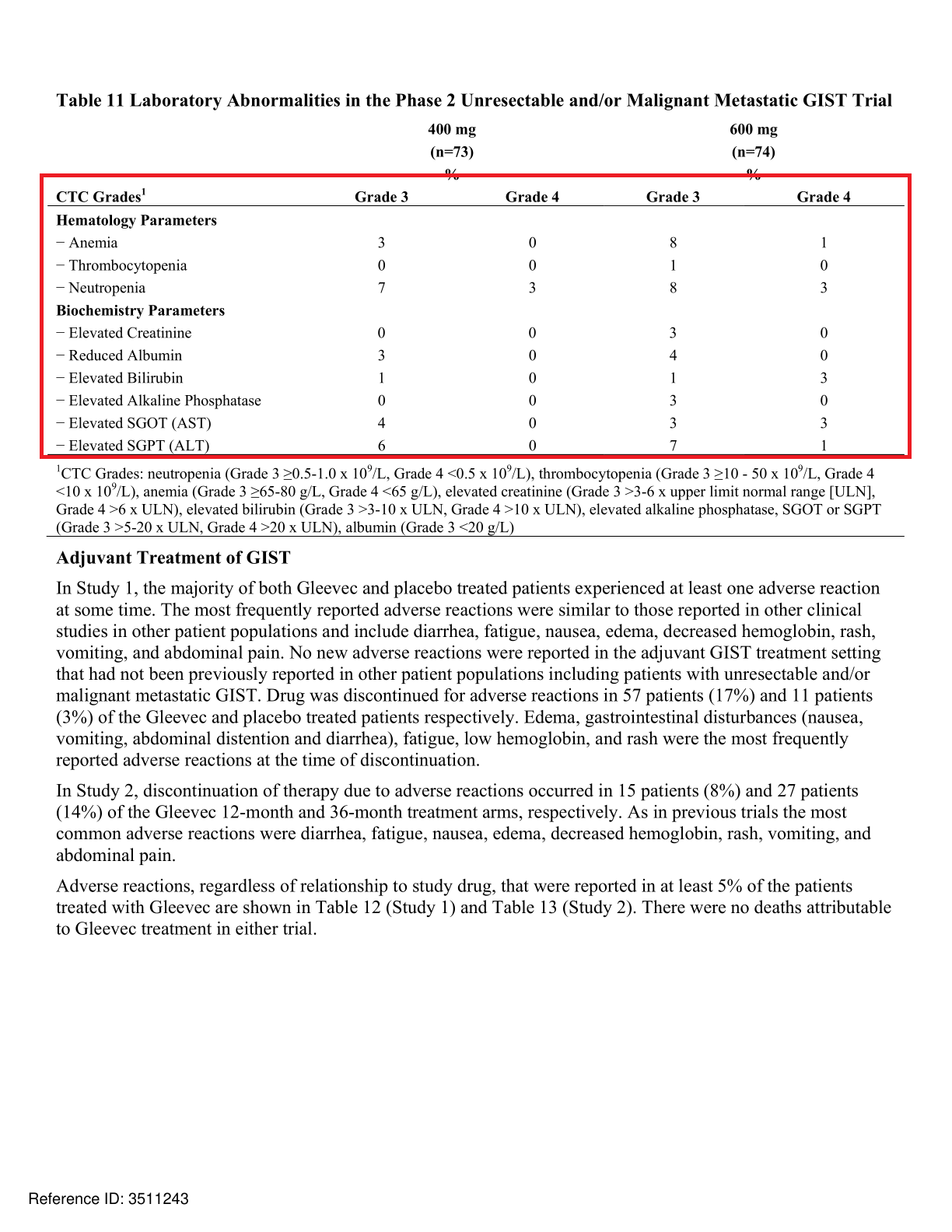}

        \caption{}
    \end{subfigure}
    \begin{subfigure}{0.50\columnwidth}
        \includegraphics[width=\textwidth]{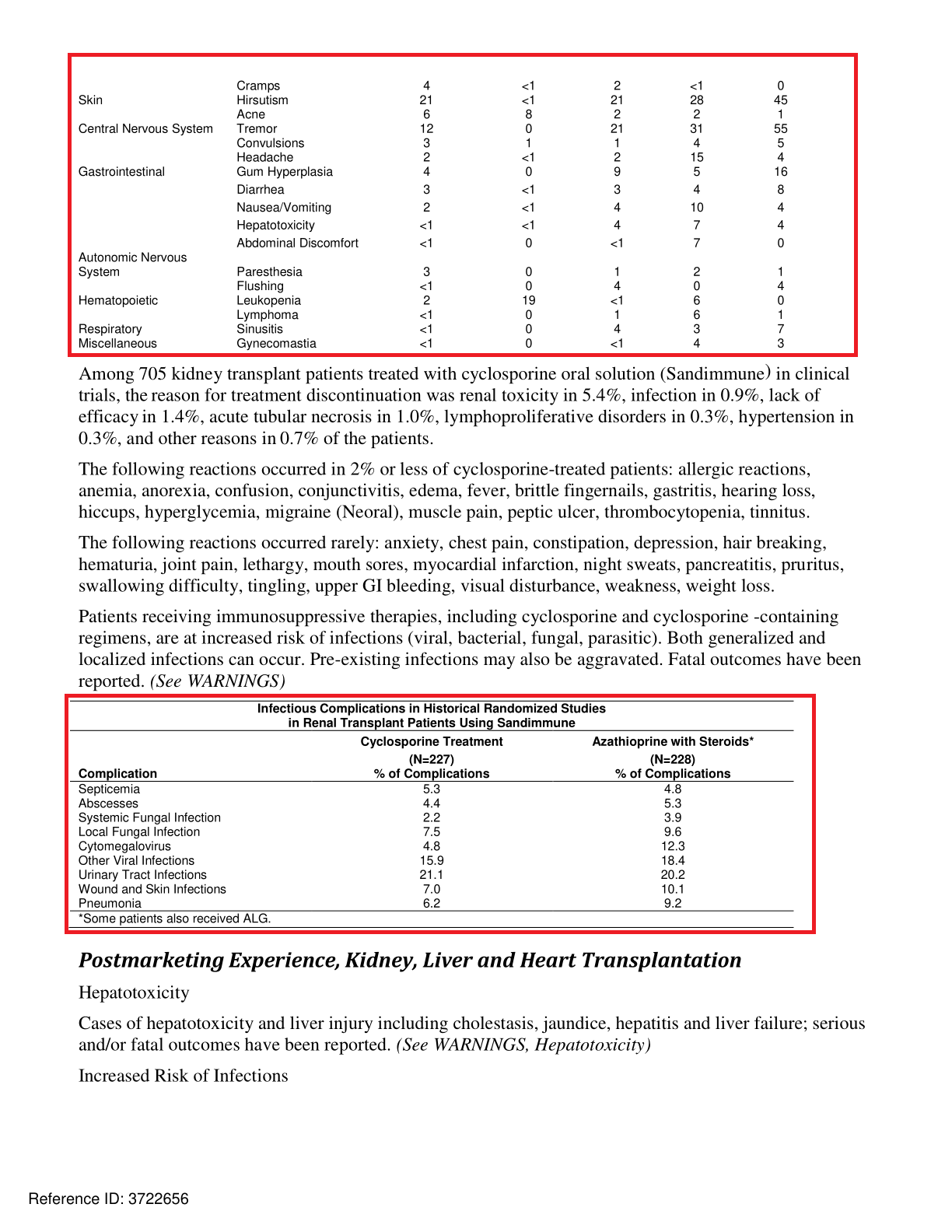}
        \caption{ }
    \end{subfigure}
\caption{Examples of images in TNCR} 
\label{fig:TNCR}
\end{figure}

TNCR \cite{abdallah2022tncr} a new table collection containing images of varied quality gathered from free access websites The TNCR dataset may be used to recognize tables in scanned document pictures and classify them into five categories. TNCR has roughly 6621 photos and 9428 captioned tables. To build numerous robust baselines, this work used state-of-the-art deep learning-based approaches for table detection. On the TNCR dataset, Deformable DERT with Resnet-50 Backbone Network delivers the best results compared to other methods, with an accuracy of 86.7\%, recall of 89.6\%, and f1 score of 88.1\%. A few samples from this dataset are shown in Figure \ref{fig:TNCR}.

\subsubsection{SynthTabNet}
\begin{figure}[h!]
    
    \begin{subfigure}{0.50\columnwidth}
        \includegraphics[width=\textwidth]{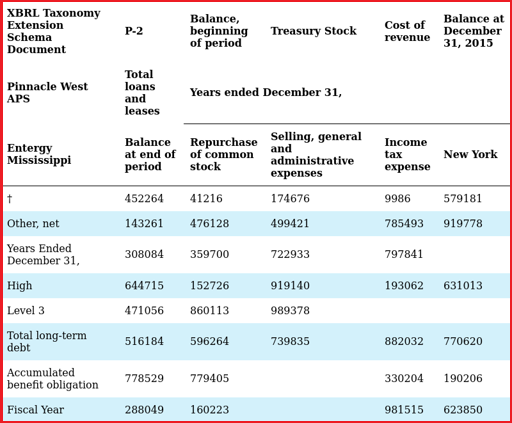}

        \caption{}
    \end{subfigure}
    \begin{subfigure}{0.50\columnwidth}
        \includegraphics[width=\textwidth]{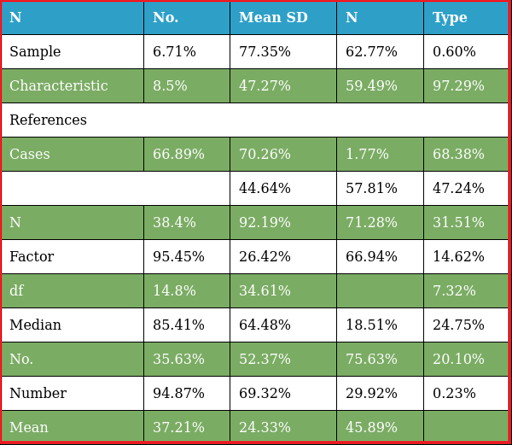}
        \caption{ }
    \end{subfigure}
\caption{Examples of images in SynthTabNet} 
\label{fig:SynthTabNet}
\end{figure}

To correct an imbalance in the earlier datasets, A Nassar \cite{nassar2022tableformer} proposes SynthTabNet, a synthetically created dataset with a variety of appearance styles and complexity. The authors have created four synthetic datasets, each with 150k samples. The most common words from PubTabNet and FinTabNet as well as randomly produced text make up the corpora used to create the table content. The first two synthetic datasets have been adjusted to closely resemble the look of the real datasets while incorporating more intricate table structures. The third one adopts a colorful style with strong contrast, while the final one has tables with little content. Last but not least, The authors have integrated all synthetic datasets into a single, 600k-example synthetic dataset. A few samples from this dataset are shown in Figure \ref{fig:SynthTabNet}.

Table \ref{tab:dataset_compression_1} presents a comparison between some of the popular datasets of table detection and structure recognition. 
\begin{table}[h!]
\caption{The table illustrates a quantitative comparison between some famous datasets in table detection.}
\begin{adjustbox}{width=1\textwidth}
\begin{tabular}{|c|c|c|c|c|c|c|} 
    \hline
      Dataset & Total pages & Total Tables &Table detection &Table Structure & Classification & Scanned \\ \hline
      ICDAR2013         & 462  &     150     & \cmark & \cmark &   \xmark &  \cmark \\ \hline
      
      ICDAR2017-POD     & 2,417  &     -     & \cmark & \xmark &   \xmark &  \cmark \\ \hline
      TabStructDB       & 2.4k  &     -      & \xmark & \cmark &   \xmark &  \cmark \\ \hline
      TABLE2LATEX-450K  &  -  &  450,000     & \xmark & \cmark &   \xmark &  \cmark \\ \hline
      RVL-CDIP (SUBSET) & 518  &     -       & \cmark & \xmark &   \xmark &  \cmark \\ \hline
      IIIT-AR-13K       & 13K  &     -       & \cmark & \xmark &   \xmark &  \cmark \\ \hline
      CamCap            & 85  &     -        & \cmark & \cmark &   \xmark &  \xmark \\ \hline
      
      UNLV              & 2889 &      -      & \cmark & \cmark &   \xmark &  \cmark \\ \hline
      UW-3 dataset      & 1600 &      -      & \cmark & \cmark &   \xmark &  \cmark \\ \hline
      Marmot            & 2000 &     -       & \cmark & \xmark &   \xmark &  \cmark \\ \hline
      TableBank         &  -   &  417,234    & \cmark & \xmark &   \xmark &  \cmark \\ \hline
      ICDAR2019         &  -   &    2000     & \cmark & \cmark &   \xmark &  \cmark \\ \hline
      DeepFigures       &  -   & 5.5 million & \cmark & \xmark &   \xmark &  \cmark \\ \hline
      PubTables-1M      &460,589& 1 million  & \cmark & \cmark &   \xmark & \cmark  \\ \hline
      SciTSR            &  -    &  15,000    & \xmark & \cmark &   \xmark & \xmark  \\ \hline 
      FinTabNet         &89,646&    112,887  & \cmark & \cmark &   \xmark & \xmark  \\ \hline 
      PubTabNet         &  -    &    568k    & \xmark & \cmark &   \xmark & \cmark  \\ \hline 
      TNCR              & 6621 &    9428     & \cmark & \xmark &   \cmark & \cmark  \\ \hline 
      SynthTabNet       & 600k &     -       & \cmark & \cmark &   \cmark & \cmark  \\ \hline 
      
\end{tabular}
\end{adjustbox}
\label{tab:dataset_compression_1}
\end{table}

\subsection{Metrics}

Table detectors use multiple criteria to measure the performance of the detectors viz., frames per second (FPS), precision, and recall. However, mean Average Precision (mAP) is the most common evaluation metric. Precision is derived from Intersection over Union (IoU), which is the ratio of the area of overlap and the area of union between the ground truth and the predicted bounding box. A threshold is set to determine if the detection is correct. If the IoU is more than the threshold, it is classified as True Positive while an IoU below it is classified as False Positive. If the model fails to detect an object present in the ground truth, it is termed a False Negative. Precision measures the percentage of correct predictions while recall measures the correct predictions with respect to the ground truth.
\begin{equation}
  \begin{aligned}
    \textrm{Average Precision (AP)} &= \frac{\textrm{True Positive (TP)}}{( \textrm{True Positive (TP)}+\textrm{False Positive (FP)} )} \\  &=\frac{True Positive } { All Observations}
\end{aligned}
\end{equation}

\begin{equation}
\begin{aligned}
    \textrm{Average Recall (AR)} &= \frac{\textrm{True Positive (TP)}}{( \textrm{True Positive (TP)}+\textrm{False Negative (FN) } )}\\ &=  \frac{True Positive } { All Ground Truth}
\end{aligned}
\end{equation}

\begin{equation}
\begin{aligned}
    \textrm{F1-score} &= \frac{ 2 * (\textrm{AP}* \textrm{AR} )}{(\textrm{AP}+\textrm{AR})}
\end{aligned}
\end{equation}

Based on the above equation, average precision is computed separately for each class. To compare performance between the detectors, the mean of average precision of all classes, called mean average precision (mAP) is used, which acts as a single metric for final evaluation.

IOU is a metric that finds the difference between ground truth annotations and predicted bounding boxes. This metric is used in most state of art object detection algorithms. In object detection, the model predicts multiple bounding boxes for each object and based on the confidence scores of each bounding box it removes unnecessary boxes based on their threshold value. We need to declare the threshold value based on our requirements.
\begin{equation}
\begin{aligned}
    \textrm{IOU} &= \frac{\textrm{Area of union } }{\textrm{area of intersection}}
\end{aligned}
\end{equation}

\section{Table detection and structure recognition Models}
Table detection has been studied for an extended period of time. Researchers used different methods that can be categorized as follows:
\begin{enumerate}
    \item heuristic-based methods
    \item machine learning-based methods
    \item deep learning-based methods
\end{enumerate}

Primarily heuristic-based methods were mainly used in the 1990s, 2000s, and early 2010. They employed different visual cues like lines, keywords, space features, etc. to detect tables.

P. Pyreddy et al. \cite{pyreddy1997tinti} proposed an approach of detecting tables using character alignment, holes, and gaps. Wang et al. \cite{wangt2001automatic}. used a statistical approach to detect table lines depending on the distance between consecutive words. Grouped horizontal consecutive words together with vertical adjacent lines were employed to propose table entity candidates. Jahan et al. \cite{jahan2014locating} presented a method that uses local thresholds for word spacing and line height for detecting table regions.

Itonori \cite{itonori1993table} proposed a rule-based approach that led to the text-block arrangement and ruled line position to localize the table in the documents. Chandran and Kasturi \cite{chandran1993structural} developed another table detection approach based on vertical and horizontal lines. Wonkyo Seo et al. \cite{seo2015junction} used junctions (intersection of the horizontal and vertical line) detection with further processing. 

Hassan et al. \cite{hassan2007table} locate and segment tables by analyzing spatial features of text blocks. Ruffolo et al. \cite{oro2009trex} introduced PDF-TREX, a heuristic bottom-up approach for table recognition in single-column PDF documents. It uses the spatial features of page elements to align and group them into paragraphs and tables. Nurminen \cite{nurminen2013algorithmic} proposed a set of heuristics to locate subsequent text boxes with common alignments and assign them the probability of being a table.

Fang et al. \cite{fang2012table} used the table header as a starting point to detect the table region and decompose its elements. Harit et al. \cite{harit2012table} proposed a technique for table detection based on the identification of unique table start and trailer patterns. Tupaj et al. \cite{tupaj1996extracting} proposed an OCR based table detection technique. The system searches for sequences of table-like lines based on the keywords 

The above methods work relatively  well on documents with uniform layouts. However, heuristic rules need to be tweaked to a wider variety of tables and are not really suited for generic solutions. Therefore, machine learning approaches started to be employed to solve the table detection problem.

Machine learning-based methods were common around the 2000s and the 2010s.

Kieninger  et al. \cite{kieninger1998t} applied an unsupervised learning approach by clustering word segments. Cesarini et al. \cite{cesarini2002trainable} used a modified XY tree supervised learning approach. Fan et al.\cite{fan2015table} uses both supervised and unsupervised approaches to table detection in PDF documents. Wang and Hu \cite{wang2002machine} applied Decision tree and SVM classifiers to layout, content type and word group features. T. Kasar et al. \cite{kasar2013learning} used the junction detection and then passed the information to the SVM classifier. Silva et al. \cite{e2009learning} applied joint probability distribution over sequential observations of visual page elements (Hidden Markov Models) to merge potential table lines into tables. Klampfl et al. \cite{klampfl2014comparison} compare two unsupervised table recognition methods from digital scientific articles. Docstrum algorithm \cite{o1993document} applies KNN to aggregate structures into lines and then uses perpendicular distance and angle between lines to combine them into text blocks. It must be noted that this algorithm was devised in 1993, earlier than other methods mentioned in this section. 

F Shafait \cite{shafait2010table} proposes a useful method for table recognition that performs well on documents with a range of layouts, including business reports, news stories, and magazine pages. The Tesseract OCR engine offers an open-source implementation of the algorithm.

As neural networks gained interest, researchers started to apply them to document layout analysis tasks. Initially, they were used at simpler tasks like table detection. Later on, as more complex architectures were developed, more work was put into table columns and overall structure recognition.

Hao et al. \cite{hao2016table} employed CNN to detect whether a certain region proposal is a table or not. Azka Gilani et al. \cite{gilani2017table} proposed a Faster R-CNN-based model to make up for the limitations of Hao et al. \cite{hao2016table} and other prior methodologies.

Sebastian Schreiber et al.  \cite{schreiber2017deepdesrt} were the first to perform table detection and structure recognition using Faster RCNN. He et al. \cite{he2017multi}, used FCN for semantic page segmentation. S. Arif et al. \cite{arif2018table} attempted to improve the accuracy of Faster R-CNN by using semantic color-coding of text. Reza et al. \cite{reza2019table} used a combination of GAN-based architecture for table detection. Agarwal et al. \cite{agarwal2021cdec} used a multistage extension of Mask R-CNN with a dual backbone for detecting tables.


Recently transformer-based models were applied to document layout analysis, Smock, Brandon et al. \cite{smock2021pubtables} applied Carion et al.\cite{carion2020end} DEtection TRansformer framework, a transformer encoder-decoder architecture, to their table dataset for both table detection and structure recognition tasks. Xu et al. \cite{li2022dit} proposed a self-supervised pre-trained Document Image Transformer model using large-scale unlabeled text images for document analysis, including table detection

\subsection{Table detection Models}
In this section, we examine the deep learning methods used for document image table detection. We have divided the methods into several deep learning ideas for the benefit of our readers' convenience. Table \ref{tab:Table Detection Methods} lists all the object identification-based table detection strategies.  It also discusses various deep learning-based methods that have been used in these methods.

A Gilani \cite{gilani2017table} has shown how to recognize tables using deep learning. Document pictures are pre-processed initially in the suggested technique. These photos are then sent into a Region Proposal Network for table detection, which is followed by a fully connected neural network. The suggested approach works with great precision on a variety of document pictures, including documents, research papers, and periodicals, with various layouts.

D Prasad \cite{prasad2020cascadetabnet} presents an automatic table detection approach for interpreting tabular data in document pictures, which primarily entails addressing two issues: table detection and table structure recognition.  Using a single Convolution Neural Network (CNN) model, provide an enhanced deep learning-based end-to-end solution for handling both table detection and structure recognition challenges. CascadeTabNet is a Cascade mask Region-based CNN High-Resolution Network (Cascade mask R-CNN HRNet)-based model that simultaneously identifies table areas and recognizes structural body cells from those tables.

SS Paliwal \cite{paliwal2019tablenet} presents TableNet which is a new end-to-end deep learning model for both table detection and structure recognition. To divide the table and column areas, the model uses the dependency between the twin objectives of table detection and table structure recognition. Then, from the discovered tabular sub-regions, semantic rule-based row extraction is performed.

Y Huang \cite{huang2019yolo} describes a table detecting algorithm based on the YOLO principle. The authors offer various adaptive improvements to YOLOv3, including an anchor optimization technique and two post-processing methods, to account for the significant differences between document objects and real objects. also employ k-means clustering for anchor optimization to create anchors that are more suited for tables than natural objects, making it easier for our model to find the exact placements of tables. The additional whitespaces and noisy page objects are deleted from the projected results during the post-processing procedure.

L Hao \cite{hao2016table} offers a new method for detecting tables in PDF documents that are based on convolutional neural networks, one of the most widely used deep learning models. The suggested method begins by selecting some table-like areas using some vague constraints, then building and refining convolutional networks to identify whether the selected areas are tables or not. Furthermore, the convolutional networks immediately extract and use the visual aspects of table sections, while the non-visual information contained in original PDF documents is also taken into account to aid in better detection outcomes.

SA Siddiqui \cite{siddiqui2018decnt} provide a novel strategy for detecting tables in documents. The approach given here takes advantage of data's potential to recognize tables with any arrangement. however, the given method works directly on photos, making it universally applicable to any format. The proposed method uses a unique mix of deformable CNN and speedier R-CNN/FPN. Because tables might be present at variable sizes and transformations, traditional CNN has a fixed receptive field, which makes table recognition difficult (orientation). Deformable convolution bases its receptive field on the input, allowing it to shape it to match the input. The network can accommodate tables of any layout because of this customization of the receptive field.

N Sun \cite{sun2019faster} presents a corner-finding approach for faster R-CNN-based table detection. The Faster R-CNN network is first used to achieve coarse table identification and corner location. then, coordinate matching is used to group those corners that belong to the same table. Untrustworthy edges are filtered at the same time. Finally, the matching corner group fine-tunes and adjusts the table borders. At the pixel level, the suggested technique enhances table boundary finding precision.

I Kavasidis\cite{kavasidis2019saliency}  propose a method for detecting tables and charts using a combination of deep CNNs, graphical models, and saliency ideas. M Holeček \cite{holevcek2019table} presented the concept of table understanding utilizing graph convolutions in structured documents like bills, extending the applicability of graph neural networks. A PDF document is used in the planned research as well. The job of line item table detection and information extraction are combined in this study to tackle the problem of table detection. Any word may be quickly identified as a line item or not using the line item technique. Following word classification, the tabular region may be easily identified since, in contrast to other text sections on bills, table lines are able to distinguish themselves rather effectively.

Á Casado-García \cite{casado2020benefits} Uses object detection techniques, The authors have shown that fine-tuning from a closer domain improves the performance of table detection after conducting a thorough examination. The authors have utilized Mask R-CNN, YOLO, SSD, and Retina Net in conjunction with object detection algorithms. Two basic datasets are chosen to be used in this investigation, TableBank and PascalVOC.

X Zheng \cite{zheng2021global} provides Global Table Extractor (GTE), a method for jointly detecting tables and recognizing cell structures that can be implemented on top of any object detection model. To train their table network with the help of cell placement predictions, the authors develop GTE-Table, which introduces a new penalty based on the inherent cell confinement limitation of tables. A novel hierarchical cell identification network called GTE-Cell makes use of table styles. Additionally, in order to quickly and inexpensively build a sizable corpus of training and test data, authors develop a method to automatically classify table and cell structures in preexisting texts.

Y Li \cite{li2019gan} provides a new network to produce the layout elements for table text and to enhance the performance of less ruled table identification. The Generative Adversarial Networks(GAN) and this feature generator model are comparable. The authors mandate that the feature generator model extract comparable features for both heavily governed and loosely ruled tables.

DD Nguyen \cite{nguyen2022tablesegnet} introduces TableSegNet, a fully convolutional network with a compact design that concurrently separates and detects tables. TableSegNet uses a shallower path to discover table locations in high resolution and a deeper path to detect table areas in low resolution, splitting the found regions into separate tables. TableSegNet employs convolution blocks with broad kernel sizes throughout the feature extraction process and an additional table-border class in the main output to increase the detection and separation capabilities.

D Zhang \cite{zhang2022yolo} suggests a YOLO-table-based table detection methodology. To enhance the network's capacity to learn the spatial arrangement aspects of tables, the authors incorporate involution into the network's core, and the authors create a simple Feature Pyramid Network to increase model efficacy. This research also suggests a table-based enhancement technique.

\begin{table}[h!]
\caption{A comparison of the benefits and drawbacks of several deep learning-based table detection methods}
\begin{adjustbox}{width=1\textwidth}
\begin{tabular}{|c|c|c|c|} 
    \hline
      Literature & Method & Benefits & Drawbacks  \\ \hline
      
      A Gilani\cite{gilani2017table}         & \makecell{Faster R-CNN}     &     \makecell{ 1) On scanned document pictures, this is the\\ first deep learning-based table detection method.
      \\ 2) The object detection technique is made easier \\by converting RGB pixels to distance measures.}     & \makecell{There are additional phases \\in the pre-processing process.}\\ \hline
      
      S Schreiber\cite{schreiber2017deepdesrt}  &  \makecell{transfer learning methods\\ + Faster R-CNN}   &    \makecell{end-to-end strategy for detecting tables and \\table structures that is straightforward \\and efficient } & \makecell{When compared to other\\ state-of-the-art techniques, \\it is less accurate. }\\ \hline
      
      SA Siddiqui \cite{siddiqui2018decnt}  & \makecell{ Deformable CNN + \\Faster R-CNN}&     \makecell{ Deformable convolutional neural networks'\\ dynamic receptive field aids in the reconfiguration \\of multiple tabular boundries.  }  & \makecell{ When compared to standard  \\convolutions, deformable  \\convolutions are computationally \\ demanding.}  \\ \hline 
      
      SS Paliwal \cite{paliwal2019tablenet}   &  \makecell{ Networks with fully\\ convolutions }   &    \makecell{ 1) First attempt at combining a single solution\\ to handle both the problem of table detection and \\structure recognition. 2) A comprehensive method\\ for  structure recognition and detection in document \\pictures. }    & \makecell{ This approach only functions on\\ column detection when used for\\ table structure extraction. }  \\ \hline  
      
      P Riba \cite{riba2019table}   &  \makecell{ OCR-based Graph NN that \\makes use of textual \\characteristics }   &    \makecell{ The suggested technique makes use of more data\\ than only spatial attributes.}    & \makecell{1) No comparisons to other state-of-\\the-art strategies. 2) Additional\\ annotations are needed using this \\strategy in addition to the tabular\\ data. }  \\ \hline 
      
      N Sun \cite{sun2019faster}   & \makecell{ Faster R-CNN + \\Locate corners} &    \makecell{1) Better outcomes are obtained using a  novel \\technique.  2) Faster R-CNN is used to identify not \\just tables, but also the corners of tabular borders}    &  \makecell{ 1) It is necessary to do postprocessing\\ operations such as corner refining. \\2) Because of the additional detections, \\the computation is more involved.}  \\ \hline 
      
       I Kavasidis \cite{kavasidis2019saliency}   &  \makecell{combination of deep CNNs,\\ graphical models, and\\ saliency }   &    \makecell{ 1) Dilated convolutions rather than conventional\\ convolutions are used. 2) Using this technique, \\saliency detection is performed in place of table\\ detection. }    & \makecell{ To provide equivalent results, many \\processing stages are necessary. }  \\ \hline 
       
       M Holeček \cite{holevcek2019table}   &  \makecell{ Graph NN + line item \\identification Method}   &    \makecell{ This approach yields encouraging outcomes when\\ used to layout-intensive documents like invoices \\and PDFs. }    & \makecell{ 1) Limited baseline approach without\\ comparisons to other state-of-the-art \\techniques 2) No publicly accessible \\table datasets are used for the\\ evaluation of the approach. }  \\ \hline 
       
      Y Huang \cite{huang2019yolo}   &  \makecell{ YOLO }   &    \makecell{ In comparison, a quicker and more effective\\ strategy }    & \makecell{ The suggested methodology relies on\\ data-driven post-processing methods.}  \\ \hline 
      
      Y Li \cite{li2019gan}   &  \makecell{Generative Adversarial Networks(GAN) }   &    \makecell{ For ruling and less ruled tables, the GAN-based\\ strategy drives the network to extract comparable\\ characteristics. }    & \makecell{ In document images with different\\ tabular layouts, the generator-based\\ model is susceptible.}  \\ \hline 
      
      M Li \cite{li2020tablebank}   &  \makecell{  Faster R-CNN }   &    \makecell{This method demonstrates how a basic Faster \\R-CNN can yield excellent results when used \\with a huge dataset like TableBank. }    & \makecell{ Just a simple Faster-RCNN\\ implementation}  \\ \hline 
      
      D Prasad \cite{prasad2020cascadetabnet}   &  \makecell{ Cascade mask Region-based\\ CNN High-Resolution\\ Network-based model }   &    \makecell{ The study shows how iterative transfer learning\\ may be used to transform pictures, which can\\ lessen the need on huge datasets. }    & \makecell{ The same as\cite{gilani2017table}, There are additional\\ phases in the pre-processing process.}  \\ \hline 
      
      Á Casado-García \cite{casado2020benefits}   &  \makecell{ Liken fine-tuning + \\ Mask R-CNN, RetinaNet\\, SSD and YOLO }   &    \makecell{ Describe the advantages of using object detection\\ networks in conjunction with domain-specific\\ fine-tuning techniques for table detection. }    & \makecell{ Closed domain fine-tuning is still\\ insufficient to get state-of-the-art\\ solutions. }  \\ \hline 
      
      M Agarwal \cite{agarwal2021cdec}   &  \makecell{ multistage extension of \\Mask R-CNN with a dual\\ backbone} & \makecell{ 1) A comprehensive object detection-based frame-\\work utilizing a composite backbone to deliver  state-\\of-the-art outcomes    2) Extensive tests on benchmark\\ datasets for table detection that are openly accessible.}   &    \makecell{ The technique is computationally \\expensive since it uses a composite \\backbone in addition to deformable\\ convolutions. }   \\ \hline

      X Zheng \cite{zheng2021global} &  \makecell{ Global Table Extractor \\(GTE) which is general \\method for object detection }   &    \makecell{ 1) The problem of table detection is benefited by the\\ extra piece-wise constraint loss introduced. 2) a\\ complete method that is compatible with all object \\detection frameworks. }    & \makecell{ Annotations for cellular borders are\\ necessary since the process of table \\detection depends on cell detection. }  \\ \hline

\end{tabular}
\end{adjustbox}
\label{tab:Table Detection Methods}
\end{table}

\subsection{Table Structure Recognition Models}
In order to recognize table structures in document images, deep learning approaches are reviewed in this part. We divided the methods into discrete deep-learning principles for the benefit of our readers. Table \ref{tab:Table Recognition1},\ref{tab:Table Recognition2} lists all methods for recognizing table structures based on object detection, as well as their benefits and drawbacks. It also discusses various deep learning-based methods that have been used in these methods.

A Zucker \cite{zucker2021clusti} presents CluSTi, a Clustering approach for recognizing the Structure of Tables in invoice scanned images, as an effective way. CluSTi makes three contributions. To begin, it uses a clustering approach to eliminate high noise from the table pictures. Second, it uses state-of-the-art text recognition to extract all text boxes. Finally, CluSTi organizes the text boxes into the correct rows and columns using a horizontal and vertical clustering technique with optimum parameters.  Z Zhang \cite{zhang2022split} present Split, Embed, and Merge (SEM) is a table structure recognizer that is accurate. M Namysl \cite{namysl2022flexible} presents a versatile and modular table extraction approach in this research.

E Koci \cite{koci2018table} offers a new method for identifying tables in spreadsheets and constructing layout areas after determining the layout role of each cell. Using a graph model, they express the spatial interrelationships between these areas. On this foundation, they present Remove and Conquer (RAC), a table recognition algorithm based on a set of carefully selected criteria.

Using the potential of deformable convolutional networks, SA Siddiqui \cite{siddiqui2019deeptabstr} proposes a unique approach for analyzing tabular patterns in document pictures. P Riba \cite{riba2019table} presents a graph-based technique for recognizing tables in document pictures in this paper. also employ the location, context, and content type instead of the raw content (recognized text), thus it's just a structural perception technique that's not reliant on the language or the quality of the text reading. E Koci \cite{koci2019genetic} use genetic-based techniques for graph partitioning, to recognize the sections of the graph matching to tables in the sheet.

SA Siddiqui \cite{siddiqui2019rethinking} described the structure recognition issue as the semantic segmentation issue. To segment the rows and columns, the authors employed fully convolutional networks. The approach of prediction tiling is introduced, which lessens the complexity of table structural identification, assuming consistency in a tabular structure. The author imported pre-trained models from ImageNet and used the structural models of FCN's encoder and decoder. The model creates features of the same size as the original input picture when given an image.

SA Khan \cite{khan2019table}  presents a robust deep learning-based solution for extracting rows and columns from a recognized table in document pictures in this work. The table pictures are pre-processed before being sent into a bi-directional Recurrent Neural Network using Gated Recurrent Units (GRU) and a fully-connected layer with softmax activation in the suggested solution. SF Rashid \cite{rashid2017table} provides a new learning-based approach for table content identification in diverse document pictures. SR Qasim \cite{qasim2019rethinking} presents a graph network-based architecture for table recognition as a superior alternative to typical neural networks. S Raja \cite{raja2020table} describes a method for recognizing table structure that combines cell detection and interaction modules to locate the cells and forecast their relationships with other detected cells in terms of row and column. Also, add structural limitations to the loss function for cell identification as extra differential components. The existing issues with end-to-end table identification were examined by Y Deng \cite{deng2019challenges}, who also highlighted the need for a larger dataset in this area.

Another study by Y Zou \cite{zou2020deep} called for the development of an image-based table structure identification technique using fully convolutional networks. the shown work divides a table's rows, columns, and cells. All of the table components' estimated bounds are enhanced using connected component analysis. Based on the placement of the row and column separators, row and column numbers are then allocated for each cell. In addition, special algorithms are used to optimize cellular borders.

To identify rows and columns in tables, KA Hashmi \cite{hashmi2021guided} suggested a guided technique for table structure identification. The localization of rows and columns may be made better, according to this study, by using an anchor optimization approach. The boundaries of rows and columns are detected in their proposed work using Mask R-CNN and optimized anchors.

Another effort to segment tabular structures is the ReS2TIM paper by W Xue \cite{xue2019res2tim} which describes the reconstruction of syntactic structures from the table. Regressing the coordinates for each cell is this model's main objective. A network that can identify the neighbors of each cell in a table is initially built using the new technique. In the study, a distance-based weighting system is given that will assist the network in overcoming the training-related class imbalance problem.

C Tensmeyer \cite{tensmeyer2019deep} has presented SPLERGE (Split and Merge), another method using dilated convolutions. Their strategy entails the use of two distinct deep learning models, the first of which establishes the grid-like layout of the table and the second of which determines if further cell spans over many rows or columns are possible. 

A Nassar \cite{nassar2022tableformer} provide a fresh identification model for table structures. The latter enhances the most recent encoder-dual-decoder from PubTabNet end-to-end deep learning model in two important aspects. First, the authors provide a brand-new table-cell object detection decoder. This allows them to easily access the content of the table cells in programmatic PDFs without having to train any proprietary OCR decoders.  The authors claim that this architectural improvement makes table-content extraction more precise and enables them to work with non-English tables. Second, transformer-based decoders take the place of LSTM decoders.

S Raja \cite{raja2022visual} suggests a novel object-detection-based deep model that is tailored for quick optimization and captures the natural alignments of cells inside tables. Dense table recognition may still be problematic even with precise cell detection because multi-row/column spanning cells make it difficult to capture long-range row/column relationships. Therefore, the authors also seek to enhance structure recognition by determining a unique rectilinear graph-based formulation. The author emphasizes the relevance of empty cells in a table from a semantics standpoint. The authors recommend a modification to a well-liked assessment criterion to take these cells into consideration. To stimulate fresh perspectives on the issue, then provide a moderately large assessment dataset with annotations that are modeled after human cognition.

X Shen \cite{shen2022rcanet} suggested two modules, referred to as Rows Aggregated (RA) and Columns Aggregated (CA). First, to produce a rough forecast for the rows and columns and address the issue of high error tolerance, feature slicing and tiling are applied. Second, the attention maps of the channels are computed to further obtain the row and column information. In order to complete the rows segmentation and columns segmentation, the authors employ RA and CA to construct a semantic segmentation network termed the Rows and Columns Aggregated Network (RCANet).

C Ma\cite{ma2022robust} present RobusTabNet, a novel method for recognizing the structure of tables and detecting their borders from a variety of document pictures. The authors suggest using CornerNet as a new region proposal network to produce higher quality table proposals for Faster R-CNN, which has greatly increased the localization accuracy of Faster R-CNN for table identification.  by utilizing only the minimal ResNet-18 backbone network. Additionally, the authors suggest a brand-new split-and-merge approach for recognizing table structures. In this method, each detected table is divided into a grid of cells using a novel spatial CNN separation line prediction module, and then a Grid CNN cell merging module is used to recover the spanning cells. Their table structure recognizer can accurately identify tables with significant blank areas and geometrically deformed (even curved) tables because the spatial CNN module can efficiently transmit contextual information throughout the whole table picture. B Xiao \cite{xiao2022table} postulates that a complex table structure may be represented by a graph, where the vertices and edges stand in for individual cells and the connections between them. Then, the authors design a conditional attention network and characterize the table structure identification issue as a cell association classification problem (CATT-Net). 

A Jain \cite{jain2022tsr} suggests training a deep network to recognize the spatial relationships between various word pairs included in the table picture in order to decipher the table structure. The authors offer an end-to-end pipeline called TSR-DSAW: TSR through Deep Spatial Association of Words, which generates a digital representation of a table picture in a structured format like HTML. The suggested technique starts by utilizing a text-detection network, such as CRAFT, to identify every word in the input table picture. Next, using dynamic programming, word pairings are created. These word pairings are underlined in each individual image and then given to a DenseNet-121 classifier that has been trained to recognize spatial correlations like same-row, same-column, same-cell, or none.  Finally, The authors apply post-processing to the classifier output in order to produce the HTML table structure.

H Li \cite{li2022two} formulate the issue as a cell relation extraction challenge and provide T2, a cutting-edge two-phase method that successfully extracts table structures from digitally preserved texts. T2 offers a broad idea known as a prime connection that accurately represents the direct relationships between cells. To find complicated table structures, it also builds an alignment graph and uses a message-passing network.

\begin{table}[h!]
\caption{A comparison of the benefits and drawbacks of several deep learning-based table Structure recognition methods}
\begin{adjustbox}{width=1\textwidth}
\begin{tabular}{|c|c|c|c|} 
    \hline
      Literature & Method & Benefits & Drawbacks  \\ \hline
      
      SF Rashid \cite{rashid2017table}  &  \makecell{Uses the geometric pos-\\ition of words + A neu-\\ral network model \\(autoMLP)}  &    \makecell{No reliance on complex layout analysis\\ Mechanism. Can be used on the diverse set\\ of documents with different layouts }  &    \makecell{limitation is in marking columns\\ boundaries due to variations in\\ the number of words in each\\ column } \\ \hline
      
      E Koci \cite{koci2018table} &  \makecell{Encoding of spatial inter-\\relations between these\\ regions using a graph rep-\\resentation, as well as\\ rules and heuristics }  &    \makecell{1) Recognition for single-table and multi-\\table spreadsheets. 2) No reliance on any \\assumptions with what regards the arran-\\gement of tables }  &    \makecell{Tables with few columns and \\empty cells are not handled\\ well. } \\ \hline
      
      SA Siddiqui \cite{siddiqui2019deeptabstr}     & \makecell{deformable CNN  \\+ Faster R-CNN}     &     \makecell{ 1) The use of deformable convolution can \\handle various tabular structures. 2) rel-\\eased a new dataset that contained table \\structure data. } & \makecell{The tables in the proposed\\ approach won't operate co-\\rrectly if they have a row a-\\nd column span. }\\ \hline
       
      SA Siddiqui \cite{siddiqui2019rethinking}   &  \makecell{ Fully CNNs }   &    \makecell{ The complexity of the task of identifying t-\\able structures is reduced by the proposed\\ prediction tiling approach.}    & \makecell{ 1) Additional post-processing p-\\rocesses are necessary when ro-\\ws or columns are excessively \\fragmented. 2) The technique is \\based on the tabular structures' \\consistency assumption. }  \\ \hline 
      
      SR Qasim \cite{qasim2019rethinking}   &  \makecell{ Graph NN + CNN  }   &    \makecell{ 1) This paper also presents a unique, memory-\\efficient training strategy based on Monte Ca-\\rlo. 2) The suggested approach makes use of \\both textual and spatial characteristics.} & \makecell{ The publicly accessible table \\datasets are not used to test\\ the system. }  \\ \hline 
      
      W Xue \cite{xue2019res2tim}   &  \makecell{Graph NN + weights\\ depending on distance }   &    \makecell{ For the cell relationship network, the class\\ imbalance issue is solved using the distance-\\based weighting method. }    & \makecell{ When dealing with sparse tab-\\les, the approach is insecure. }  \\ \hline 
      
      C Tensmeyer \cite{tensmeyer2019deep}  &  \makecell{Dilated Convolutions \\+ Fully CNN }   &    \makecell{ The technique is effective with both scanned \\and PDF document images. }    & \makecell{ The post-processing heuristics\\ determine how the merging p-\\ortion of the method works. }  \\ \hline 
      
      SA Khan \cite{khan2019table}  &  \makecell{ RNN }   &    \makecell{ The reduced receptive field of CNNs is solved\\ by the bi-directional GRU. }    & \makecell{ Pre-processing procedures including\\ binarization, noise reduction, and \\morphological modification are ne-\\cessary. }  \\ \hline 
      
      P Riba \cite{riba2019table} &  \makecell{Graph Neural Networks \\approach }  &    \makecell{ 1) It is not constrained to rigid tabular lay-\\outs in terms of single rows, columns or pr-\\esence of rule lines. 2) The model is langua-\\ge independent }  &    \makecell{1) The method may have problems\\ when dealing with border cond-\\itions. 2) There is a small amount\\ of training data in the RVL-CDIP\\ dataset and F1, Precision and Re-\\call metrics are lower than other\\ methods. } \\ \hline

      Y Deng \cite{deng2019challenges}  &  \makecell{ Encoder decoder net }   &    \makecell{1) In the work that is given, issues with end-\\to-end table recognition are examined. 2) Mad\\-e a contribution with yet another sizable data\\-set in the area of table comprehension. }    & \makecell{ The other publicly accessible \\table recognition datasets are not\\ used to assess the suggested base-\\line technique. }  \\ \hline 
      
      E Koci \cite{koci2019genetic} &  \makecell{Graph model + Appl-\\ication of genetic-based \\approaches }  &    \makecell{Requires little to no involvement of domain \\experts}  &    \makecell{The accuracy of GE depends on\\ the number of edges. Specifica-\\lly, we determined that GE ach-\\ieves an accuracy of only 19\% \\for multi-table graphs } \\ \hline
      
      D Prasad \cite{prasad2020cascadetabnet}   &  \makecell{ Cascade mask Reg-\\ionbased CNN High-\\Resolution Network-\\based model }   &    \makecell{ Direct regression occurs at cellular bound-\\aries using an end-to-end method.  }    & \makecell{ Tables with/out ruling  lin-\\es must undergo further \\post-processing.}  \\ \hline 
      
      S Raja \cite{raja2020table}        &  \makecell{Mask R-CNN +  \\ResNet-101 based\\ Net}   &    \makecell{1) An additional alignment loss is sugges-\\ted for precise cell detection. 2) A train-\\able top-down for cell identification and \\bottom-up for structure recognition coll-\\ection is proposed. } & \makecell{When cells are empty, the \\strategy is weak. }\\ \hline
      
\end{tabular}
\end{adjustbox}
\label{tab:Table Recognition1}
\end{table}

\begin{table}[h!]
\caption{A comparison of the benefits and drawbacks of several deep learning-based table Structure recognition methods (continue Table \ref{tab:Table Recognition1})}
\begin{adjustbox}{width=1\textwidth}
\begin{tabular}{|c|c|c|c|} 
    \hline
      Literature & Method & Benefits & Drawbacks  \\ \hline

       B Xiao \cite{xiao2022table} &  \makecell{cells’ bounding boxes\\ + conditional attention \\network}  &    \makecell{Only utilizes visual features without any meta-\\data }  &    \makecell{1) Assumes that the coordinates \\of cells in the table are known.\\ 2) Difficulties with tables with-\\out borders} \\ \hline
      
      Y Zou \cite{zou2020deep}   &  \makecell{ Fully CNNs }   &    \makecell{1) Using linked component analysis enhances\\ the outcomes. 2) In a table, cells are segmen-\\ted in addition to the rows and columns. }    & \makecell{To provide comparison findings,\\ a small number of post-process-\\ing procedures utilizing specific\\ algorithms are necessary. }  \\ \hline 
      
      X Zhong \cite{zhong2020image}  &  \makecell{ Dual decoder with\\ attention-based enc-\\oding }   &    \makecell{  1) To assess table recognition techniques, the \\methodology offers a unique evaluation metric \\called TEDS. 2) released a huge table dataset.  }    & \makecell{ The technique cannot be readily\\ compared to other state-of-the-\\art techniques. }  \\ \hline 
      
      KA Hashmi \cite{hashmi2021guided}   &  \makecell{ Utilizing an optimi-\\zation technique for \\anchors+  Mask R-\\CNN }   &    \makecell{ Networks of region proposals converge mo-\\re quickly and effectively thanks to optim-\\ized anchoring. }    & \makecell{ This study relies on the pre\\liminary pre-processing phase\\ of clustering the ground truth\\ to find appropriate anchors. }  \\ \hline 
      
      A Zucker \cite{zucker2021clusti} &  \makecell{Character Region Awareness \\for Text Detection (CRAFT)\\ and Density-Based Spatial \\Clustering of Applications\\ with Noise (DBSCAN) }  &    \makecell{A bottom-up method, which emphasizes that\\ the table structure is formed by relative pos-\\itions of text cells, and not by inherent bou-\\ndaries }  &    \makecell{Cannot handle spreading rows \\or columns well } \\ \hline

      X Zheng \cite{zheng2021global}   &  \makecell{ Method for object \\detecting generally }   &    \makecell{ An additional innovative cluster-based tech\\nique combined with a hierarchical network\\ to detect tabular forms. }    & \makecell{ Accurately classifying a table\\ is a prerequisite for final cell\\ structure identification. }  \\ \hline

      Z Zhang \cite{zhang2022split} &  \makecell{A combination of fully convo-\\lutional network (FCN)+ RoI-\\Align  + the pretrained BERT \\model +  Gated Recurrent Unit \\(GRU) decoder }  &    \makecell{Directly operates on table images with no \\dependency on meta-information, can pro-\\cess simple and complex tables}  &    \makecell{Oversegments tables when space\\ between cells is large, doesn’t\\ handle merged cells well} \\ \hline
      
       M Namysl \cite{namysl2022flexible} &  \makecell{Rule-based algorithms +\\ graph-based table inter-\\pretation method }  &    \makecell{1) Approach allows processing images and \\digital documents. 2) Processing steps can \\be adapted separately }  &    \makecell{ 1) Support the most frequent ta-\\ble formats only. Reliance on the\\ presence of predefined keywords. \\2) Prone to the errors propagated\\ from the upstream components\\ of system. 3) Focus on the ta-\\bles with rulings} \\ \hline
       
       A Nassar \cite{nassar2022tableformer} &  \makecell{End-to-end neural network\\  + CNN Backbone + tran-\\sformer based layers }  &    \makecell{1) Handles different languages without being\\ trained on them. 2) Predicts tables structure\\ and bounding boxes for the table content }  &    \makecell{Work with PDF documents} \\ \hline
       
       A Jain \cite{jain2022tsr}  &  \makecell{spatial associations +  dyna-\\mic programming techniques}  &    \makecell{Recognizing complex table structures having\\ multi-span rows/columns and missing cells }  &    \makecell{Uses OCR to read words from\\ images Not language agnostic } \\ \hline
       
       S Raja \cite{raja2022visual} &  \makecell{object detection}  &    \makecell{Better detection of empty cells}  &    \makecell{Fails for very sparse tables wh-\\ere most of the cells are empty} \\ \hline
       
    \end{tabular}
\end{adjustbox}
\label{tab:Table Recognition2}
\end{table}


\section{Methodology}
\label{sec:methodology}
In this section, we will extend the methodology and methods used for the TNCR dataset in our previous work \cite{abdallah2022tncr}. in the previous work, we described Cascade R-CNN, Cascade Mask R-CNN, Cascade RPN, Hybrid Task Cascade (HTC), YOLO, and Deformable DETR. In this section, we will describe additional four methodologies of using object detection and classification for Faster R-CNN, Mask R-CNN, HRNets, Resnest, and Dynamic R-CNN. 

\subsection{Faster R-CNN}
\begin{figure}[ht!]
      \centering
      \includegraphics[width=\textwidth,height=.60\textwidth]{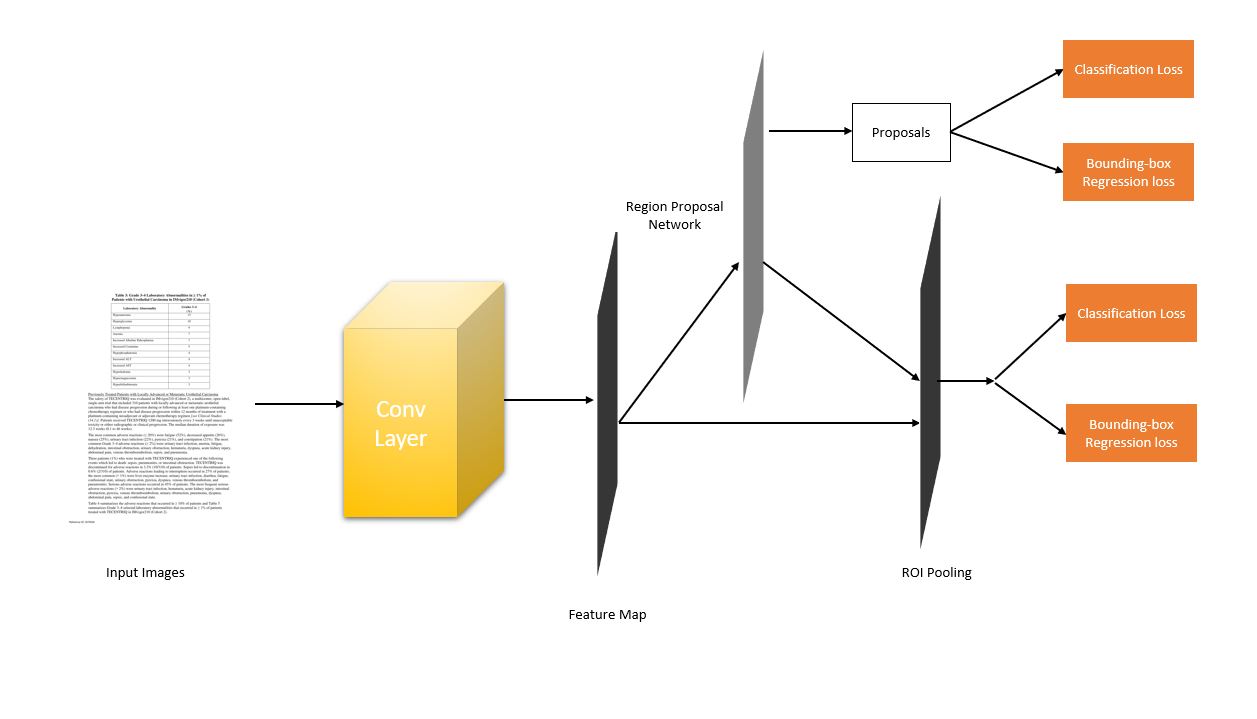}
      \caption{Faster R-CNN}
      \label{fig:Faster-CNN}
\end{figure}

Faster R-CNN \cite{ren2015faster} contains two modules: The RPN is a fully-convolutional network that generates region proposals, and the Fast-RCNN detector takes the proposal from RPN as input and generates object detection results as seen in Fig. \ref{fig:Faster-CNN}.  A feature extraction network, which is often a pretrained CNN, is employed in a Faster R-CNN object detection network, similar to what we utilized for its predecessor. Following that, there are two trainable subnetworks. The first is a Region Proposal Network (RPN), which is used to produce object proposals as its name indicates, and the second is used to predict the object's real class. The RPN that is put after the last convolutional layer is thus the major differentiator for Faster R-CNN. This has been taught to generate region proposals without the use of any external mechanisms such as Selective Search. Then, similar to Fast R-CNN, utilize ROI pooling, an upstream classifier, and a bounding box regressor.

\subsection{Mask R-CNN}
\begin{figure}[ht!]
      \centering
      \includegraphics[width=\textwidth,height=.60\textwidth]{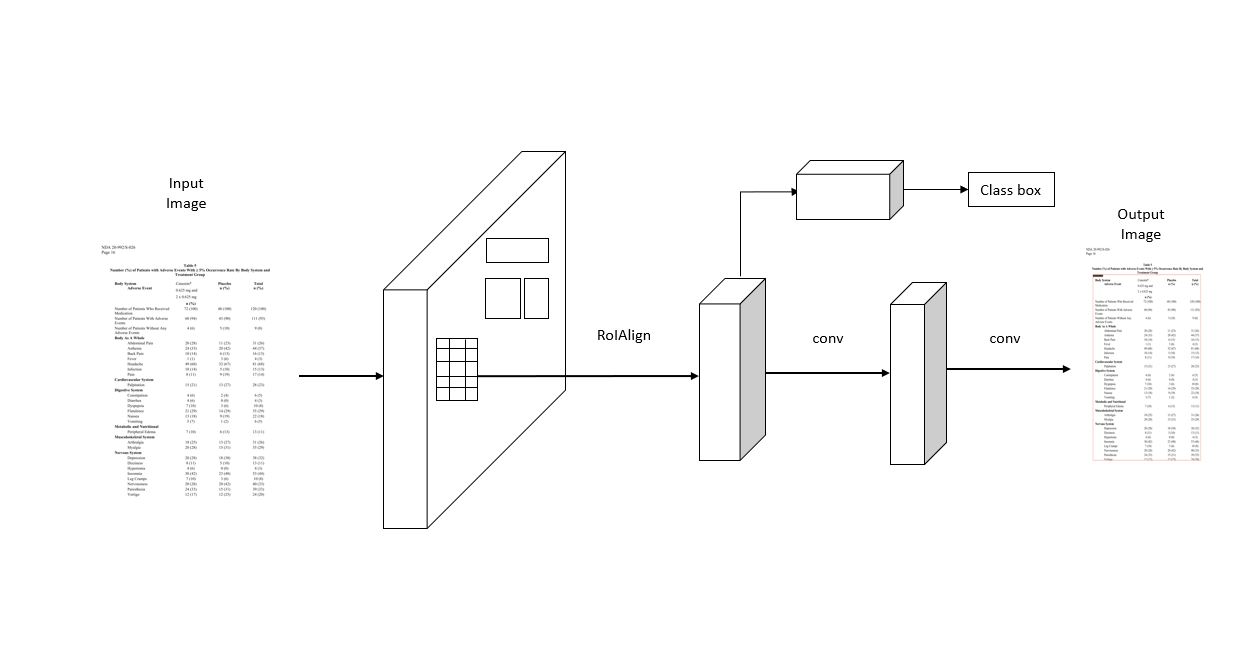}
      \caption{Mask R-CNN}
      \label{fig:Mask-RCNN}
\end{figure}

Mask R-CNN \cite{He_2017} uses R-CNN to effectively detect objects in an image while also performing object segmentation tasks for each region of interest. As a result, segmentation runs concurrently with classification and bounding box regression. The high-level architecture of the Mask R-CNN is shown in Fig. \ref{fig:Mask-RCNN}. 
Mask RCNN is divided into two phases. First, it generates proposals based on the input image for regions where an object might be present. Second, based on the first stage proposals, it predicts the object's class, refines the bounding box, and creates a mask at the pixel level of the object. The backbone structure is related to both phases. The concept of Mask R-CNN is simple: Faster R-CNN outputs a class label and a bounding-box offset for each candidate object; Mask R-CNN adds a third branch that outputs the object mask.

\subsection{HRNets}

\begin{figure}[ht!]
      \centering
      \includegraphics[width=\textwidth,height=.60\textwidth]{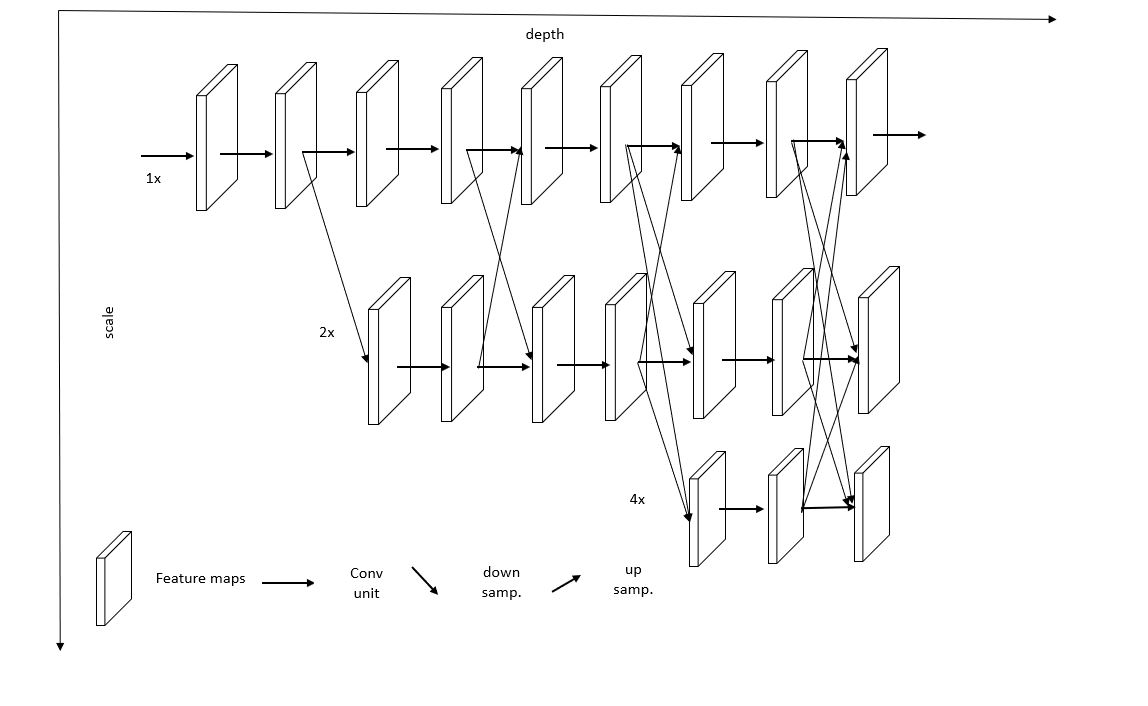}
      \caption{HRNet}
      \label{fig:HRNet}
\end{figure}

Ke Sun et at,\cite{SunXLW19,SunZJCXLMWLW19} present a novel architecture called High-Resolution Net, which is capable of maintaining high-resolution representations through the entire process. The first stage of the HRNet project is to build a high-resolution subnetwork. The next stage is to add more high-to-low resolutions subnetworks. The multi-scale fusions are carried out by HRNet through a parallel multi-resolution network as seen in Fig. \ref{fig:HRNet}.

Compared to existing widely-used networks \cite{newell2016stacked,insafutdinov2016deepercut,xiao2018simple,yang2017learning}, HRNet has two advantages: it can connect high-to-low-resolution subnetworks in parallel, and it can provide better pose estimation.
Most existing fusion schemes combine low-resolution and high-resolution representations. Instead of doing so, HRNet performs multiscale fusions to boost both the high-resolution and low-resolution representations.

\subsection{Resnest}

\begin{figure}[ht!]
      \centering
      \includegraphics[width=\textwidth,height=.60\textwidth]{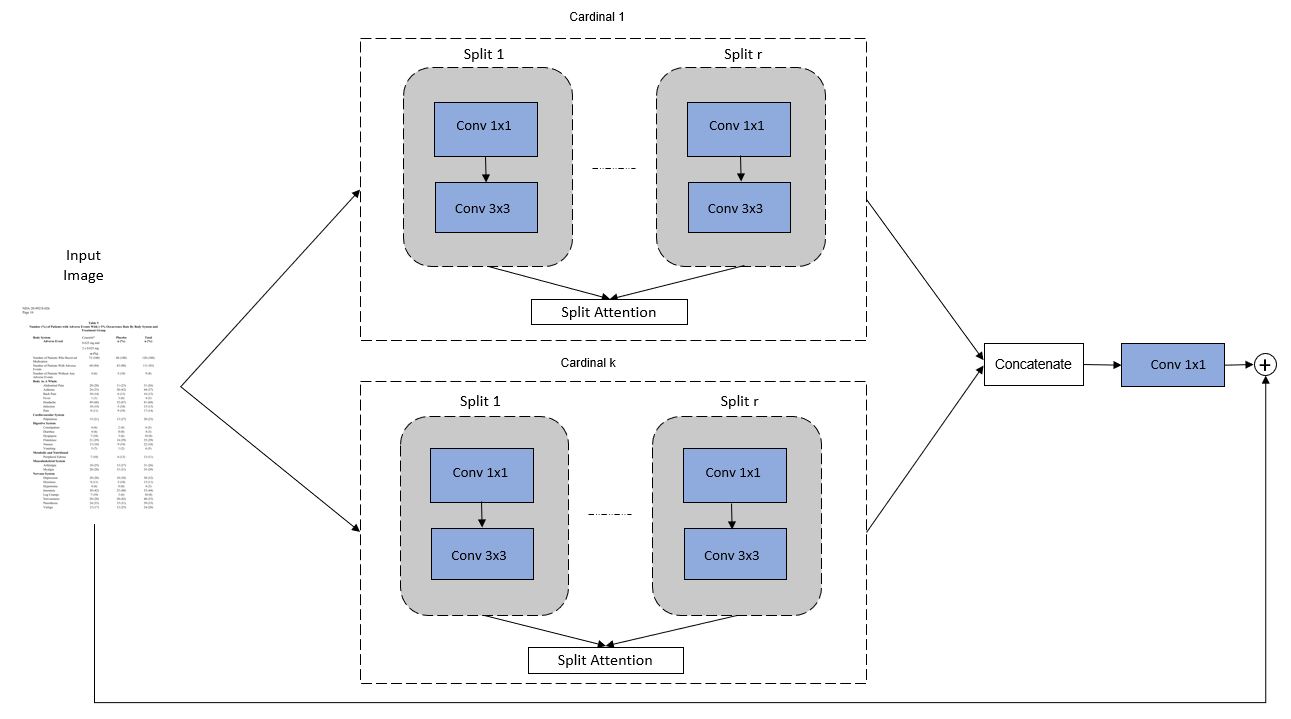}
      \caption{Resnest}
      \label{fig:Resnest}
\end{figure}

Resnest \cite{zhang2020resnest} is a simple architecture that combines the features of a multipath network with a channel-wise attention strategy. It allows  for the preservation of independent representations in the meta structure. As in a multi-path network, a Resnet network module performs a set of transformations on low-dimensional embeddings and concatenates their outputs. Each transformation is carried out with a different attention strategy to capture the interdependencies of a feature map. The key difference between the two is that the attention strategy is focused on the specific channel and not the whole network. The Split-Attention block is a computing unit that combines feature map group and split attention operations. A Split-Attention Block is depicted in Fig. \ref{fig:Resnest}.

\subsection{Dynamic R-CNN }

Hongkai Zhang et al, \cite{DynamicRCNN} proposes Dynamic RCNN, a simple but effective method for maximizing the dynamic quality of object detection proposals. It is made up of two parts: Dynamic Label Assignment and Dynamic SmoothL1 Loss, which are used for classification and regression, respectively. First, adjust the IoU threshold for positive/negative samples based on the proposals distribution in the training procedure to train a better classifier that is discriminative for high IoU proposals. 
Set the threshold as the proposal's IoU at a certain percentage because this can reflect the overall distribution's quality. Change the shape of the regression loss function for regression to adaptively fit the regression label distribution change and ensure the contribution of high-quality samples to training.

\section{Experiments Results}
\label{Experiments_Results}

\subsection{Experiment Settings}

The MMdetection \cite{chen2019mmdetection} library for PyTorch has been used to implement each of the proposed and tested models. A vast variety of object detection and instance segmentation methods, as well as associated parts and modules, are included in the object detection toolkit known as MMDetection. It gradually transforms into a single platform that includes several widely used detection techniques and contemporary modules. Three Tesla V100-SXM GPUs with 16 GB of GPU memory, 16 GB of RAM, two Intel Xeon E-5-2680 CPUs, and four NVIDIA Tesla k20x GPUs were used in the trials, which were carried out on the Google Colaboratory platform. With pictures scaled to a constant size of 1300 $\times$ 1500 and a batch size of 16, all models have been trained and evaluated. The optimizer with a momentum of 0.9, a weight decay of 0.0001, and a learning rate of 0.02 is known as SGD. The Feature Pyramid Network (FPN) neck is used by all models.                                                                                                                                 

\subsection{Results of TNCR dataset }

The Faster R-CNN model has achieved good performance in table detection compared with Cascade-RCNN and Cascade Mask-RCNN in most of the backbones. We have trained the Faster R-CNN model with L1 Loss \cite{wu2019iou} with Resnet-50 for bounding box regression. As shown in Table \ref{tab:FasterR-CNN}, it achieves f1-score of 0.921. Resnet-101 backbone achieves the highest F1 score over 50\% to 65\%, ResNeXt-101-64x4d  achieves the highest F1 score over 70\% to 95\% and ResNeXt-101-64x4d achieves the highest F1 score over 50\%:95\% of 0.786. Resnet-50 backbone with 1$\times$ Lr schedule achieves the lowest performance over 50\% to 60\% IoUs. Also, the Resnet-50 backbone with L1 Los achieves the lowest performance from 65\% to 95\% IoUs and also achieves the lowest performance over 50\%:95\%.
\begin{table}[h!]
\caption{Faster R-CNN}
\begin{center}
\begin{adjustbox}{width=1\textwidth}
\begin{tabular}{|c|c|c|c|c|c|c|c|c|c|c|c|c|c|}
    \hline
    \multirow{2}{*}{Backbone}& \multirow{2}{*}{Lr schd/Losses}& \multirow{2}{*}{} & \multicolumn{11}{|c|}{ IoU } \\
    \cline{4-14}
     & & & 50\% & 55\% & 60\% & 65\% & 70\% & 75\% & 80\% & 85\% & 90\% & 95\% & 50\%:95\% \\
    
    \hline
    \multirow{3}{*}{Resnet-50}  & \multirow{3}{*}{L1Loss} 
    & Precision   & 0.875 & 0.872 & 0.872 & 0.866 & 0.858 & 0.844 & 0.823 & 0.782 & 0.688 & 0.424 & 0.649 \\
    & & Recall    & 0.973 & 0.972 & 0.970 & 0.964 & 0.956 & 0.941 & 0.922 & 0.890 & 0.812 & 0.577 & 0.775 \\
    & & F1-Score  & 0.921 & 0.919 & 0.918 & \textbf{0.912}* & \textbf{0.904}* & \textbf{0.889}* & \textbf{0.869}* & \textbf{0.832}* & \textbf{0.744}* & \textbf{0.488}* & \textbf{0.706}* \\
    
    \hline
    \multirow{3}{*}{Resnet-50}  & \multirow{3}{*}{1x}     
    & Precision   & 0.874 & 0.872 & 0.871 & 0.869 & 0.863 & 0.844 & 0.827 & 0.783 & 0.693 & 0.431 & 0.653 \\
    & & Recall    & 0.972 & 0.969 & 0.968 & 0.966 & 0.959 & 0.942 & 0.929 & 0.895 & 0.823 & 0.587 & 0.779 \\
    & & F1-Score  & \textbf{0.920}* & \textbf{0.917}* & \textbf{0.916}* & 0.914 & 0.908 & 0.890 & 0.875 & 0.835 & 0.752 & 0.497 & 0.710 \\

    \hline
    \multirow{3}{*}{Resnet-101}  & \multirow{3}{*}{1x} 
    & Precision   & 0.885 & 0.885 & 0.882 & 0.879 & 0.870 & 0.867 & 0.849 & 0.820 & 0.763 & 0.555 & 0.720 \\
    & & Recall    & 0.973 & 0.973 & 0.971 & 0.969 & 0.961 & 0.956 & 0.943 & 0.920 & 0.870 & 0.698 & 0.835 \\
    & & F1-Score  & \textbf{0.926} & \textbf{0.926} & \textbf{0.924} & \textbf{0.921} & 0.913 & 0.909 & 0.893 & 0.867 & 0.812 &  0.618 & 0.773 \\

    \hline
    \multirow{3}{*}{ResNeXt-101-32x4d}  & \multirow{3}{*}{1x}     
    & Precision   & 0.880 & 0.879 & 0.877 & 0.875 & 0.872 & 0.866 & 0.845 & 0.817 & 0.760 & 0.575 & 0.727 \\
    & & Recall    & 0.976 & 0.976 & 0.975 & 0.972 & 0.969 & 0.962 & 0.944 & 0.921 & 0.871 & 0.711 & 0.843 \\
    & & F1-Score  & 0.925 & 0.924 & 0.923 & 0.920 & 0.917 & 0.911 & 0.891 & 0.865 & 0.811 & 0.635 & 0.780 \\
    
    \hline
    \multirow{3}{*}{ResNeXt-101-64x4d}  & \multirow{3}{*}{1x}     
    & Precision   & 0.884 & 0.884 & 0.880 & 0.879 & 0.876 & 0.871 & 0.856 & 0.833 & 0.780 & 0.581 & 0.733 \\
    & & Recall    & 0.972 & 0.970 & 0.969 & 0.967 & 0.965 & 0.961 & 0.950 & .931 & 0.884 & 0.724 &  0.848\\
    & & F1-Score  & 0.925 & 0.925 & 0.922 & 0.920 & \textbf{0.918} & \textbf{0.913} & \textbf{0.900} & \textbf{0.879} & \textbf{0.828} & \textbf{0.644} & \textbf{0.786} \\

    \hline
\end{tabular}
\end{adjustbox}
\end{center}
\label{tab:FasterR-CNN}
\end{table}

We implemented Mask R-CNN \cite{He_2017} to use R-CNN for table objects in an image and also used for performing object segmentation for each ROI. As seen in Table \ref{tab:Mask R-CNN}, Mask R-CNN shows good performance in our dataset in precision, recall, and F1 score for all backbones. Resnet-101 backbone has achieved the highest F1 score of 0.774 over 50\%:95\% and maintains the highest F1 score at various IoUs. ResNeXt-101-32x4d achieves the lowest performance over 50\% to 95\% IoUs and also achieves an f1 score of 0.512 over 50\%:95\%. ResNeXt-101-64x4d also achieves the lowest performance at various IoUs except for 95\% IoU.
\begin{table}[h!]
\caption{Mask R-CNN}
\begin{center}
\begin{adjustbox}{width=1\textwidth}
\begin{tabular}{|c|c|c|c|c|c|c|c|c|c|c|c|c|c|}
    \hline
    \multirow{2}{*}{Backbone}& \multirow{2}{*}{Lr schd}& \multirow{2}{*}{} & \multicolumn{11}{|c|}{ IoU } \\
    \cline{4-14}
     & & & 50\% & 55\% & 60\% & 65\% & 70\% & 75\% & 80\% & 85\% & 90\% & 95\% & 50\%:95\% \\
    
    \hline
    \multirow{3}{*}{Resnet-50}  & \multirow{3}{*}{1x} 
    & Precision   & 0.877 & 0.876 & 0.874 & 0.871 & 0.868 & 0.858 & 0.834 & 0.800 & 0.728 & 0.506 & 0.692 \\
    & & Recall    & 0.973 & 0.972 & 0.970 & 0.967 & 0.963 & 0.952 & 0.932 & 0.903 & 0.840 & 0.651 & 0.812 \\
    & & F1-Score  & 0.922 & 0.921 & 0.919 & 0.916 & 0.913 & 0.902 & 0.880 & 0.848 & 0.779 & 0.569 & 0.747 \\
    
    \hline
    \multirow{3}{*}{Resnet-101}  & \multirow{3}{*}{1x} 
    & Precision   & 0.878 & 0.877 & 0.875 & 0.874 & 0.869 & 0.861 & 0.847 & 0.812 & 0.762 & 0.553 & 0.716 \\
    & & Recall    & 0.977 & 0.976 & 0.974 & 0.973 & 0.966 & 0.959 & 0.949 & 0.918 & 0.874 & 0.711 & 0.844 \\
    & & F1-Score  & \textbf{0.924} & \textbf{0.923} & \textbf{0.921} & \textbf{0.920} & \textbf{0.914} & \textbf{0.907} & \textbf{0.895} & \textbf{0.861} & \textbf{0.814} & \textbf{0.622} & \textbf{0.774} \\

    \hline
    \multirow{3}{*}{ResNeXt-101-32x4d}  & \multirow{3}{*}{1x} 
    & Precision   & 0.778 & 0.777 & 0.774 & 0.769 & 0.759 & 0.749 & 0.713 & 0.651 & 0.477 & 0.407 & 0.434 \\
    & & Recall    & 0.975 & 0.974 & 0.968 & 0.964 & 0.952 & 0.941 & 0.913 & 0.856 & 0.725 & 0.695 & 0.626 \\
    & & F1-Score  & \textbf{0.865}* & \textbf{0.864}* & \textbf{0.860}* & \textbf{0.855}* & \textbf{0.844}* & \textbf{0.834}* & \textbf{0.800}* & \textbf{0.739}* & \textbf{0.575}* & \textbf{0.513}* & \textbf{0.512}* \\

    \hline
    \multirow{3}{*}{ResNeXt-101-64x4d}  & \multirow{3}{*}{1x} 
    & Precision   & 0.778 & 0.777 & 0.774 & 0.769 & 0.759 & 0.749 & 0.713 & 0.651 & 0.477 & 0.417 & 0.434 \\
    & & Recall    & 0.975 & 0.974 & 0.968 & 0.964 & 0.952 & 0.941 & 0.913 & 0.856 & 0.725 & 0.705 & 0.626\\
    & & F1-Score  & 0.865 & 0.864 & 0.860 & 0.855 & 0.844 & 0.834 & 0.800 & 0.739 & 0.575 & 0.524 & 0.512 \\

    \hline
    
\end{tabular}
\end{adjustbox}
\end{center}
\label{tab:Mask R-CNN}
\end{table}

In the following  \cref{tab:HRNets-CascadeR-CNN,tab:HRNets-Faster R-CNN,tab:HRNets-HTC,tab:hrnets-Mask R-CNN,tab:hrnet-CascadeMaskR-CNN,tab:hrnets-FCOS} show the comparative analysis we have trained HRNets with different methods and each method with different backbones for  object detection and instance segmentation models. In Table \ref{tab:HRNets-CascadeR-CNN}, we evaluated and calculated f1-score. It shows that HRNetV2p-W40 achieves better performance over 50\% to 65\%. Also, HRNetV2p-W18  achieves better performance over 70\%, 75\%, 80\%, and  95\%. HRNetV2p-W18  achieves an f1 score of 0.842 over 50\%:95\% IoU. HRNetV2p-W32 backbone achieves the lowest performance over 50\% to 70\%, and 90\% IoUs and HRNetV2p-W40 achieve the lowest performance over 75\% to 85\% and 95\%. HRNetV2p-W40  achieves f1 score of 0.841 over 50\%:95\% IoU.

\begin{table}[h!]
\caption{HRNets - Cascade R-CNN}
\begin{center}
\begin{adjustbox}{width=1\textwidth}
\begin{tabular}{|c|c|c|c|c|c|c|c|c|c|c|c|c|c|}
    \hline
    \multirow{2}{*}{Backbone}& \multirow{2}{*}{Lr schd}& \multirow{2}{*}{} & \multicolumn{11}{|c|}{ IoU } \\
    \cline{4-14}
     & & & 50\% & 55\% & 60\% & 65\% & 70\% & 75\% & 80\% & 85\% & 90\% & 95\% & 50\%:95\% \\
    
    \hline
    \multirow{3}{*}{HRNetV2p-W18}  & \multirow{3}{*}{20e} 
    & Precision   & 0.894 & 0.894 & 0.894 & 0.892 & 0.892 & 0.886 & 0.880 & 0.862 & 0.825 & 0.712 & 0.803 \\
    & & Recall    & 0.962 & 0.962 & 0.960 & 0.960 & 0.960 & 0.954 & 0.950 & 0.937 & 0.906 & 0.813 & 0.887 \\
    & & F1-Score  & 0.926 & 0.926 & 0.925 & 0.924 & \textbf{0.924} & \textbf{0.918} & \textbf{0.913} & 0.897 & 0.863 & \textbf{0.759} & \textbf{0.842} \\
  
    \hline
    \multirow{3}{*}{HRNetV2p-W32}  & \multirow{3}{*}{20e} 
    & Precision   & 0.895 & 0.895 & 0.893 & 0.893 & 0.893 & 0.889 & 0.881 & 0.869 & 0.828 & 0.717 & 0.806 \\
    & & Recall    & 0.955 & 0.955 & 0.954 & 0.954 & 0.953 & 0.949 & 0.943 & 0.933 & 0.900 & 0.810 & 0.882 \\
    & & F1-Score  & \textbf{0.924}* & \textbf{0.924}* & \textbf{0.922}* & \textbf{0.922}* & \textbf{0.922}* & 0.918 & 0.910 & \textbf{0.899} & \textbf{0.862}* & 0.760 & 0.842 \\

    \hline
    \multirow{3}{*}{HRNetV2p-W40}  & \multirow{3}{*}{20e} 
    & Precision   & 0.893 & 0.891 & 0.891 & 0.891 & 0.888 & 0.880 & 0.871 & 0.854 & 0.831 & 0.705 & 0.799 \\
    & & Recall    & 0.967 & 0.965 & 0.965 & 0.964 & 0.961 & 0.956 & 0.948 & 0.935 & 0.914 & 0.811 & 0.889 \\
    & & F1-Score  & \textbf{0.928} & \textbf{0.926} & \textbf{0.926} & \textbf{0.926} & 0.923 & \textbf{0.916}* & \textbf{0.907}* & \textbf{0.892}* & \textbf{0.870} &  \textbf{0.754}* & \textbf{0.841}* \\
    \hline

\end{tabular}
\end{adjustbox}
\end{center}
\label{tab:HRNets-CascadeR-CNN}
\end{table}

Table \ref{tab:HRNets-Faster R-CNN} shows the performance of the HRNets Faster R-CNN detector with various backbone structures with combinations of Lr Schedule. The HRNetV2p-W18 with 1$\times$  Lr Schedule backbone shows a low performance compared with other backbones. it achieves an f1 score of 0.770. It achieves 3.2\% less than HRNetV2p-W18 with 2$\times$  Lr Schedule. HRNetV2p-W40 with 1$\times$  Lr Schedule backbone achieves better performance over 50\% to 85\% IoUs and HRNetV2p-W40 with 2$\times$  Lr Schedule backbone achieves better performance over 90\% and 95\% IoUs. HRNetV2p-W18 with 2$\times$  Lr Schedule backbone achieves an f1 score of 0.802 over 50\%:95\%.  HRNetV2p-W32 with 1$\times$  Lr Schedule backbone share same performance over 50\% to 60\%.
\begin{table}[h!]
\caption{HRNets - Faster R-CNN}
\begin{center}
\begin{adjustbox}{width=1\textwidth}
\begin{tabular}{|c|c|c|c|c|c|c|c|c|c|c|c|c|c|}
    \hline
    \multirow{2}{*}{Backbone}& \multirow{2}{*}{Lr schd}& \multirow{2}{*}{} & \multicolumn{11}{|c|}{ IoU } \\
    \cline{4-14}
     & & & 50\% & 55\% & 60\% & 65\% & 70\% & 75\% & 80\% & 85\% & 90\% & 95\% & 50\%:95\% \\
    
    \hline
    \multirow{3}{*}{HRNetV2p-W18 }  & \multirow{3}{*}{1x} 
    & Precision   & 0.867 & 0.865 & 0.863 & 0.859 & 0.853 & 0.845 & 0.827 & 0.806 & 0.750 & 0.556 & 0.711 \\
    & & Recall    & 0.972 & 0.970 & 0.968 & 0.964 & 0.959 & 0.952 & 0.940 & 0.915 & 0.869 & 0.711 & 0.842 \\
    & & F1-Score  & \textbf{0.916}* & \textbf{0.914}* & \textbf{0.912}* & \textbf{0.908}* & \textbf{0.902}* & \textbf{0.895}* & \textbf{0.879}* & \textbf{0.857}* & \textbf{0.805}* & \textbf{0.624}* &  \textbf{0.770}* \\

    \hline
    \multirow{3}{*}{HRNetV2p-W18}  & \multirow{3}{*}{2x} 
    & Precision   & 0.876 & 0.873 & 0.872 & 0.869 & 0.867 & 0.857 & 0.845 & 0.817 & 0.776 & 0.628 & 0.752 \\
    & & Recall    & 0.962 & 0.960 & 0.958 & 0.955 & 0.953 & 0.946 & 0.937 & 0.910 & 0.874 & 0.759 & 0.860 \\
    & & F1-Score  & 0.916 & 0.914 & 0.912 & 0.909 & 0.907 & 0.899 & 0.888 & 0.860 & 0.822 & 0.687 & \textbf{0.802} \\

    \hline
    
    \multirow{3}{*}{HRNetV2p-W32 }  & \multirow{3}{*}{1x} 
    & Precision   & 0.877 & 0.876 & 0.874 & 0.869 & 0.862 & 0.859 & 0.839 & 0.822 & 0.759 & 0.579 & 0.728 \\
    & & Recall    & 0.969 & 0.968 & 0.967 & 0.963 & 0.957 & 0.954 & 0.939 & 0.922 & 0.870 & 0.728 & 0.849 \\
    & & F1-Score  & \textbf{0.920} & \textbf{0.919} & \textbf{0.918} & 0.913 & 0.907 & 0.904 & 0.886 & 0.869 & 0.810 & 0.645 & 0.783 \\

    \hline
    \multirow{3}{*}{HRNetV2p-W32}  & \multirow{3}{*}{2x} 
    & Precision   & 0.877 & 0.877 & 0.877 & 0.874 & 0.869 & 0.864 & 0.847 & 0.820 & 0.785 & 0.592 & 0.735 \\
    & & Recall    & 0.964 & 0.964 & 0.963 & 0.960 & 0.956 & 0.951 & 0.939 & 0.918 & 0.886 & 0.734 &  0.849 \\
    & & F1-Score  & 0.918 & 0.918 & 0.917 & 0.914 & 0.910 & 0.905 & 0.890 & 0.866 & 0.832 & 0.655 & 0.787 \\

    \hline
    
    \multirow{3}{*}{HRNetV2p-W40 }  & \multirow{3}{*}{1x} 
    & Precision   & 0.875 & 0.874 & 0.873 & 0.872 & 0.868 & 0.862 & 0.851 & 0.827 & 0.779 & 0.612 & 0.743 \\
    & & Recall    & 0.970 & 0.969 & 0.968 & 0.967 & 0.964 & 0.958 & 0.949 & 0.930 & 0.888 & 0.753 & 0.862 \\
    & & F1-Score  & \textbf{0.920} & \textbf{0.919} & \textbf{0.918} & \textbf{0.917} & \textbf{0.913} & \textbf{0.907} & \textbf{0.897} & \textbf{0.875} & 0.829 & 0.675 & 0.798 \\
    
    \hline
    \multirow{3}{*}{HRNetV2p-W40}  & \multirow{3}{*}{2x} 
    & Precision   & 0.880 & 0.880 & 0.877 & 0.877 & 0.873 & 0.861 & 0.852 & 0.834 & 0.802 & 0.629  & 0.754 \\
    & & Recall    & 0.957 & 0.957 & 0.954 & 0.954 & 0.951 & 0.943 & 0.935 & 0.918 & 0.890 & 0.755 &  0.856\\
    & & F1-Score  & 0.916 & 0.916 & 0.913 & 0.913 & 0.910 & 0.900 & 0.891 & 0.873 & \textbf{0.843} & \textbf{0.686} & 0.801 \\

    \hline
    
\end{tabular}
\end{adjustbox}
\end{center}
\label{tab:HRNets-Faster R-CNN}
\end{table}

Table \ref{tab:HRNets-HTC} shows the performance of  the HRNets HTC method with the same  Lr Schedule.  HRNetV2p-W40 backbone suffers from overfitting through the dataset. HRNetV2p-W18  achieves f1 score of 0.840, precision of 0.901 and recall 0.788 over 50\%:95\%. HRNetV2p-W18  achieves better performance over various IoUs. HRNetV2p-W32 shows less performance compare with  HRNetV2p-W18 with 8.3\% over 50\%:95\% for the f1 score.
\begin{table}[h!]
\caption{HRNets - HTC}
\begin{center}
\begin{adjustbox}{width=1\textwidth}
\begin{tabular}{|c|c|c|c|c|c|c|c|c|c|c|c|c|c|}
    \hline
    \multirow{2}{*}{Backbone}& \multirow{2}{*}{Lr schd}& \multirow{2}{*}{} & \multicolumn{11}{|c|}{ IoU } \\
    \cline{4-14}
     & & & 50\% & 55\% & 60\% & 65\% & 70\% & 75\% & 80\% & 85\% & 90\% & 95\% & 50\%:95\% \\
    
    \hline
    \multirow{3}{*}{HRNetV2p-W18}  & \multirow{3}{*}{20e} 
    & Precision   & 0.885 & 0.885 & 0.883 & 0.882 & 0.881 & 0.875 & 0.862 & 0.849 & 0.808 & 0.691 &  0.788\\
    & & Recall    & 0.987 & 0.987 & 0.984 & 0.984 & 0.982 & 0.976 & 0.966 & 0.954 & 0.915 & 0.816 &  0.901\\
    & & F1-Score  & \textbf{0.933} & \textbf{0.933} & \textbf{0.930} & \textbf{0.930} & \textbf{0.928} & \textbf{0.922} & \textbf{0.911} & \textbf{0.898} & \textbf{0.858} & \textbf{0.748} & \textbf{0.840} \\
    \hline
    \multirow{3}{*}{HRNetV2p-W32}  & \multirow{3}{*}{20e} 
    & Precision   & 0.851 & 0.851 & 0.849 & 0.846 & 0.843 & 0.834 & 0.816 & 0.792 & 0.737 & 0.516 & 0.684 \\
    & & Recall    & 0.985 & 0.985 & 0.984 & 0.981 & 0.976 & 0.968 & 0.951 & 0.929 & 0.885 & 0.710 & 0.848 \\
    & & F1-Score  & \textbf{0.913}* & \textbf{0.913}* & \textbf{0.911}* & \textbf{0.908}* & \textbf{0.904}* & \textbf{0.896}* & \textbf{0.878}* & \textbf{0.855}* & \textbf{0.804}* & \textbf{0.597}* & \textbf{0.757}* \\

    \hline
    
\end{tabular}
\end{adjustbox}
\end{center}
\label{tab:HRNets-HTC}
\end{table}

Table \ref{tab:hrnets-Mask R-CNN} shows the performance of  HRNets HTC method with the same  Lr Schedule.  HRNetV2p-W32 has achieved the highest f1 score of 0.871 over 50\%:95\% and continues to achieve the highest F1 score at various IoUs. HRNetV2p-W32  shows good performance compare with HRNetV2p-W18 with 12\% over 50\%:95\% for f1 score. HRNetV2p-W40 with 1$\times$ and 2$\times$ Lr Schedule backbones suffer from overfitting through the dataset.  Table \ref{tab:hrnet-CascadeMaskR-CNN} shows the performance of  HRNets Cascade Mask R-CNN method.  HRNetV2p-W32 and HRNetV2p-W40 backbones suffer from overfitting through the dataset.  HRNetV2p-W18 achieve f1 score of 0.903 over 50\%:95\%.
\begin{table}[h!]
\caption{HRNets - Mask R-CNN}
\begin{center}
\begin{adjustbox}{width=1\textwidth}
\begin{tabular}{|c|c|c|c|c|c|c|c|c|c|c|c|c|c|}
    \hline
    \multirow{2}{*}{Backbone}& \multirow{2}{*}{Lr schd}& \multirow{2}{*}{} & \multicolumn{11}{|c|}{ IoU } \\
    \cline{4-14}
     & & & 50\% & 55\% & 60\% & 65\% & 70\% & 75\% & 80\% & 85\% & 90\% & 95\% & 50\%:95\% \\
    
    \hline
    \multirow{3}{*}{HRNetV2p-W18}  & \multirow{3}{*}{1x} 
    & Precision   & 0.848 & 0.845 & 0.840 & 0.839 & 0.835 & 0.829 & 0.817 & 0.793 & 0.736 & 0.521 & 0.684 \\
    & & Recall    & 0.971 & 0.969 & 0.966 & 0.964 & 0.960 & 0.956 & 0.947 & 0.928 & 0.876 & 0.698 & 0.834 \\
    & & F1-Score  & \textbf{0.905}* & \textbf{0.902}* & \textbf{0.898}* & \textbf{0.897}* & \textbf{0.893}* & \textbf{0.887}* & \textbf{0.877}* & \textbf{0.855}* & \textbf{0.799}* & \textbf{0.596}* & \textbf{0.751}*\\
  
    \hline
    \multirow{3}{*}{HRNetV2p-W32}  & \multirow{3}{*}{1x} 
    & Precision   & 0.859 & 0.857 & 0.857 & 0.857  & 0.852 & 0.848 & 0.833 & 0.816 & 0.764 & 0.585& 0.816 \\
    & & Recall    & 0.971 & 0.969 & 0.969 & 0.969  & 0.965 & 0.960 & 0.947 & 0.934 & 0.889 & 0.744& 0.934 \\
    & & F1-Score  & \textbf{0.911} & \textbf{0.909} & \textbf{0.909} & \textbf{0.909}  & \textbf{0.904} & \textbf{0.900} & \textbf{0.886} & \textbf{0.871} & \textbf{0.821} & \textbf{0.654}& \textbf{0.871}\\

    \hline

\end{tabular}
\end{adjustbox}
\end{center}
\label{tab:hrnets-Mask R-CNN}
\end{table}


\begin{table}[h!]
\caption{HRNets - Cascade Mask R-CNN}
\begin{center}
\begin{adjustbox}{width=1\textwidth}
\begin{tabular}{|c|c|c|c|c|c|c|c|c|c|c|c|c|c|}
    \hline
    \multirow{2}{*}{Backbone}& \multirow{2}{*}{Lr schd}& \multirow{2}{*}{} & \multicolumn{11}{|c|}{ IoU } \\
    \cline{4-14}
     & & & 50\% & 55\% & 60\% & 65\% & 70\% & 75\% & 80\% & 85\% & 90\% & 95\% & 50\%:95\% \\
    
    \hline
    \multirow{3}{*}{HRNetV2p-W18}  & \multirow{3}{*}{20e} 
    & Precision   & 0.888 & 0.887 & 0.887 & 0.886 & 0.885 & 0.884 & 0.872 & 0.858 & 0.828 & 0.732 & 0.810 \\
    & & Recall    & 0.970 & 0.970 & 0.970 & 0.967 & 0.967 & 0.965 & 0.955 & 0.942 & 0.918 & 0.836 & 0.903 \\
    & & F1-Score  & 0.927 & 0.926 & 0.926 & 0.924 & 0.924 & 0.922 & 0.911 & 0.898 & 0.870 & 0.780 & 0.903 \\

    \hline
\end{tabular}
\end{adjustbox}
\end{center}
\label{tab:hrnet-CascadeMaskR-CNN}
\end{table}

Table \ref{tab:hrnets-FCOS} shows the performance of HRNets with Fully Convolutional One-Stage (FCOS) Object Detection. HRNets FCOS achieves less performance compared with other models. HRNetV2p-W18  with 2$\times$  Lr Schedule achieve an increment of 25\% f1 score from HRNetV2p-W18  with 1$\times$  Lr Schedule and  HRNetV2p-W18  with 2$\times$  Lr Schedule achieve an increment of 22.5\% f1 score from HRNetV2p-W32.  HRNetV2p-W18  with 2$\times$  Lr Schedule achieve an f1 score of 0.648.

\begin{table}[h!]
\caption{HRNets - FCOS}
\begin{center}
\begin{adjustbox}{width=1\textwidth}
\begin{tabular}{|c|c|c|c|c|c|c|c|c|c|c|c|c|c|}
    \hline
    \multirow{2}{*}{Backbone}& \multirow{2}{*}{Lr schd}& \multirow{2}{*}{} & \multicolumn{11}{|c|}{ IoU } \\
    \cline{4-14}
     & & & 50\% & 55\% & 60\% & 65\% & 70\% & 75\% & 80\% & 85\% & 90\% & 95\% & 50\%:95\% \\
    
    \hline
    \multirow{3}{*}{HRNetV2p-W18}  & \multirow{3}{*}{1x} 
    & Precision   & 0.511 & 0.507 & 0.498 & 0.485 & 0.467 & 0.441 & 0.405 & 0.328 & 0.222 & 0.086 & 0.298 \\
    & & Recall    & 0.959 & 0.946 & 0.930 & 0.910 & 0.885 & 0.844 & 0.798 & 0.697 & 0.517 & 0.244 & 0.601 \\
    & & F1-Score  & \textbf{0.666}* & \textbf{0.660}* & \textbf{0.648}* & \textbf{0.632}* & \textbf{0.611}* &\textbf{ 0.579}* & \textbf{0.537}* & \textbf{0.446}* & \textbf{0.310}* & \textbf{0.127}* & \textbf{0.398}* \\

    \hline
    \multirow{3}{*}{HRNetV2p-W18}  & \multirow{3}{*}{2x} 
    & Precision   & 0.790 & 0.788 & 0.782 & 0.779 & 0.770 & 0.759 & 0.729 & 0.691 & 0.596 & 0.335 & 0.563 \\
    & & Recall    & 0.983 & 0.978 & 0.972 & 0.969 & 0.959 & 0.947 & 0.917 & 0.878 & 0.786 & 0.545 & 0.764 \\
    & & F1-Score  & \textbf{0.875} & \textbf{0.872} & \textbf{0.866} & \textbf{0.863} & \textbf{0.854} & \textbf{0.842} & \textbf{0.812} & \textbf{0.773} & \textbf{0.677} & \textbf{0.414} & \textbf{0.648} \\

    \hline
    \multirow{3}{*}{HRNetV2p-W32}  & \multirow{3}{*}{1x} 
    & Precision   & 0.566 & 0.561 & 0.555 & 0.539 & 0.528 & 0.504 & 0.469 & 0.400 & 0.275 & 0.086 & 0.326 \\
    & & Recall    & 0.970 & 0.964 & 0.956 & 0.928 & 0.906 & 0.868 & 0.818 & 0.730 & 0.571 & 0.241 & 0.605 \\
    & & F1-Score  & 0.714 & 0.709 & 0.702 & 0.681 & 0.667 & 0.637 & 0.596 & 0.516 & 0.371 & 0.126 & 0.423 \\

    \hline
    
\end{tabular}
\end{adjustbox}
\end{center}
\label{tab:hrnets-FCOS}
\end{table}

In the following  \cref{tab:Resnest-CascadeR-CNN,tab:Resnest-Faster R-CNN} show the comparative analysis, we have trained ResNeSt with Cascade R-CNN and Faster R-CNN methods, and each method with different backbones(Resnest-50, Resnest-101). For Cascade R-CNN S-101 backbone achieve an f1 score of 0.845 over 50\%:95\% IoU and also achieves the highest performance compare with the S-50 backbone and the Faster R-CNN method. The Faster R-CNN S-101 backbone achieves an f1 score of 0.748 over 50\%:95\% IoU. Cascade R-CNN S-101 backbone has an increment of  9.2\% over 50\%:95\% for the f1 score.

Dynamic RCNN proposes by \cite{DynamicRCNN}, it is a simple but effective method for maximizing the dynamic quality of object detection proposals. Table \ref{tab:Dynamic R-CNN} show the Dynamic RCNN with  Resnet-50 achieves an f1 score of 0.628, the precision of 0.561, recall of 0.714  over 50\%:95\%.

\begin{table}[h!]
\caption{Resnest - Cascade R-CNN}
\begin{center}
\begin{adjustbox}{width=1\textwidth}
\begin{tabular}{|c|c|c|c|c|c|c|c|c|c|c|c|c|c|}
    \hline
    \multirow{2}{*}{Backbone}& \multirow{2}{*}{Lr schd}& \multirow{2}{*}{} & \multicolumn{11}{|c|}{ IoU } \\
    \cline{4-14}
    & & & 50\% & 55\% & 60\% & 65\% & 70\% & 75\% & 80\% & 85\% & 90\% & 95\% & 50\%:95\% \\
    
    \hline
    \multirow{3}{*}{S-50}  & \multirow{3}{*}{1x} 
    & Precision   & 0.895 & 0.894 & 0.891 & 0.885 & 0.881 & 0.875 & 0.870 & 0.854 & 0.808 & 0.659 & 0.777 \\
    & & Recall    & 0.977 & 0.976 & 0.974 & 0.969 & 0.965 & 0.959 & 0.954 & 0.940 & 0.903 & 0.784 & 0.880 \\
    & & F1-Score  & 0.934 & 0.933 & 0.930 & 0.925 & 0.921 & 0.915 & 0.910 & 0.894 & 0.852 & 0.716 & 0.825 \\

    \hline
    \multirow{3}{*}{S-101}  & \multirow{3}{*}{1x} 
    & Precision   & 0.905 & 0.903 & 0.902 & 0.899 & 0.893 & 0.891 & 0.884 & 0.876 & 0.826 & 0.693 & 0.799 \\
    & & Recall    & 0.985 & 0.984 & 0.983 & 0.979 & 0.976 & 0.972 & 0.965 & 0.958 & 0.917 & 0.811 & 0.898 \\
    & & F1-Score  & 0.943 & \textbf{0.941} & \textbf{0.940} & \textbf{0.937} & \textbf{0.932} & \textbf{0.929} & \textbf{0.922} & \textbf{0.915} & \textbf{0.869} & \textbf{0.747} & \textbf{0.845} \\

    \hline
    
\end{tabular}
\end{adjustbox}
\end{center}
\label{tab:Resnest-CascadeR-CNN}
\end{table}

\begin{table}[h!]
\caption{Resnest - Faster R-CNN}
\begin{center}
\begin{adjustbox}{width=1\textwidth}
\begin{tabular}{|c|c|c|c|c|c|c|c|c|c|c|c|c|c|}
    \hline
    \multirow{2}{*}{Backbone}& \multirow{2}{*}{Lr schd}& \multirow{2}{*}{} & \multicolumn{11}{|c|}{ IoU } \\
    \cline{4-14}
     & & & 50\% & 55\% & 60\% & 65\% & 70\% & 75\% & 80\% & 85\% & 90\% & 95\% & 50\%:95\% \\
    
    \hline
    \multirow{3}{*}{S-50}  & \multirow{3}{*}{1x} 
    & Precision   & 0.884 & 0.884 & 0.880 & 0.879 & 0.872 & 0.861 & 0.844 & 0.809 & 0.709 & 0.429 & 0.656 \\
    & & Recall    & 0.970 & 0.970 & 0.968 & 0.967 & 0.961 & 0.951 & 0.935 & 0.906 & 0.824 & 0.597 & 0.784 \\
    & & F1-Score  & 0.925 & 0.925 & 0.921 & 0.920 & 0.914 & 0.903 & 0.887 & 0.854 & 0.762 & 0.499 & 0.714 \\

    \hline
    \multirow{3}{*}{S-101}  & \multirow{3}{*}{1x} 
    & Precision   & 0.893 & 0.893 & 0.890 & 0.888 & 0.879 & 0.876 & 0.862 & 0.823 & 0.747 & 0.495 & 0.694 \\
    & & Recall    & 0.981 & 0.979 & 0.977 & 0.975 & 0.967 & 0.963 & 0.950 & 0.921 & 0.861 & 0.645 & 0.813 \\
    & & F1-Score  & \textbf{0.934} & \textbf{0.934} & \textbf{0.931} & \textbf{0.929} & \textbf{0.920} & \textbf{0.917} & \textbf{0.903} & \textbf{0.869} & \textbf{0.799} & \textbf{0.560} &  \textbf{0.748}\\

    \hline
    
\end{tabular}
\end{adjustbox}
\end{center}
\label{tab:Resnest-Faster R-CNN}
\end{table}

\begin{table}[h!]
\caption{Dynamic R-CNN}
\begin{center}
\begin{adjustbox}{width=1\textwidth}
\begin{tabular}{|c|c|c|c|c|c|c|c|c|c|c|c|c|c|}
    \hline
    \multirow{2}{*}{Backbone}& \multirow{2}{*}{Lr schd}& \multirow{2}{*}{} & \multicolumn{11}{|c|}{ IoU } \\
    \cline{4-14}
     & & & 50\% & 55\% & 60\% & 65\% & 70\% & 75\% & 80\% & 85\% & 90\% & 95\% & 50\%:95\% \\
    
    \hline
    \multirow{3}{*}{Resnet-50}  & \multirow{3}{*}{1x} 
    & Precision   & 0.855 & 0.854 & 0.853 & 0.849 & 0.839 & 0.823 & 0.802 & 0.764 & 0.646 & 0.267 & 0.561 \\
    & & Recall    & 0.978 & 0.977 & 0.975 & 0.971 & 0.963 & 0.943 & 0.925 & 0.888 & 0.793 & 0.451 & 0.714 \\
    & & F1-Score  & 0.912 & 0.911 & 0.909 & 0.905 & 0.896 & 0.878 & 0.859 & 0.821 & 0.711 & 0.335 & 0.628 \\

    \hline
    
\end{tabular}
\end{adjustbox}
\end{center}
\label{tab:Dynamic R-CNN}
\end{table}

\begin{table}[h!]
\caption{Table detection}
\begin{center}
\begin{adjustbox}{width=1\textwidth}
\begin{tabular}{|c|c|c|c|c|c|c|c|c|c|c|c|c|c|c|c|}
    \hline
    \multirow{2}{*}{Approach}& \multirow{2}{*}{Dataset}& \multirow{2}{*}{Method} & \multirow{2}{*}{} & \multicolumn{11}{|c|}{ IoU } & \multirow{2}{*}{Year}\\
    \cline{5-15}
     &  & & & 50\% & 55\% & 60\% & 65\% & 70\% & 75\% & 80\% & 85\% & 90\% & 95\% & 50\%:95\%  & \\
     
     \hline
    \multirow{3}{*}{Tesseract \cite{shafait2010table}}  & \multirow{3}{*}{UNLV}  & \multirow{3}{*}{Tab-stop Detection} 
    & Precision     & - & - & - & - & - & - & - & - & 86.00 & - & - & \multirow{3}{*}{2010} \\
    & & & Recall    & - & - & - & - & - & - & - & - & 79.00 & - & - & \\
    & & & F1-Score  & - & - & - & - & - & - & - & - & 82.35 & - & - & \\
    
    \hline
    \multirow{3}{*}{A Gilani\cite{gilani2017table}}  & \multirow{3}{*}{UNLV}  & \multirow{3}{*}{Faster R-CNN} 
    & Precision     & - & - & - & - & - & - & - & - & 82.30 & - & - & \multirow{3}{*}{2017} \\
    & & & Recall    & - & - & - & - & - & - & - & - & 90.67 & - & - & \\
    & & & F1-Score  & - & - & - & - & - & - & - & - & 86.29 & - & - & \\
    
    \hline
    \multirow{3}{*}{SA Siddiqui\cite{siddiqui2018decnt}}  & \multirow{3}{*}{UNLV}  & \multirow{3}{*}{ \makecell{Deformable CNN \\ + Faster R-CNN}} 
    & Precision     & 78.6 & - & - & - & - & - & - & - & - & - & - & \multirow{3}{*}{2018} \\
    & & & Recall    & 74.9 & - & - & - & - & - & - & - & - & - & - & \\
    & & & F1-Score  & 76.7 & - & - & - & - & - & - & - & - & - & - & \\
    
    \hline
    \multirow{3}{*}{Á Casado-García\cite{casado2020benefits}}  & \multirow{3}{*}{UNLV}  & \multirow{3}{*}{YOLO} 
    & Precision     & - & - &  93.0 & - & 92.0 & - & 83.0 & - & 48.0 & - & - & \multirow{3}{*}{2020} \\
    & & & Recall    & - & - &  95.0 & - & 94.0 & - & 85.0 & - & 49.0 & - & - & \\
    & & & F1-Score  & - & - &  94.0 & - & 93.0 & - & 84.0 & - & 49.0 & - & - &\\
    
    \hline
    \multirow{3}{*}{ M Agarwal \cite{agarwal2021cdec}}  & \multirow{3}{*}{UNLV} & \multirow{3}{*}{\makecell{\makecell{Cascade mask\\ R-CNN }}} 
    & Precision     & 96.0 & - & 94.4 & - & 91.5 & - & 82.6 & - & 61.8 & - & - & \multirow{3}{*}{2018} \\
    & & & Recall    & 77.0 & - & 75.8 & - & 73.4 & - & 66.3 & - & 49.6 & - & - & \\
    & & & F1-Score  & 86.5 & - & 85.1 & - & 82.5 & - & 74.4 & - & 55.7 & - & - & \\
    
    \hline
    \multirow{3}{*}{S Schreiber\cite{schreiber2017deepdesrt}}  & \multirow{3}{*}{ICDAR2013}  & \multirow{3}{*}{Mask R-CNN} 
    & Precision     & 97.40 & - &  - & - & - & - & - & - & - & - & - & \multirow{3}{*}{2017} \\
    & & & Recall    & 96.15 & - &  - & - & - & - & - & - & - & - & - & \\
    & & & F1-Score  & 96.77 & - &  - & - & - & - & - & - & - & - & - &\\

    \hline
    \multirow{3}{*}{ SA Siddiqui\cite{siddiqui2019deeptabstr}}  & \multirow{3}{*}{ICDAR2013}  & \multirow{3}{*}{\makecell{Deformable CNN}} 
    & Precision     & 99.6 & - &  - & - & - & - & - & - & - & - & - & \multirow{3}{*}{2018} \\
    & & & Recall    & 99.6 & - &  - & - & - & - & - & - & - & - & - & \\
    & & & F1-Score  & 99.6 & - &  - & - & - & - & - & - & - & - & - &\\

    \hline
    \multirow{3}{*}{ I Kavasidis\cite{kavasidis2019saliency}}  & \multirow{3}{*}{ICDAR2013}  & \multirow{3}{*}{\makecell{Semantic Image \\Segmentation}} 
    & Precision     & 97.5 & - &  - & - & - & - & - & - & - & - & - & \multirow{3}{*}{2019} \\
    & & & Recall    & 98.1 & - &  - & - & - & - & - & - & - & - & - & \\
    & & & F1-Score  & 97.8 & - &  - & - & - & - & - & - & - & - & - &\\

    \hline
    \multirow{3}{*}{ Y Huang\cite{huang2019yolo}}  & \multirow{3}{*}{ICDAR2013}  & \multirow{3}{*}{\makecell{YOLO}} 
    & Precision     & 100 & - &  98.6 & - & - & - & 89.2 & - & - & - & - & \multirow{3}{*}{2019} \\
    & & & Recall    & 94.9 & - &  93.6 & - & - & - & 84.6 & - & - & - & - & \\
    & & & F1-Score  & 97.3 & - &  96.1 & - & - & - & 86.8 & - & - & - & - &\\
    
    \hline
    \multirow{3}{*}{ SS Paliwal\cite{paliwal2019tablenet}}  & \multirow{3}{*}{ICDAR2013}  & \multirow{3}{*}{\makecell{fully convolutions}} 
    & Precision     & 96.97 & - &  - & - & - & - & - & - & - & - & - & \multirow{3}{*}{2019} \\
    & & & Recall    & 96.28 & - &  - & - & - & - & - & - & - & - & - & \\
    & & & F1-Score  & 96.62 & - &  - & - & - & - & - & - & - & - & - &\\
    
    \hline
    \multirow{3}{*}{Á Casado-García\cite{casado2020benefits}}  & \multirow{3}{*}{ICDAR2013}  & \multirow{3}{*}{Mask R-CNN} 
    & Precision     & - & - &  70.0 & - & 70.0 & - & 70.0 & - & 47.0 & - & - & \multirow{3}{*}{2020} \\
    & & & Recall    & - & - &  97.0 & - & 97.0 & - & 97.0 & - & 65.0 & - & - & \\
    & & & F1-Score  & - & - &  81.0 & - & 81.0 & - & 81.0 & - & 54.0 & - & - &\\

    \hline
    \multirow{3}{*}{D Prasad\cite{prasad2020cascadetabnet}}  & \multirow{3}{*}{ICDAR2013}  & \multirow{3}{*}{\makecell{\makecell{Cascade mask\\ R-CNN HRNet}}} 
    & Precision     & 100 & - &  - & - & - & - & - & - & - & - & - & \multirow{3}{*}{2020} \\
    & & & Recall    & 100 & - &  - & - & - & - & - & - & - & - & - & \\
    & & & F1-Score  & 100 & - &  - & - & - & - & - & - & - & - & - &\\
    
    \hline
    \multirow{3}{*}{ M Li\cite{li2020tablebank}}  & \multirow{3}{*}{ICDAR2013}  & \multirow{3}{*}{\makecell{Faster R-CNN}} 
    & Precision     & 96.58 & - &  - & - & - & - & - & - & - & - & - & \multirow{3}{*}{2020} \\
    & & & Recall    & 95.94 & - &  - & - & - & - & - & - & - & - & - & \\
    & & & F1-Score  & 96.25 & - &  - & - & - & - & - & - & - & - & - &\\
    \hline
    
     \hline
    \multirow{3}{*}{ M Agarwal \cite{agarwal2021cdec}}  & \multirow{3}{*}{ICDAR2013}  & \multirow{3}{*}{\makecell{\makecell{Cascade mask\\ R-CNN }}} 
    & Precision     & 100.0 & - &  100.0 & - & 98.7 & - & 94.2 & - & 66.0 & - & - & \multirow{3}{*}{2021} \\
    & & & Recall    & 100.0 & - &  100.0 & - & 98.7 & - & 94.2 & - & 66.0 & - & - & \\
    & & & F1-Score  & 100.0 & - &  100.0 & - & 98.7 & - & 94.2 & - & 66.0 & - & - &\\
    
    \hline
    \multirow{3}{*}{ X Zheng\cite{zheng2021global}}  & \multirow{3}{*}{ICDAR2013}  & \multirow{3}{*}{\makecell{\makecell{object detection\\ networks }}} 
    & Precision     & 98.97 & - &  - & - & - & - & - & - & - & - & - & \multirow{3}{*}{2021} \\
    & & & Recall    & 99.77 & - &  - & - & - & - & - & - & - & - & - & \\
    & & & F1-Score  & 99.31 & - &  - & - & - & - & - & - & - & - & - &\\
    
    \hline
    \multirow{3}{*}{ SA Siddiqui\cite{siddiqui2019deeptabstr}}  & \multirow{3}{*}{ICDAR2017}  & \multirow{3}{*}{\makecell{Deformable CNN}} 
    & Precision     & - & - &  96.5 & -  & - & - & 96.7  & - & - & - & - & \multirow{3}{*}{2018} \\
    & & & Recall    & - & - &  97.1 & - & - & - & 93.7 & - & - & - & - & \\
    & & & F1-Score  & - & - &  96.8 & - & - & - & 95.2 & - & - & - & - &\\
    
    \hline
    \multirow{3}{*}{ Y Huang\cite{huang2019yolo}}  & \multirow{3}{*}{ICDAR2017}  & \multirow{3}{*}{\makecell{YOLO}} 
    & Precision     & - & - &  97.8 & - & - & - & 97.5 & - & - & - & - & \multirow{3}{*}{2019} \\
    & & & Recall    & - & - &  97.2 & - & - & - & 96.8 & - & - & - & - & \\
    & & & F1-Score  & - & - &  97.5 & - & - & - & 97.1 & - & - & - & - &\\
    
    \hline
    
\end{tabular}
\end{adjustbox}
\end{center}
\label{tab:Table Detection1}
\end{table}
    
\begin{table}[h!]
\caption{Table detection (Continue Table \ref{tab:Table Detection1})}
\begin{center}
\begin{adjustbox}{width=1\textwidth}
\begin{tabular}{|c|c|c|c|c|c|c|c|c|c|c|c|c|c|c|c|}
    \hline
    \multirow{2}{*}{Approach}& \multirow{2}{*}{Dataset}& \multirow{2}{*}{Method} & \multirow{2}{*}{} & \multicolumn{11}{|c|}{ IoU } & \multirow{2}{*}{Year}\\
    \cline{5-15}
     &  & & & 50\% & 55\% & 60\% & 65\% & 70\% & 75\% & 80\% & 85\% & 90\% & 95\% & 50\%:95\%  & \\

    \hline
    \multirow{3}{*}{ Y Li\cite{li2019gan}}  & \multirow{3}{*}{ICDAR2017}  & \multirow{3}{*}{\makecell{\makecell{Generative Adversarial \\Networks(GAN) }}} 
    & Precision     & - & - &  94.4 & - & - & - & 90.3 & - & - & - & - & \multirow{3}{*}{2019} \\
    & & & Recall    & - & - &  94.4 & - & - & - & 90.3 & - & - & - & - & \\
    & & & F1-Score  & - & - &  94.4 & - & - & - & 90.3 & - & - & - & - &\\

    \hline
    \multirow{3}{*}{ N Sun \cite{sun2019faster}}  & \multirow{3}{*}{ICDAR2017}  & \multirow{3}{*}{Faster R-CNN} 
    & Precision     & - & - &  - & - & - & - & 94.3 & - & - & - & - & \multirow{3}{*}{2019} \\
    & & & Recall    & - & - &  - & - & - & - & 95.6 & - & - & - & - & \\
    & & & F1-Score  & - & - &  - & - & - & - & 94.9 & - & - & - & - &\\
    
    \hline
    \multirow{3}{*}{Á Casado-García\cite{casado2020benefits}}  & \multirow{3}{*}{ICDAR2017}  & \multirow{3}{*}{RetinaNet} 
    & Precision     & - & - &  92.0 & - & 92.0 & - & 89.0 & - & 79.0 & - & - & \multirow{3}{*}{2020} \\
    & & & Recall    & - & - &  87.0 & - & 87.0 & - & 84.0 & - & 75.0 & - & - & \\
    & & & F1-Score  & - & - &  89.0 & - & 89.0 & - & 86.0 & - & 77.0 & - & - &\\
    
    \hline
    \multirow{3}{*}{  M Agarwal \cite{agarwal2021cdec}}  & \multirow{3}{*}{ICDAR2017}  & \multirow{3}{*}{\makecell{\makecell{Cascade mask\\ R-CNN }}} 
    & Precision     & - & - &  96.9 & - & - & - & - & - & - & - & - & \multirow{3}{*}{2021} \\
    & & & Recall    & - & - &  89.9 & - & - & - & - & - & - & - & - & \\
    & & & F1-Score  & - & - &  93.4 & - & - & - & - & - & - & - & - &\\
    
    \hline
    \multirow{3}{*}{D Prasad\cite{prasad2020cascadetabnet}}  & \multirow{3}{*}{ICDAR2019}  & \multirow{3}{*}{\makecell{\makecell{Cascade mask\\ R-CNN HRNet}}} 
    & Precision     & - & - &  - & - & - & - & - & - & - & - & - & \multirow{3}{*}{2020} \\
    & & & Recall    & - & - &  - & - & - & - & - & - & - & - & - & \\
    & & & F1-Score  & - & - &  94.3 & - & 93.4 & - & 92.5 & - & 90.1 & - & - &\\
    \hline

     \hline
    \multirow{3}{*}{ M Agarwal \cite{agarwal2021cdec}}  & \multirow{3}{*}{ICDAR2019}  & \multirow{3}{*}{\makecell{\makecell{Cascade mask\\ R-CNN }}} 
    & Precision     & 98.7 & - & 98.0 & - & 97.7 & - & 97.1 & - & 93.4 & - & - & \multirow{3}{*}{2021} \\
    & & & Recall    & 94.6 & - & 93.9 & - & 93.6 & - & 93.0 & - & 89.5 & - & - & \\
    & & & F1-Score  & 96.6 & - & 95.9 & - & 95.6 & - & 95.0 & - & 91.5 & - & - &\\
    
    \multirow{3}{*}{ X Zheng\cite{zheng2021global}}  & \multirow{3}{*}{ICDAR2019}  & \multirow{3}{*}{\makecell{\makecell{object detection\\ networks }}} 
    & Precision     & - & - &  - & - & - & - & 96.0 & - & 90.0 & - & - & \multirow{3}{*}{2021} \\
    & & & Recall    & - & - &  - & - & - & - & 95.0 & - & 89.0 & - & - & \\
    & & & F1-Score  & - & - &  - & - & - & - & 95.5 & - & 95.5 & - & - &\\
    \hline
    
    \multirow{3}{*}{SA Siddiqui\cite{siddiqui2018decnt}}  & \multirow{3}{*}{Mormot}  & \multirow{3}{*}{ \makecell{Deformable CNN}} 
    & Precision     & 84.9 & - & - & - & - & - & - & - & - & - & - & \multirow{3}{*}{2018} \\
    & & & Recall    & 94.6 & - & - & - & - & - & - & - & - & - & - & \\
    & & & F1-Score  & 89.5 & - & - & - & - & - & - & - & - & - & - & \\
    
    \hline
    \multirow{3}{*}{ M Agarwal \cite{agarwal2021cdec}}  & \multirow{3}{*}{TableBank}  & \multirow{3}{*}{\makecell{\makecell{Cascade mask\\ R-CNN }}} 
    & Precision     & 93.4 & - &  99.5 & - & - & - & - & - & - & - & - & \multirow{3}{*}{2021} \\
    & & & Recall    & 92.4 & - &  97.8 & - & - & - & - & - & - & - & - & \\
    & & & F1-Score  & 92.9 & - &  98.6 & - & - & - & - & - & - & - & - &\\

    \hline
    \multirow{3}{*}{P Riba \cite{riba2019table}}  & \multirow{3}{*}{RVL-CDIP}  & \multirow{3}{*}{Graph NN} 
    & Precision     & 15.2 & - &  - & - & - & - & - & - & - & - & - & \multirow{3}{*}{2019} \\
    & & & Recall    & 36.5 & - &  - & - & - & - & - & - & - & - & - & \\
    & & & F1-Score  & 21.5 & - &  - & - & - & - & - & - & - & - & - &\\
    
    \hline
    
\end{tabular}
\end{adjustbox}
\end{center}
\label{tab:Table Detection2}
\end{table}

\begin{table}[h!]
\caption{Table Structure Recognition}
\begin{center}
\begin{adjustbox}{width=1\textwidth}
\begin{tabular}{|c|c|c|c|c|c|c|c|c|c|c|c|c|c|c|c|}
    \hline
    \multirow{2}{*}{Approach}& \multirow{2}{*}{Dataset}& \multirow{2}{*}{Method} & \multirow{2}{*}{} & \multicolumn{11}{|c|}{ IoU } & \multirow{2}{*}{Year}\\
    \cline{5-15}
     &  & & & 50\% & 55\% & 60\% & 65\% & 70\% & 75\% & 80\% & 85\% & 90\% & 95\% & 50\%:95\%  & \\

    \hline
    \multirow{3}{*}{S Schreiber\cite{schreiber2017deepdesrt}}  & \multirow{3}{*}{ICDAR2013}  & \multirow{3}{*}{Fully CNN} 
    & Precision     & 95.93 & - &  - & - & - & - & - & - & - & - & - & \multirow{3}{*}{2017} \\
    & & & Recall    & 87.36 & - &  - & - & - & - & - & - & - & - & - & \\
    & & & F1-Score  & 91.44 & - &  - & - & - & - & - & - & - & - & - &\\

    \hline
    \multirow{3}{*}{SA Siddiqui\cite{siddiqui2019deeptabstr}}  & \multirow{3}{*}{ICDAR2013}  & \multirow{3}{*}{Deformable CNN} 
    & Precision     & 93.19 & - &  - & - & - & - & - & - & - & - & - & \multirow{3}{*}{2019} \\
    & & & Recall    & 93.08 & - &  - & - & - & - & - & - & - & - & - & \\
    & & & F1-Score  & 92.98 & - &  - & - & - & - & - & - & - & - & - &\\
    
    \hline
    \multirow{3}{*}{W Xue\cite{xue2019res2tim}}  & \multirow{3}{*}{ICDAR2013}  & \multirow{3}{*}{\makecell{Graph NN + weights\\ depending on distance}}
    & Precision     & 92.6 & - &  - & - & - & - & - & - & - & - & - & \multirow{3}{*}{2019} \\
    & & & Recall    & 44.7 & - &  - & - & - & - & - & - & - & - & - & \\
    & & & F1-Score  & 60.3 & - &  - & - & - & - & - & - & - & - & - &\\
    
    \hline
    \multirow{3}{*}{ SS Paliwal\cite{paliwal2019tablenet}}  & \multirow{3}{*}{ICDAR2013}  & \multirow{3}{*}{\makecell{fully CNN}} 
    & Precision     & 92.15 & - &  - & - & - & - & - & - & - & - & - & \multirow{3}{*}{2019} \\
    & & & Recall    & 89.87 & - &  - & - & - & - & - & - & - & - & - & \\
    & & & F1-Score  & 90.98 & - &  - & - & - & - & - & - & - & - & - &\\

    \hline
    \multirow{3}{*}{SA Khan\cite{khan2019table}}  & \multirow{3}{*}{ICDAR2013}  & \multirow{3}{*}{\makecell{Bi-directional RNN}} 
    & Precision     & 96.92 & - &  - & - & - & - & - & - & - & - & - & \multirow{3}{*}{2019} \\
    & & & Recall    & 90.12 & - &  - & - & - & - & - & - & - & - & - & \\
    & & & F1-Score  & 93.39 & - &  - & - & - & - & - & - & - & - & - &\\

    \hline
    \multirow{3}{*}{C Tensmeyer\cite{tensmeyer2019deep}}  & \multirow{3}{*}{ICDAR2013}  & \multirow{3}{*}{\makecell{Dilated Convolutions \\+ Fully CNN}} 
    & Precision     & 95.8 & - &  - & - & - & - & - & - & - & - & - & \multirow{3}{*}{2019} \\
    & & & Recall    & 94.6 & - &  - & - & - & - & - & - & - & - & - & \\
    & & & F1-Score  & 95.2 & - &  - & - & - & - & - & - & - & - & - &\\ 
    
    \hline
    \multirow{3}{*}{Z Chi\cite{chi2019complicated}}  & \multirow{3}{*}{ICDAR2013}  & \multirow{3}{*}{\makecell{Fully CNN}} 
    & Precision     & 88.5 & - &  - & - & - & - & - & - & - & - & - & \multirow{3}{*}{2019} \\
    & & & Recall    & 86.0 & - &  - & - & - & - & - & - & - & - & - & \\
    & & & F1-Score  & 87.2 & - &  - & - & - & - & - & - & - & - & - &\\

    \hline
    \multirow{3}{*}{Á Casado-García\cite{casado2020benefits}}  & \multirow{3}{*}{ICDAR2013}  & \multirow{3}{*}{Mask R-CNN} 
    & Precision     & - & - &  70.0 & - & 70.0 & - & 70.0 & - & 47.0 & - & - & \multirow{3}{*}{2020} \\
    & & & Recall    & - & - &  97.0 & - & 97.0 & - & 97.0 & - & 65.0 & - & - & \\
    & & & F1-Score  & - & - &  81.0 & - & 81.0 & - & 81.0 & - & 54.0 & - & - &\\

    \hline
    \multirow{3}{*}{S Raja\cite{raja2020table}}  & \multirow{3}{*}{ICDAR2013}  & \multirow{3}{*}{\makecell{Object Detection Methods}} 
    & Precision     & 92.7 & - &  - & - & - & - & - & - & - & - & - & \multirow{3}{*}{2020} \\
    & & & Recall    & 91.1 & - &  - & - & - & - & - & - & - & - & - & \\
    & & & F1-Score  & 91.9 & - &  - & - & - & - & - & - & - & - & - &\\  
    
    \hline
    \multirow{3}{*}{KA Hashmi\cite{hashmi2021guided}}  & \multirow{3}{*}{ICDAR2013}  & \multirow{3}{*}{\makecell{Object Detection Methods}} 
    & Precision     & 95.37 & - &  - & - & - & - & - & - & - & - & - & \multirow{3}{*}{2021} \\
    & & & Recall    & 95.56 & - &  - & - & - & - & - & - & - & - & - & \\
    & & & F1-Score  & 95.46 & - &  - & - & - & - & - & - & - & - & - &\\  
    
    \hline
    \multirow{3}{*}{D Prasad\cite{prasad2020cascadetabnet}}  & \multirow{3}{*}{ICDAR2019}  & \multirow{3}{*}{\makecell{Object Detection Methods}} 
    & Precision     & - & - &  - & - & - & - & - & - & - & - & - & \multirow{3}{*}{2020} \\
    & & & Recall    & - & - &  - & - & - & - & - & - & - & - & - & \\
    & & & F1-Score  & - & - &  43.8 & - & 35.4 & - & 19.0 & - & 3.6 & - & - &\\

    \hline
    \multirow{3}{*}{Y Zou\cite{zou2020deep}}  & \multirow{3}{*}{ICDAR2019}  & \multirow{3}{*}{\makecell{Fully CNN}} 
    & Precision     & - & - &  18.79 & - & - & - & 1.71 & - & - & - & - & \multirow{3}{*}{2021} \\
    & & & Recall    & - & - &  10.07 & - & - & - & 0.92 & - & - & - & - & \\
    & & & F1-Score  & - & - &  13.11 & - & - & - & 1.19 & - & - & - & - &\\

    \hline
    \multirow{3}{*}{X Zheng\cite{zheng2021global}}  & \multirow{3}{*}{ICDAR2019}  & \multirow{3}{*}{\makecell{Object Detection Methods}} 
    & Precision     & - & - &  - & - & - & - & - & - & - & - & - & \multirow{3}{*}{2021} \\
    & & & Recall    & - & - &  - & - & - & - & - & - & - & - & - & \\
    & & & F1-Score  & 54.8 & - &  38.5 & - & - & - & - & - & - & - & - &\\  
    
    \hline
    
\end{tabular}
\end{adjustbox}
\end{center}
\label{tab:Table Recognition}
\end{table}

\begin{table}[h!]
\caption{ Open source code for most of the studies articles in Table Detection }
\begin{center}
\begin{adjustbox}{width=1\textwidth}
\begin{tabular}{|c|c|c|c|c|} 
    \hline
     Article  &   Model    & Year &   Framework  & Link \\ \hline
     
     Z Chi \cite{chi2019complicated} & SciTSR &   2019    &   Pytorch & \url{https://github.com/Academic-Hammer/SciTSR}    \\ \hline
     
     D Prasad \cite{prasad2020cascadetabnet} & CascadeTabNet &   2020    &  Pytorch  &   \url{https://github.com/DevashishPrasad/CascadeTabNet    } \\ \hline
     
     Á Casado-García \cite{casado2020benefits} & - &   2020    &  mxnet  & \url{https://github.com/holms-ur/fine-tuning}    \\ \hline
     
     M Li \cite{prasad2020cascadetabnet} & TableBank &   2020    & Pytorch, Detectron2    &      \url{https://github.com/doc-analysis/TableBank}    \\ \hline

     S Raja S Raja \cite{raja2020table}& TabStructNet &   2020    &  tensorflow   &      \url{https://github.com/sachinraja13/TabStructNet.git}    \\ \hline

     X Zhong \cite{zhong2020image} & PubTabNet &   2020    &  -   & \url{https://github.com/ibm-aur-nlp/PubTabNet}    \\ \hline

     M Agarwal \cite{agarwal2021cdec} & CDeC-Net &   2021    &   PyTorch  &   \url{https://github.com/mdv3101/CDeCNet}    \\ \hline

\end{tabular}
\end{adjustbox}
\end{center}
\label{tab:sourcecode}
\end{table}


\subsection{Result of different datasets}
\subsubsection{Table Detection Evaluations}
To identify the tabular region in the document image and regress the coordinates of a bounding box that has been designated as a tabular region is the task of table detection.Tables \ref{tab:Table Detection1},\ref{tab:Table Detection2} demonstrates how several table detection techniques that have been thoroughly researched compare in terms of performance. The ICDAR-2013, ICDAR-2017-POD, ICDAR-2019, and UNLV datasets are typically used to assess the effectiveness of table detection algorithms. 

Additionally described in Tables \ref{tab:Table Detection1},\ref{tab:Table Detection2} is the Intersection Over Union (IOU) criterion used to determine precision and recall. The most accurate results across all relevant datasets are underlined. It is important to note that several of the approaches did not specify the IOU threshold value but instead compared their findings to those of other approaches that did. So, for those procedures, we have taken into consideration the same threshold value.

\subsubsection{Table Recognition Evaluations}
Table recognition entails both segmenting the structure of tables (the task of structural segmentation is assessed based on how precisely the rows or columns of the tables are split) and extracting the data from the cells. We will review the evaluations of the few previously discussed approaches in this part. 

Table \ref{tab:Table Recognition} provides a summary of the findings. It is important to note that the different datasets and evaluation measures used in these procedures mean that the provided methodologies are not directly comparable to one another.

\subsection{Open source code}
Several open source frameworks for creating generic deep learning models, most of which are written in Python, are available online, including TensorFlow, Keras, PyTorch, and MXNet.The open-source projects for table detection and structure recognition  are summarized in Table \ref{tab:sourcecode}. Many of the authors have also made open-source implementations of their proposed models available. TensorFlow and PyTorch are the most often utilized frameworks in these open source projects.

\section{Conclusion}
In the field of document analysis, table analysis is a significant and extensively researched problem.  The challenge of interpreting tables has been dramatically transformed and new standards have been set thanks to the use of deep learning ideas. 
\\

As we said at the paper's main contribution's paragraph at the Introduction section,
we have addressed several current processes that have advanced the process of information extraction from tables in document pictures by implementing deep learning concepts.  We have discussed methods that use deep learning to detect, identify, and classify tables.  We have also shown the most and least well-known techniques that have been used to detect and identify tables, respectively. 
\\

As we did at section 7, all of the datasets that are publicly accessible and their access details have been compiled.  On numerous datasets, we have presented a thorough performance comparison of the methodologies that have been addressed.  On well-known datasets that are freely accessible to the public, state-of-the-art algorithms for table detection have produced results that are almost flawless.  Once the tabular region has been identified, the work of structurally segmenting tables and then recognizing them follows. 
\\

We conclude that both of these areas still have opportunities for development. 

 \bibliographystyle{elsarticle-num} 
 \bibliography{elsarticle-template-num}






\end{document}